\documentclass[11pt, a4paper, twoside]{report}

\usepackage[utf8]{inputenc}
\usepackage[english]{babel}
\usepackage{csquotes}
\usepackage{amsmath}
\usepackage{amssymb}
\usepackage{amsthm}
\usepackage{graphicx}
    \graphicspath{ {./figures/} }
\usepackage[left=2.5cm, right=2.5cm, top=3cm, bottom=3cm, twoside]{geometry}
\usepackage{float}
\usepackage[section]{placeins}
\usepackage{subcaption}
\usepackage{newpxtext}
\usepackage[vvarbb]{newpxmath}
\usepackage{xfrac}
\usepackage[super]{nth}

\usepackage[dvipsnames,table,xcdraw]{xcolor}
\usepackage{bm}
	
\usepackage[mathcal]{eucal}
\usepackage{tikz}
    \usetikzlibrary{cd, positioning, arrows.meta, backgrounds, calc, 3d}
    \tikzstyle{every picture}+=[remember picture]
\usepackage{enumitem}
\usepackage[many,breakable]{tcolorbox}
\usepackage[backend=biber,
    sorting=nyt,
    maxbibnames=99,
    style=authoryear-comp,
    dashed=false,
    backref=true
]{biblatex}
\usepackage{hyperref}
    \hypersetup{
    colorlinks=true,
    linkcolor=RoyalBlue,
    filecolor=magenta,      
    urlcolor=RoyalBlue,
    citecolor=RoyalBlue,
    breaklinks=true,
    linktoc=all,
}
\usepackage{enumitem}
    \setlist{itemsep=.2em}
\usepackage{fancyhdr}
\usepackage{epigraph}

\usepackage{array}

\DeclareEmphSequence{\bfseries,\mdseries}

\DefineBibliographyStrings{english}{%
  backrefpage = {page},% originally "cited on page"
  backrefpages = {pages},% originally "cited on pages"
}

\theoremstyle{definition}
\newtheorem{theorem}{Theorem}[chapter]

\newtheorem{lemma}[theorem]{Lemma}
\newtheorem{corollary}[theorem]{Corollary}

\theoremstyle{definition}
\newtheorem{definition}[theorem]{Definition}
\newtheorem{example}[theorem]{Example}

\theoremstyle{remark}
\newtheorem{remark}[theorem]{Remark}

\theoremstyle{definition}

\tcolorboxenvironment{definition}{
	blanker,
	before skip=\topsep,
	after skip=\topsep,
	enhanced jigsaw,
	colback=black!10!white,
	left=5pt,
	right=5pt, % I'd avoid this % don't tell me what to do
	top=5pt,
	bottom=5pt,
	rounded corners=all,
	arc=0pt,
	outer arc=0pt
}

\tcolorboxenvironment{theorem}{
	blanker,
	before skip=\topsep,
	after skip=\topsep,
	enhanced jigsaw,
	colback=black!10!white,
	left=5pt,
	right=5pt, % I'd avoid this
	top=5pt,
	bottom=5pt,
	rounded corners=all,
	arc=0pt,
	outer arc=0pt
}

\tcolorboxenvironment{lemma}{
	blanker,
	before skip=\topsep,
	after skip=\topsep,
	enhanced jigsaw,
	colback=black!10!white,
	left=5pt,
	right=5pt, % I'd avoid this
	top=5pt,
	bottom=5pt,
	rounded corners=all,
	arc=0pt,
	outer arc=0pt
}

\tcolorboxenvironment{proposition}{
	blanker,
	before skip=\topsep,
	after skip=\topsep,
	enhanced jigsaw,
	colback=black!10!white,
	left=5pt,
	right=5pt, % I'd avoid this
	top=5pt,
	bottom=5pt,
	rounded corners=all,
	arc=0pt,
	outer arc=0pt
}

\tcolorboxenvironment{remark}{
	blanker,
	before skip=1.5em,
	after skip=1.5em,
	borderline west={1.5pt}{1.5pt}{black},
	breakable,
	left=8pt,
	right=8pt, % I'd avoid this
}

\DeclareMathOperator*{\argmin}{arg\,min}

\DeclareMathOperator*{\Var}{Var}

\renewcommand\d[1]{\mathop{}\!\mathit{d}\nobreak\hspace{-0.1em}#1}

\DeclareMathOperator{\diam}{diam}
\DeclareMathOperator{\supp}{supp}

\DeclareMathOperator{\relu}{ReLU}

\newcommand{\rident}[1]{\mathrm{#1}}
\newcommand{\iident}[1]{\mathit{#1}}

\makeatletter
\newcommand*{\bigcdot}{}% Check if undefined
\DeclareRobustCommand*{\bigcdot}{%
  \mathbin{\mathpalette\bigcdot@{}}%
}
\newcommand*{\bigcdot@scalefactor}{.7}
\newcommand*{\bigcdot@widthfactor}{1.15}
\newcommand*{\bigcdot@}[2]{%
  \sbox0{$#1\vcenter{}$}% math axis
  \sbox2{$#1\cdot\m@th$}%
  \hbox to \bigcdot@widthfactor\wd2{%
    \hfil
    \raise\ht0\hbox{%
      \scalebox{\bigcdot@scalefactor}{%
        \lower\ht0\hbox{$#1\bullet\m@th$}%
      }%
    }%
    \hfil
  }%
}
\makeatother

\title{\textsc{\huge Mathematics of Neural Networks}}

\author{\scshape Bart M.N. Smets}

\date{November 12, 2022}

\begin{document}

\captionsetup{width=.75\linewidth, format=hang, labelfont=bf, font=small}

\maketitle

\epigraph{\itshape ``You insist that there is something a machine cannot do. If you tell me precisely what it is a machine cannot do, then I can always make a machine which will do just that.''}{\scshape -- John von Neumann}

\epigraph{\itshape ``As a technical discussion grows longer, the probability of someone suggesting deep learning as a solution approaches $1$.''}{\scshape -- comment on Youtube anno 2020}

\vfill

\begin{center}
\parbox{12cm}{
Thanks to my colleagues Gijs Bellaard, Remco Duits, Jim Portegies and Alessandro Di Bucchianico for valuable input and feedback during the writing of these lecture notes.
}
\end{center}
%\paragraph{Acknowledgements}
%...

{
\hypersetup{linkcolor=black}
\tableofcontents
}

\newpage
\pagestyle{fancy}
%%%%%%%%%%%%%%%%%%%%%%%%%%%%%%%%%%%%%%%%%%%%%%%%%%%%%%%%%%%%%%%%%%%%%%%%%%%%%%%%%

\chapter*{Introduction}
\addcontentsline{toc}{chapter}{Introduction}

The last decade has seen great experimental progress being made in machine learning, spearheaded by deep learning methods that make use of so-called \emph{deep neural networks}.
Many challenging, high-dimensional tasks that were previously beyond reach have become feasible with remarkably simple (in the algorithmic sense) techniques coupled with modern computational resources.
Particularly in the fields of computer vision and natural language processing, deep learning is currently the go-to tool.

Machine learning is usually positioned under the umbrella of \emph{artificial intelligence}. 
While artificial intelligence is a fairly nebulous and broad term, the term machine learning refers specifically to algorithms that can improve their performance at a given task when presented with more data about the problem.
Machine learning algorithms are used in a wide variety of applications such as speech recognition and computer vision, where it is difficult or impossible to develop conventional algorithms to perform the required task.
These learning algorithms generally start from very general model that is then \emph{trained} based in sample data to learn how to perform a specific task.
When the underlying model being used is a (deep) neural network then we speak about deep learning.

The idea of using computational models that are inspired by the workings of biological neurons dates as far back as \textcite{mcculloch1943logical}.
Over the following decades more of the ingredients that we think of a standard today were added: training the network on data was tried by \textcite{ivakhnenko1966cybernetic}, \textcite{fukushima1987neural} originated the ancestor of our current convolutional neural networks and \textcite{lecun1989backpropagation} introduced backpropagation as a training mechanism for neural networks.

However, non of these efforts led to a breakthrough in the use of neural networks in practice and such models were mainly regarded as a academic curiosity during this time period.
This state of affairs only changed at the start of the new millennium with the appearance of programmable \emph{GPUs} (Graphics Processing Unit), while initially designed with rendering 3D graphics in mind these devices could be leveraged for other purposes, such as by \textcite{oh2004gpu} for neural networks.
This eventually led to such breakthroughs as on the ImageNet \parencite[][]{deng2009imagenet} image classification challenge  by \textcite{krizhevsky2012imagenet}, where neural networks managed to dominate other, more traditional, techniques.
These events can be thought of as the start of the modern deep learning era.

Despite more than a decade of impressive experimental results, theoretical understanding of why deep learning works as well as it does is lacking.
This presents an opportunity both for understanding and improvement, particularly for mathematicians.

This course presents an introduction to neural networks from the point of view of a mathematician.
We cover the basic vocabulary and functioning of neural networks in Chapter~\ref{ch:basics}.
In Chapter~\ref{ch:deeplearning} we look at deep neural networks and the associates techniques that allow them to work.
Chapter~\ref{ch:equivariance} covers a novel application of geometry to neural networks.
We will discuss how the theory of Lie groups and homogeneous spaces can be leveraged to endow neural networks with certain structural symmetries, i.e. make them equivariant under certain geometric transformations.

% Chapter~\ref{ch:universality} contains some (unfinished) material about the universal approximation property of neural networks but is not part of the course material.

%%%%%%%%%%%%%%%%%%%%%%%%%%%%%%%%%%%%%%%%%%%%%%%%%%%%%%%%%%%%%%%%%%%%%%%%%%%%%%%%%
%%%
%%%
%%%
%%%
%%%
%%%%%%%%%%%%%%%%%%%%%%%%%%%%%%%%%%%%%%%%%%%%%%%%%%%%%%%%%%%%%%%%%%%%%%%%%%%%%%%%%
\chapter{The Basics}
\label{ch:basics}

%%%%%%%%%%%%%%%%%%%%%%%%%%%%%%%%%%%%%%%%%%%%%%%%%%%%%%%%%%%%%%%%%%%%%%%%%%%%%%%%%
\section{Supervised Learning}
\label{sec:supervisedlearning}

While there are different kinds of machine learning, we will be focusing on \emph{supervised learning}.
Supervised learning is a machine learning paradigm where the available data consists of a pairing of inputs with know, correct, outputs.
What is unknown is exactly how the mapping between those inputs and outputs works.
The goal is to infer, from the available data, the general structure of the mapping in the hope that this will generalize to unseen situations.
Formally we can state the problem as follows.

\paragraph{The supervised learning problem}
Given an unknown function $f:X \to Y$ between spaces $X$ and $Y$, find a good approximation of $f$ using only a dataset of $N$ samples:
\begin{equation*}
    \mathcal{D} = \big\{ \vspace{2em}\left( x_i, y_i \right) \big\}_{i=1}^N
    \quad\text{ with }\quad y_i = f(x_i) \quad\text{for all $i=1 \ldots N$}.
\end{equation*}
The space $X$ is also called the \emph{feature space} and its elements are referred to as feature vectors.
The space $Y$ is also called the \emph{label space} and its elements are referred to as labels.

\begin{example}
\label{example:dogcat}
Let $X$ be the space of all images of cats and dogs and $f$ the classifier that maps to $Y=\left\{ \text{"dog"}, \text{"cat"} \right\}$.
Which we can express numerically as $X = [0,1]^{3 \times H \times W}$ (i.e. an image of height $H$ and width $W$ with 3 color channels) and $Y=[0,1]^2$ (one probability for each of the two classes).
\end{example}

The general approach to solving this problem consists of three main steps.
\begin{enumerate}
    \item Choose a model $F:X \times W \to Y$ parametrized by a parameter space $W$ ($W$ for weights, since that is how parameters in NNs are often referred to).
    
    \item We need to quantify the quality of the model's output, so we need to choose a \emph{loss function} $\ell: Y \times Y \to \mathbb{R}$, where $\ell(y_1,y_2)$ indicates ``how different'' $y_1$ and $y_2$ are.
    
    \item Based on the dataset and the loss function choose $w \in W$ so that $F(\ \cdot\ ; w)$ is the ``best'' possible approximation of the target function $f$.
\end{enumerate}

This high-level approach leaves some open questions however.
\begin{itemize}
    \item How to choose the model?
    \item How to choose the loss function?
    \item How to optimize?
\end{itemize}
None of these questions have canonical answers and are largely handled by trial-and-error and heuristics simply because we have to resolve them before we can make any progress.
Regardless, these choices will have a large impact on the final result even though they are in no way supported by the available data or some form of first principle.
The collective set of assumptions we make to be able to proceed with the problem is called \emph{inductive bias} in the machine learning field.

For the purpose of this course we will of course pick neural networks as our models of choice and we will talk about those later.
The second thing we have to do is chose a loss function.
A \emph{loss function} is a function
$\ell:Y\times Y \to \mathbb{R}$, it works much like a metric, but it is less restrictive.
\begin{itemize}
    \item Usually but not always $Y \times Y \to \mathbb{R}^+$
    
    \item Since we want to minimize loss, the minimum should exist so that $\min_{y \in Y} \ell(y_0,y)$ is well posed.
    
    \item Differentiable (a.e. at least) would be necessary since we want to do gradient descent.
    
    \item Metric properties like identity of indiscernibles, symmetry and triangle inequality need not hold.
\end{itemize}

Having chosen a model and a loss function we move on to optimization, or:
what is the ``best'' choice of $w \in W$? 
The straightforward (but not necessarily preferred) answer: minimize the loss on the dataset, i.e. find:
\begin{equation*}
    w^* = \argmin_{w \in W} \sum_{i=1}^N \ell \left( F(\, x_i\,;w), y_i \right)
    .
\end{equation*}

\begin{example}[Linear least squares]
\begin{equation*}
    F(x;w) = \sum_{j=1}^n w_j \varphi_j(x)
\end{equation*}
and $\ell(y_1,y_2)=| y_1 - y_2 |^2$ for some basis functions $\{ \varphi_j \}_j$.
Then we get the familiar \emph{linear least square} setting.
\end{example}

What is the ``best'' $w\in W$ is not necessarily the one that minimizes the loss on the dataset, \emph{overfitting} is often an issue in any optimization problem.
Which brings us to \emph{regularization}.

We will discuss regularization techniques for NNs more in the future. 
But for now we will mention that \emph{parameter regularization} is a common technique from regression that is often used in NNs.
This type of regularization is characterized by the addition of a penalty term to the data loss that discourages parameter values for straying into undesirable areas (such as becoming too large).
The modified optimization problem becomes:
\begin{equation*}
    w^*
    =
    \argmin_{w \in W} \sum_{i=1}^N \ell \left( F(x_i;w), y_i \right)
    + \lambda C(w)
\end{equation*}
with $\lambda>0$ and $C:W \to \mathbb{R}^+$ to penalize complexity in some fashion.
\begin{example}[Tikhonov regularization]
$W=\mathbb{R}^n$ and $C(w) = \| w \|^2$.
In the context of neural networks this type of parameter regularization is also called \emph{weight decay}.
\end{example}

\paragraph{Regression \& classification}

Supervised learning should sound familiar by now. Indeed if $X=\mathbb{R}^n$ and $Y=\mathbb{R}^m$ then it amounts to \emph{regression}.

Regression is a large part of supervised learning but it is more generally formulated so that is encompasses other tasks such as \emph{classification}.

Classification is the ML designation for (multiclass) logistic regression where $X=\mathbb{R}^n$ and we try to assign each $x \in X$ to one of $m$ classes.
The numeric output for this type of problem is normally a discrete probability distribution over $m$ classes:
\begin{equation*}
    Y = 
    \left\{ (y_1,\ldots,y_m) \in [0,1]^m \ \middle| \ \textstyle\sum_{i=1}^m y_i = 1 \right\}.
\end{equation*}
Example~\ref{example:dogcat} is of this type.

\begin{remark}[Statistical learning theory viewpoint]

In SLT the assumption is made that the data samples $(x_i,y_i)$ are drawn i.i.d. from some probability distribution $\mu$ on $X \times Y$.
\textit{Think about why this is a fairly big leap.}
What we are then interested in is minimizing the \emph{population risk}:
\begin{equation*}
    R(w) 
    := \mathbb{E}_{(x,y) \sim \mu} \left[ \ell\left( F(x;w), y \right) \right]
    := \int_{X \times Y} \ell\left( F(x;w), y \right) \d\mu(x,y)
    .
\end{equation*}
The goal would be finding the parameter set that minimizes this population risk:
\begin{equation*}
    w^* = \argmin_{w \in W} R(w).
\end{equation*}

But in reality we do not know $\mu$ and so we cannot even calculate the population risk, let alone minimize it. 
So we do the next best thing: minimize the \emph{empirical risk}:
\begin{equation*}
    \hat{R}(w)
    :=
    \frac{1}{N} \sum_{i=1}^N  \ell\left( F(x_i;w), y_i \right)
    .
\end{equation*}
The parameter set that minimize the empirical risk is called the \emph{empirical minimizer}:
\begin{equation*}
    \hat{w} = \argmin_{w \in W} \hat{R}(w),
\end{equation*}
which is the same thing as minimizing the loss on the dataset in the supervised learning setting.

When we add a regularization term as before it is called \emph{structural risk minimization}.
\begin{equation*}
    \hat{w} = \argmin_{w \in W} \hat{R}(w) + \lambda C(w).
\end{equation*}

SLT is then concerned with studying things such as bounds on $\hat{R}(\hat{w})-R(w^*)$. For more see 2MMS80 Statistical Learning Theory.
\end{remark}

%%%%%%%%%%%%%%%%%%%%%%%%%%%%%%%%%%%%%%%%%%%%%%%%%%%%%%%%%%%%%%%%%%%%%%%%%%%%%%%%%
\section{Artificial Neurons \& Activation Functions}

\begin{figure}[H]
    \centering
    \begin{minipage}[c]{0.65\textwidth}
        \includegraphics[width=\linewidth]{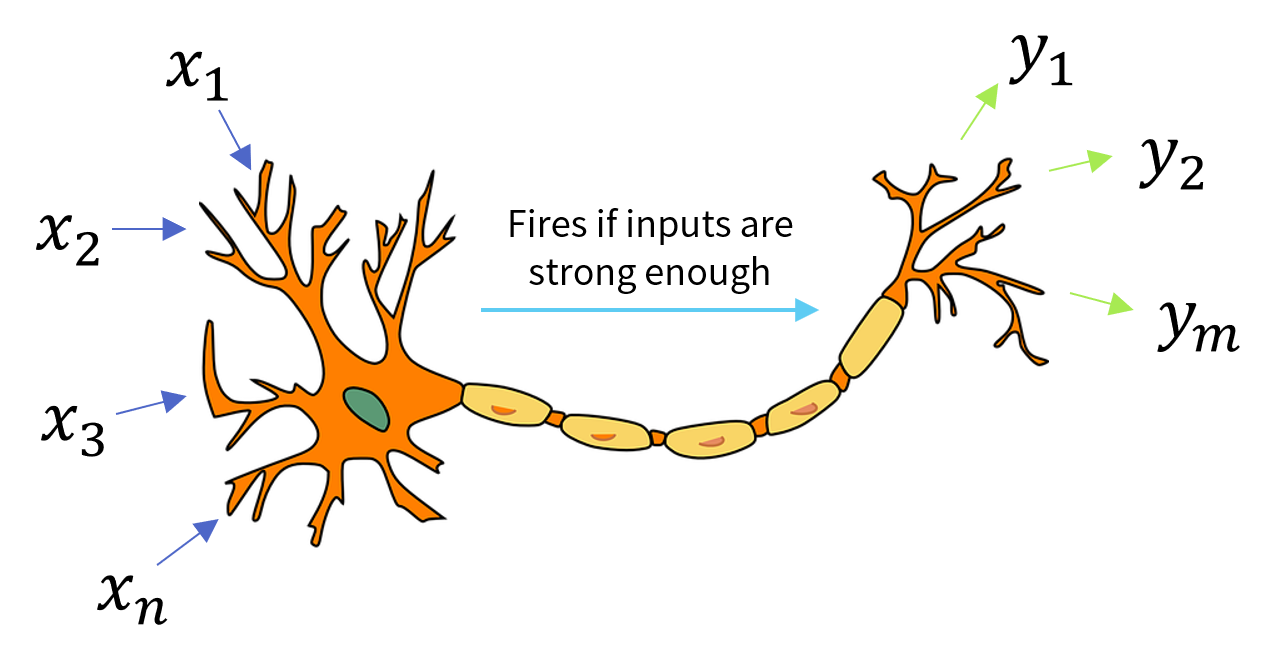}
    \end{minipage}
    ~\hfill~
    \begin{minipage}[c]{0.3\textwidth}
        \captionsetup{width=\linewidth, format=plain}
        \caption{A simplified biological neuron. The dendrites on the left receive electric signals from other neurons, once a certain threshold is reached the neuron will fire a signal along its axon and through its synapses on the right relay a signal to other neurons.}
    \label{fig:biologicalneuron}
    \end{minipage}
\end{figure}

As the name suggests, artificial neural networks are inspired by biology.
Just like their biological counterparts their constituent parts are artificial neurons.
The basic structure of a biological neuron is illustrated in Figure~\ref{fig:biologicalneuron}.
Each neuron can send and receive signals to and from other neurons so that together they form a network.

In similar fashion artificial neural networks consist of artificial neurons.
Each artificial neurons takes some inputs (usually in the form of real numbers) and produces one or more outputs (again, usually real numbers) that it passes to other neurons.
The most common model neuron is given by an affine transform followed by a non-linear function.
Let $\bm{x} \in \mathbb{R}^n$ be the input signal and $\bm{y} \in \mathbb{R}^m$ be the output signal then we calculate
\begin{equation}
    \label{eq:affineneuron}
    \bm{y} = \sigma ( A \bm{x} + \bm{b}),
\end{equation}
where $A \in \mathbb{R}^{m \times n}$, $\bm{b} \in \mathbb{R}^m$ and $\sigma$ is a choice of \emph{activation function}.
The components of the matrix $A$ are referred to as the \emph{linear weights}, the vector $\bm{b}$ is called the \emph{bias}.
The intermediate value $A \bm{x} + \bm{b}$ is sometimes referred to as the \emph{activation}.
The activation function $\sigma$ is also sometimes called the transfer function or the non-linearity.

Inputs and outputs need not span the real numbers.
Depending on the application we could encounter:
\begin{itemize}
    \item $\{0, 1\}$: binary,
    \item $\mathbb{R}^+$: non-negative reals,
    \item $[0,1]$: intervals (probabilities for example),
    \item $\mathbb{C}$: complex,
    \item etc.
\end{itemize}

A historically significant choice of activation function is the \emph{Heaviside} function, given by
\begin{equation*}
    H(x) 
    := \mathbb{1}_{x \geq 0}(x)
    :=
    \begin{cases}
        1 \qquad &\text{if } x \geq 0,
        \\
        0 &\text{else}.
    \end{cases}
\end{equation*}
A neuron of the type \eqref{eq:affineneuron} that uses the Heaviside function as its activation function is called a \emph{perceptron}.
Let us see what we can do with it.
Let $\bm{w} \in \mathbb{R}^n$ and $b \in \mathbb{R}$ then a neuron with input $\bm{x} \in \mathbb{R}^n$ and output $y \in \mathbb{R}$ would look like
\begin{equation*}
    \mathcal{N}(\bm{x}) = H(\bm{w}^T \bm{x} + b)
    =
    \begin{cases}
        1 \qquad &\text{if } \bm{w}^T \bm{x} \geq -b,
        \\
        0 &\text{if } \bm{w}^T \bm{x} < -b.
    \end{cases}
\end{equation*}
Which is nothing but a linear binary classifier on $\mathbb{R}^n$ since $\bm{w}^T \bm{x} = -b$ is a hyperplane in $\mathbb{R}^n$.
This hyperplane divides the space into two and assign the value $0$ to one half and $1$ to the other.

\begin{example}[Boolean gate]
We can (amongst other things) use a perceptron to model some elementary Boolean logic.
Let $x_1,x_2 \in \{0,1\}$ and let $\operatorname{AND}(x_1,x_2) := H(x_1+x_2-1.5)$ then the neuron behaves as follows.
\begin{center}
\begin{tabular}{ccc}
     $x_1$ & $x_2$ & $\operatorname{AND}(x_1,x_2)$  
     \\
     \hline
     0 & 0 & 0
     \\
     0 & 1 & 0
     \\
     1 & 0 & 0
     \\
     1 & 1 & 1
\end{tabular}
\end{center}
\end{example}

The Heaviside function is an example of a \emph{scalar} or \emph{pointwise} activation function. 
Often when a use a scalar function as activation function we abuse notation to let it accepts vector (and matrices) as follows. Let $\sigma: \mathbb{R} \to \mathbb{R}$ then we allow
\begin{equation*}
    \sigma \left(
        \begin{bmatrix}
            x_1 \\ \vdots \\ x_n
        \end{bmatrix}
    \right)
    \equiv
    \begin{bmatrix}
        \sigma(x_1)
        \\
        \vdots
        \\
        \sigma(x_n)
    \end{bmatrix}
    .
\end{equation*}

\begin{figure}[ht!]
    \centering
    \begin{center}
    \resizebox{\linewidth}{!}{
    \begin{tikzpicture}
\definecolor{royalblue}{rgb}{0.25, 0.41, 0.88}
\definecolor{plum}{rgb}{0.56, 0.27, 0.52}
\definecolor{pinegreen}{rgb}{0.0, 0.47, 0.44}
\definecolor{deepsaffron}{rgb}{1.0, 0.6, 0.2}

\def\xw{2.5}
\def\yw{2}

\begin{scope}[domain=-2.5:1.8]
      \draw[-latex] (-\xw,0) -- (\xw,0) node[right] {$x$};
      \draw[-latex] (0,-1.1) -- (0,\yw) node[above] {$y$};
      \draw[-] (-1,-0.05) -- (-1,0.05) node[below] {$-1$};
      \draw[-] (1,-0.05) -- (1,0.05) node[below] {$1$};
      \draw[-] (-0.05,-1) -- (0.05,-1) node[right] {$-1$};
      \draw[-] (-0.05,1) -- (0.05,1) node[right] {$1$};
      
      \draw[color=royalblue,variable=\x,style=very thick] plot ({\x},{max(\x,0)}) node[below right] {$ReLU(x)$};
\end{scope}

\begin{scope}[domain=-2.5:2.5, shift={(6,0)}]
	\draw[-latex] (-\xw,0) -- (\xw,0) node[right] {$x$};
      \draw[-latex] (0,-1.1) -- (0,\yw) node[above] {$y$};
      \draw[-] (-1,-0.05) -- (-1,0.05) node[below] {$-1$};
      \draw[-] (1,-0.05) -- (1,0.05) node[below] {$1$};
      \draw[-] (-0.05,-1) -- (0.05,-1) node[right] {$-1$};
      \draw[-] (-0.05,1) -- (0.05,1) node[right] {$1$};
      
       \draw[color=plum,variable=\x,style=very thick] plot ({\x},{1/(1+exp(-\x))}) node[above left] {$\sigma(x)$};
\end{scope}

\begin{scope}[domain=-2.5:2.5, shift={(12,0)}]
	\draw[-latex] (-\xw,0) -- (\xw,0) node[right] {$x$};
      \draw[-latex] (0,-1.1) -- (0,\yw) node[above] {$y$};
      \draw[-] (-1,-0.05) -- (-1,0.05) node[below] {$-1$};
      \draw[-] (1,-0.05) -- (1,0.05) node[below] {$1$};
      \draw[-] (-0.05,-1) -- (0.05,-1) node[right] {$-1$};
      \draw[-] (-0.05,1) -- (0.05,1) node[right] {$1$};
      
       \draw[color=pinegreen,variable=\x,style=very thick] plot ({\x},{tanh(\x)}) node[above left] {$\tanh(x)$};
\end{scope}

\begin{scope}[domain=-2.5:2, shift={(18,0)}]
	\draw[-latex] (-\xw,0) -- (\xw,0) node[right] {$x$};
      \draw[-latex] (0,-1.1) -- (0,\yw) node[above] {$y$};
      \draw[-] (-1,-0.05) -- (-1,0.05) node[below] {$-1$};
      \draw[-] (1,-0.05) -- (1,0.05) node[below] {$1$};
      \draw[-] (-0.05,-1) -- (0.05,-1) node[right] {$-1$};
      \draw[-] (-0.05,1) -- (0.05,1) node[right] {$1$};
      
       \draw[color=deepsaffron,variable=\x,style=very thick] plot ({\x},{\x/(1+exp(-\x))}) node[above left] {$\operatorname{swish_1}(x)$};
\end{scope}

    \end{tikzpicture}
    }
    \end{center}
    \caption{Some common scalar activation functions. From left to right: the rectified linear unit, the logistic sigmoid, the hyperbolic tangent and the swish function with $\beta=1$.}
    \label{fig:scalaractivationfunctions}
\end{figure}
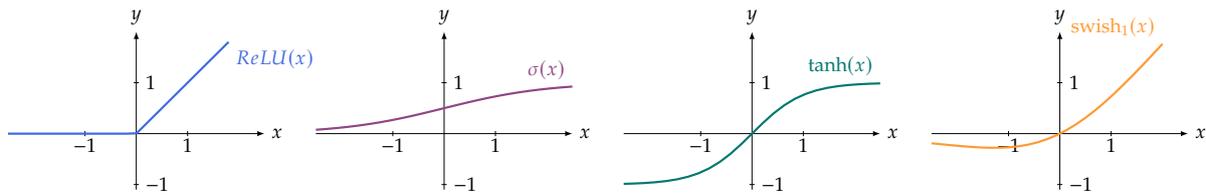

We list some commonly used scalar activation functions which are also illustrated in Figure~\ref{fig:scalaractivationfunctions}.
\begin{itemize}
    \item \emph{Rectified Linear Unit} (ReLU): arguably the most used activation function in modern neural networks, it is calculated as
        \begin{equation*}
            \sigma(\lambda) = \relu(\lambda) := \max\{0, \lambda \}
            .
        \end{equation*}
    \item \emph{Sigmoid} (also known as logistic sigmoid or soft-step):
        \begin{equation*}
            \sigma(\lambda) := \frac{1}{1 + e^{-\lambda}}
            .
        \end{equation*}
        The sigmoid was commonly used as activation function in early neural networks, which is the reason that activations functions in general are still often labeled with a $\sigma$.
    \item \emph{Hyperbolic tangent}: very similar to the sigmoid, it is given by
        \begin{equation*}
            \tanh(\lambda) := \frac{e^\lambda - e^{-\lambda}}{e^{\lambda}+e^{-\lambda}}
            .
        \end{equation*}
    \item \emph{Swish}: a more recent choice of activation function that can be thought of as a smooth variant of the ReLU.
    It is given by the multiplication of the input itself with the sigmoid function:
        \begin{equation*}
            \operatorname{swish_\beta}(\lambda)
            :=
            \lambda \, \sigma(\beta\lambda)
            :=
            \frac{\lambda}{1 + e^{-\beta \lambda}}
            ,
        \end{equation*}
        where $\beta > 0$.
        The $\beta$ parameter is usually chosen to be $1$ but could be treated as a trainable parameter if desired.
        In case of $\beta=1$, this function is also called the \emph{sigmoid-weighted linear unit} or SiLU.
\end{itemize}

Activation functions need not be scalar, we list some common multivariate functions.
\begin{itemize}
    \item \emph{Softmax}, also known as the normalized exponential function: $\operatorname{softmax}:\mathbb{R}^n \to [0,1]^n$ is given by
    \begin{equation*}
        \operatorname{softmax}
        \left(
        \begin{bmatrix}
            x_1 \\ x_2 \\ \vdots \\ x_n
        \end{bmatrix}
        \right)
        :=
        \frac{1}{\sum_{i=1}^n e^{x_i}}
        \begin{bmatrix}
            e^{x_1} \\ e^{x_2} \\ \vdots \\ e^{x_n}
        \end{bmatrix}
        .
    \end{equation*}
    Softmax has the useful property that its output is a discrete probability distribution, i.e. each value is a non-negative real in the range $[0,1]$, and all the values in its output add up to exactly $1$.
    
    \item \emph{Maxpool}: here each output is the maximum of a certain subset of the inputs: the function $\operatorname{maxpool}:\mathbb{R}^n \to \mathbb{R}^m$ is given by
    \begin{equation*}
        \operatorname{maxpool}
        \left(
        \begin{bmatrix}
            x_1 \\ x_2 \\ \vdots \\ x_n
        \end{bmatrix}
        \right)
        :=
        \begin{bmatrix}
            \max_{j \in I_1} x_j
            \\
            \max_{j \in I_2} x_j
            \\ \vdots \\ 
            \max_{j \in I_m} x_j
        \end{bmatrix}
        .
    \end{equation*}
    Where for each $i \in \{ 1, \ldots, m \}$ we have a $I_i \subset \{1,\ldots,n \}$ that specifies over which inputs to take the maximum for each output.
    Maxpooling can easily be generalised by replacing the max operation with min, the average, the mean, etc.
    
    \item \emph{Normalization}, sometimes it is desirable to re-center and re-scale a signal:
        \begin{equation*}
            \operatorname{normalize}
            \left(
            \begin{bmatrix}
                x_1 \\ x_2 \\ \vdots \\ x_n
            \end{bmatrix}
            \right)
            :=
            \begin{bmatrix}
                \frac{x_1 - \mu}{\sigma}
                \\[5pt]
                \frac{x_2 - \mu}{\sigma}
                \\ \vdots \\ 
                \frac{x_n - \mu}{\sigma}
            \end{bmatrix}
            ,
        \end{equation*}
        where $\mu = \mathbb{E}[\bm{x}]$ and $\sigma^2  = \Var(\bm{x})$.
        There are many variants on normalization where the difference is how $\mu$ and $\sigma$ are computed: over time, over subsets of the incoming signals, etc.
\end{itemize}

All the previous examples of activation functions are deterministic, but \emph{stochastic} activation functions are also used. 
\begin{itemize}
    \item \emph{Dropout} is a stochastic function that is often used during the training process but is removed once the training is finished. It works by randomly setting individual values of a signal to zero with probability $p$:
    \begin{equation*}
        \left(\operatorname{dropout}_p(\bm{x})\right)_i
        :=
        \begin{cases}
            0 \qquad &\text{with probability } p,
            \\
            x_i &\text{with probability } 1-p
            .
        \end{cases}
    \end{equation*}
    
    \item \emph{Heatbath} is a scalar function that outputs $1$ or $-1$ with a probability that depends on the input:
    \begin{equation*}
        \operatorname{heatbath}(\lambda)
        :=
        \begin{cases}
            \hphantom{-}1 \qquad &\text{with probability } \frac{1}{1+e^{-\lambda}}
            \\
            -1 \qquad &\text{otherwise}.
        \end{cases}
    \end{equation*}
\end{itemize}

All the activation functions we seen are essentially fixed functions, swish and dropout have a parameter but it is usually fixed to some chosen value.
That means that the trainable parameters of a neural network are usually the linear weights and biases.
There is however no a-priori reason why that needs to be the case, in fact we will see a class of non-linear operators with trainable parameters at the end of Chapter~\ref{ch:equivariance}.
Regardless, having parameters in the non-linear part of a network is somewhat rare in practice at the time of this writing.

%%%%%%%%%%%%%%%%%%%%%%%%%%%%%%%%%%%%%%%%%%%%%%%%%%%%%%%%%%%%%%%%%%%%%%%%%%%%%%%%%
\section{Shallow Networks}

While we are interested in deep networks we start out with a look at shallow networks.
Because of their simplicity we can still approach them constructively and gain some intuition about neural networks in general along the way.
You can also think about the study of shallow networks as the study of single layers of deep networks.

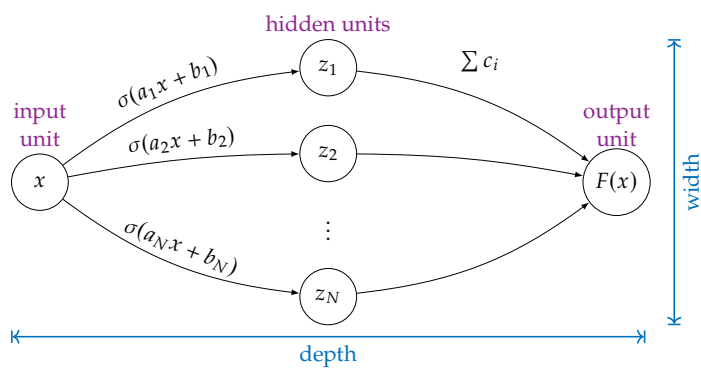
\begin{figure}[ht!]
\centering
\begin{minipage}[c]{0.65\textwidth}
\begin{center}
    \resizebox{0.9\linewidth}{!}{
    \begin{tikzpicture}
	\begin{scope}[minimum size=28]
		\node[circle,draw] (x) at (-5,0.5){$x$};
		\node[circle,draw] (z1) at (0,2.5){$z_1$};
		\node[circle,draw] (z2) at (0,1){$z_2$};
		\node[circle] (z3) at (0,-0.25){$\vdots$};
		\node[circle,draw] (z4) at (0,-1.5){$z_N$};
		\node[circle,draw] (y) at (5,0.5){$F(x)$};
	\end{scope}
	
	\draw[-latex] (x) to[bend left=15] node[above,sloped]{$\sigma(a_1 x + b_1)$} (z1);
	\draw[-latex] (x) to[bend left=5] node[above,sloped]{$\sigma(a_2 x + b_2)$} (z2);
	\draw[-latex] (x) to[bend right=15] node[above,sloped]{$\sigma(a_N x + b_N)$} (z4);
	
	\draw[-latex] (z1) to[bend left=15] node[above,yshift=10]{$\sum c_i$} (y);
	\draw[-latex] (z2) to[bend left=5] (y);
	\draw[-latex] (z4) to[bend right=15] (y);
	
	\draw[|<->|, RoyalBlue,thick] (-5.5,-2.2) -- node[below]{depth} (5.5,-2.2);
	\draw[|<->|, RoyalBlue,thick] (6,-2.0) -- node[below,sloped]{width} (6,3.0);
	
	\node (l1) [align=center, anchor=south, Plum] at (-5, 1.0) {input\\unit};
	\node (l2) [align=center, anchor=south, Plum] at (0, 3.0) {hidden units};
	\node (l3) [align=center, anchor=south, Plum] at (5, 1.0) {output\\unit};
\end{tikzpicture}
}
\end{center}
\end{minipage}
    ~\hfill~
    \begin{minipage}[c]{0.3\textwidth}
    %\vspace{0pt}
    \captionsetup{width=\linewidth, format=plain}
    \caption{Diagrammatic representation of a shallow $\mathbb{R} \to \mathbb{R}$ neural network per \eqref{eq:scalarshallownetwork}.
    In deep learning literature the input and output of a network are often referred to as the input unit respectively the output unit.
    The intermediate values are often called the hidden units.
    What is commonly referred to as the \emph{width} and \emph{depth} of the network is also indicated.
    }
    \label{fig:shallownetwork}
    \end{minipage}

\end{figure}

Let us consider a shallow ReLU network with scalar in- and output.
Let $\bm{w}=(\bm{a},\bm{b},\bm{c}) \in W=(\mathbb{R}^N)^3$ be our set of parameters for some $N \in \mathbb{N}$, then we define our model $F:\mathbb{R} \times W \to \mathbb{R}$ as
\begin{equation}
    \label{eq:scalarshallownetwork}
    F(x; \bm{w})
    :=
    \sum_{i=1}^N c_i \, \sigma(a_i x + b_i)
    ,
\end{equation}
where $\sigma$ is the ReLU.
Diagrammatically this network is represented in Figure~\ref{fig:shallownetwork}.

We will restrict ourselves to $x \in [0,1]$ for the time being.
In practice this would not be much of a restriction since real world data is compactly supported.

Let us explore what types of functions our network can express: the output is a linear combination of functions of the type
\begin{equation*}
    x \mapsto \sigma(a x + b)
    .
\end{equation*}
Which, depending on the value of $a$, gives us one of the following types of functions.

\begin{center}
\resizebox{\linewidth}{!}{
\begin{tikzpicture}
\def\xw{2.5}
\def\yw{2}

\begin{scope}[domain=-2.3:2.3]
      \draw[-latex] (-\xw,0) -- (\xw,0) node[right] {$x$};
      \draw[-latex] (0,-0.4) -- (0,\yw) node[above] {$y$};
      \node[] at (-1,2) {$a=0$};
      \draw[color=RoyalBlue,variable=\x,style=very thick] plot ({\x},{0.5}) node[above left] {$y=\sigma(b)$};
\end{scope}

\begin{scope}[domain=-2.3:2.3, shift={(6,0)}]
    \draw[-latex] (-\xw,0) -- (\xw,0) node[right] {$x$};
    \draw[-latex] (0,-0.4) -- (0,\yw) node[above] {$y$};
    \node[] at (-1,2) {$a>0$};
    \draw[-,style=thick,color=Plum] (0.5,0.1) -- (0.5,-0.1) node[below] {$-\frac{b}{a}$};
    \draw[color=Plum,variable=\x,style=very thick,samples=1000] plot ({\x},{max(\x-0.5,0)}) node[above left] {};
    \draw[-,color=Plum,style=thick] (1.2,0.7) -- (1.6,0.7) arc[start angle=0, end angle=45, radius=0.4] node[right,xshift=3,yshift=-1] {$a$};
\end{scope}

\begin{scope}[domain=-2.3:2.3, shift={(12,0)}]
	\draw[-latex] (-\xw,0) -- (\xw,0) node[right] {$x$};
    \draw[-latex] (0,-0.4) -- (0,\yw) node[above] {$y$};
    \node[] at (-1,2) {$a<0$};
    \draw[-,style=thick,color=PineGreen] (-0.5,0.1) -- (-0.5,-0.1) node[below] {$-\frac{b}{a}$};
    \draw[color=PineGreen,variable=\x,style=very thick,samples=1000] plot ({\x},{max(-\x-0.5,0)}) node[above left] {};
    \draw[-,color=PineGreen,style=thick] (-1.5,1) -- (-1.1,1) arc[start angle=0, end angle=-45, radius=0.4] node[right,xshift=3,yshift=1] {$a$};
\end{scope}

\end{tikzpicture}
}
\end{center}

All three of these classes of functions are piecewise linear functions (or piecewise affine functions really), so any linear combination of these functions would again be a piecewise linear function.
Hence, our model \eqref{eq:scalarshallownetwork} is really just a particular parameterization for a piecewise linear function on $[0,1]$.

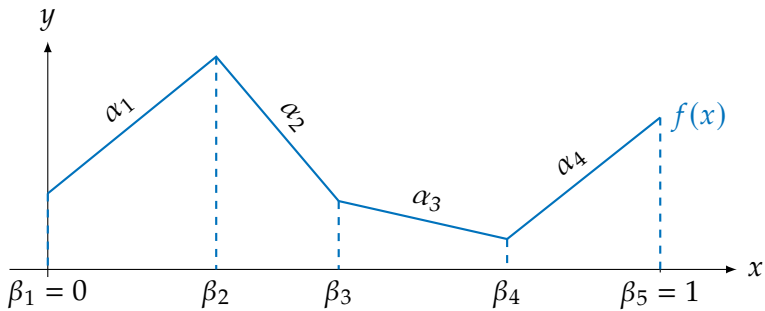
\begin{figure}[ht!]
    \centering
    \begin{minipage}[c]{0.65\textwidth}
    \begin{center}
\resizebox{\linewidth}{!}{
\begin{tikzpicture}

\coordinate (p1) at (0,1);
\coordinate (p2) at (2.2,2.8);
\coordinate (p3) at (3.8,0.9);
\coordinate (p4) at (6.0,0.4);
\coordinate (p5) at (8,2);

\begin{scope}[domain=0:8]
      \draw[-latex] (-0.5,0) -- (9,0) node[right] {$x$};
      \draw[-latex] (0,-0.1) -- (0,3) node[above] {$y$};
      %\draw[] at (0,0) node[below left] {$0$};
      \draw[-] (8,0.04) -- (8,-0.04) node[below] {};
      \draw[color=RoyalBlue,style=thick] 
        (p1)-- node[above,sloped,black]{$\alpha_1$} 
        (p2)-- node[above,sloped,black]{$\alpha_2$}
        (p3)-- node[above,sloped,black]{$\alpha_3$}
        (p4)-- node[above,sloped,black]{$\alpha_4$}
        (p5) node[right] {$f(x)$};
      \draw[dashed,color=RoyalBlue, style=thick] (p1)--(0,0) node[below,black] {$\beta_1=0$};
      \draw[dashed,color=RoyalBlue, style=thick] (p2)--(2.2,0) node[below,black] {$\beta_2$};
      \draw[dashed,color=RoyalBlue, style=thick] (p3)--(3.8,0) node[below,black] {$\beta_3$};
      \draw[dashed,color=RoyalBlue, style=thick] (p4)--(6.0,0) node[below,black] {$\beta_4$};
      \draw[dashed,color=RoyalBlue, style=thick] (p5)--(8,0) node[below,black] {$\beta_5=1$};
\end{scope}

\end{tikzpicture}
}
\end{center}
    \end{minipage}
    ~\hfill~
    \begin{minipage}[c]{0.3\textwidth}
    %\vspace{0pt}
    \captionsetup{width=\linewidth, format=plain}
    \caption{Example of a piecewise linear function on $[0,1]$ with $4$ pieces. The location of the inflection points are labeled with $\beta$'s and the slope of each piece is denoted with an $\alpha$.}
    \label{fig:piecewiselinearfunction}
    \end{minipage}
\end{figure}

Can we then represent any piecewise linear function on $[0,1]$ by a shallow ReLU network?
Let $f:[0,1] \to \mathbb{R}$ be piecewise linear function with $N$ pieces.
We will denote the inflection points with
\begin{equation*}
    0 = \beta_1 < \beta_2 < \ldots < \beta_{N+1} = 1
\end{equation*}
and the slopes (i.e. the constant derivatives of each piece) with $\alpha_1, \ldots, \alpha_N$.
An example of this setup is illustrated in Figure~\ref{fig:piecewiselinearfunction}.

Now choose a network of the type in \eqref{eq:scalarshallownetwork} with $N+1$ neurons:
\begin{equation*}
    F(x) := \sum_{i=1}^{N+1} c_i \, \sigma(a_i x + b_i)
    .
\end{equation*}
Next we choose the parameters $\bm{a},\bm{b}$ and $\bm{c}$ of the network as:
\begin{align*}
    & a_{N+1} = 0, & &b_{N+1} = 1, & &c_1=\alpha_1,
    \\
    & a_1, \ldots, a_N = 1, & &b_i = - \beta_i \ (\text{for } i = 1\ldots N), & &c_i=\alpha_i-\alpha_{i-1} \ (\text{for } i=2\ldots N).
\end{align*}
This turns our model into
\begin{equation}
    \label{eq:fittedscalarshallow}
    F(x)
    =
    f(0) + \alpha_1 x
    +
    \sum_{i=2}^N
    (\alpha_i - \alpha_{i-1})
    \,\sigma(x-\beta_i).
\end{equation}
When we examine the third term of \eqref{eq:fittedscalarshallow} we see that the ReLU's will vanish each term until $x$ reaches the appropriate threshold, i.e. $\sigma(x-\beta_i) = 0$ until $x>\beta_i$.
Hence for $x \in [\beta_1,\beta_2]$ we get the function $x \mapsto f(0)+\alpha_1 x$ which exactly matches the function $f$ in that interval.
As we proceed to the $[\beta_2,\beta_3]$ interval the first term of the right hand sum becomes non-zero and the model becomes
\begin{align*}
    x \mapsto 
    &f(0) + \alpha_1 x + (\alpha_2-\alpha_1)(x-\beta_2)
    \\
    &=
    f(0) + \alpha_1 x - \alpha_1 x + \alpha_1 \beta_2 + \alpha_2(x-\beta_2)
    \\
    &=
    f(\beta_2) + \alpha_2 (x-\beta_2),
\end{align*}
which is exactly $f$ in the interval $[\beta_2,\beta_3]$.
We can keep advancing along the $x$-axis like this and every time we pass an inflection point a new term will activate and bend the line towards a new heading.
The model can effectively be rewritten as
\begin{equation*}
    F(x)
    =
    \begin{cases}
        f(0) + \alpha_1 x \quad &\text{if } x \in [\beta_1,\beta_2],
        \\
        f(\beta_2) + \alpha_2 (x-\beta_2) &\text{if } x \in [\beta_2,\beta_3],
        \\
        \qquad \vdots
        \\
        f(\beta_{N-1}) + \alpha_N (x-\beta_{N-1}) &\text{if } x \in [\beta_{N-1},\beta_N],
    \end{cases}
\end{equation*}
which matches $f$ exactly.
Hence, we may conclude that on compact intervals the shallow scalar ReLU neural networks are exactly the space of piecewise linear functions.

Piecewise linear function are a simple class of function but can be used to approximate many other classes of function to an arbitrary degree, as the following lemma shows.
\begin{lemma}
    \label{lem:pwldense}
    Let $f \in C([0,1],\mathbb{R})$ then for all $\varepsilon > 0$ there exists a piecewise linear function $F$ so that
    \begin{equation*}
        \sup_{x\in[0,1]} \left| f(x) - F(x) \right| < \varepsilon
        .
    \end{equation*}
\end{lemma}

\begin{proof}
    Let $\varepsilon>0$ be arbitrary. Since $f$ is a continuous function on a compact domain it is also uniformly continuous on said domain, i.e.
    \begin{equation*}
        \exists \delta > 0 \ \forall x_1,x_2 \in [0,1]:
        |x_1-x_2| < \delta \Rightarrow |f(x_1)-f(x_2)| < \frac{\varepsilon}{2}.
    \end{equation*}
    Now choose $N > \frac{1}{\delta}$ and partition $[0,1]$ as $x_i = \frac{i}{N}$ for $i=0 \ldots N$.
    Define the piecewise linear function $F$ as
    \begin{equation*}
        F(x)
        =
        \sum_{i=1}^N
        \mathbb{1}_{[x_{i-1},x_i)}(x)
        \left(
            f(x_{i-1})
            +
            \frac{f(x_i)-f(x_{i-1})}{x_i-x_{i-1}} (x - x_{i-1})
        \right)
        .
    \end{equation*}
    Note that for all $i-0 \ldots N$ that $F(x_i)=f(x_i)$.
    Now let $x \in [0,1]$ be arbitrary, $x$ will always fall inside some interval $[x_{i-1},x_i)$ where we have
    \begin{align*}
        |f(x)-F(x)|
        &\leq
        |f(x)-f(x_i)| + |f(x_i)-F(x)|
        \\
        &<
        \frac{\varepsilon}{2} + |F(x_i)-F(x)|
        \\
        &<
        \frac{\varepsilon}{2} + |F(x_i)-F(x_{i-1})|
        \\
        &=
        \frac{\varepsilon}{2} + |f(x_i)-f(x_{i-1})|
        \\
        &<
        \frac{\varepsilon}{2} + \frac{\varepsilon}{2}
        =
        \varepsilon
    \end{align*}
\end{proof}

Since shallow scalar ReLU networks represent the piecewise linear function on $[0,1]$ and those piecewise linear functions are dense in $C([0,1])$ we get the following.

\begin{corollary}
    \label{cor:universal}
    Shallow scalar ReLU neural networks of arbitrary width \eqref{eq:scalarshallownetwork} are universal approximators of $C([0,1])$ in the supremum norm.
\end{corollary}

This corollary is our first \emph{universality} result. In the context of deep learning, universality is the study of what classes of functions can be approximated arbitrarily well by particular neural networks architectures.
Notice that Corollary~\ref{cor:universal} has 4 ingredients:
\begin{enumerate}
    \item the type of neural network (shallow scalar ReLU network),
    \item the growth direction of the network (width),
    \item the space of function to approximate ($C([0,1])$),
    \item and how the approximation is measured (with the supremum norm).
\end{enumerate}

Generalizing this universal approximation result to $\mathbb{R}^n$ is not possible with just one layer, one of the reasons deep networks are necessary.
In general constructive proofs such as for Lemma~\ref{lem:pwldense} are not possible/available and we will have to contend ourselves with existence results.

Universality is theoretically interesting since it tells us that neural networks can in principle closely approximate most reasonable types of functions (continuous, $L^p$, etc.), thus explaining to some degree why they are powerful.
But universality does not consider many other important facets of neural networks in practice:
\begin{itemize}
    \item economy of representation: how does the number of parameters scale with the desired accuracy,
    \item economy of finding a good approximation: just because a good approximation exist does not mean it is easy to find.
\end{itemize}

We will not discuss universality further for now. %Chapter~\ref{ch:universality} has more.
Just remember that, under some mild assumptions, neural networks are universal approximators.

%%%%%%%%%%%%%%%%%%%%%%%%%%%%%%%%%%%%%%%%%%%%%%%%%%%%%%%%%%%%%%%%%%%%%%%%%%%%%%%%%
\section{Stochastic Gradient Descent}

Recall the elements of supervised learning:
\begin{itemize}
    \item a dataset $\mathcal{D} = \{ (x_i,y_i) \in X \times Y \}_{i=1}^N$ for some input space $X$ and output space $Y$,
    \item a model $F:X \times W \to Y$ for some parameter space $W$,
    \item and a loss function $\ell:Y \times Y \to \mathbb{R}$.
\end{itemize}
We can then look at the total loss over the dataset for a given choice of parameters:
\begin{equation}
    \label{eq:totaldatasetloss}
    \ell_{\text{total}}(w)
    :=
    \frac{1}{|\mathcal{D}|} \sum_{(x,y) \in \mathcal{D}} \ell \left( F(x;w), y \right)
    .
\end{equation}
Note how the loss is a function of the parameters $w$ only, it could also include some additional regularization terms (as in Section~\ref{sec:supervisedlearning}) but we will leave those details aside for now.

Knowing little about $F$ we cannot find a minimum directly, but assuming everything is differentiable we can use gradient descent:
\begin{equation*}
    w_{t} = w_{t-1} - \eta\, \nabla_{w} \ell_{\text{total}} (w_{t-1})
    ,
\end{equation*}
where $\eta > 0$ is called the \emph{learning rate} and where we chose some $w_0 \in W$ as our starting point.

Sadly, doing this calculation is not practical in the real world due to:
\begin{itemize}
    \item the datasets being huge,
    \item the models having an enormous amount of parameters.
\end{itemize}
Just calculating $\ell_{\text{total}}$ is a major undertaking, calculating $\nabla_w \ell_{\text{total}}$ is entirely out of the question.

Instead, we take a divide and conquer approach.
We randomly divide the dataset into (roughly) equal \emph{batches} and tackle the problem batch by batch.
Denote a batch by an index set $I \subset \{ 1 \ldots N \}$, then we consider the loss over that single batch:
\begin{equation}
    \label{eq:batchloss}
    \ell_{I}(w)
    :=
    \frac{1}{|I|}
    \sum_{i \in I}
    \ell \left( F(x_i; w), y_i \right)
    .
\end{equation}

Say we divide the dataset into batches $I_1,I_2,\ldots, I_B$, we can then split our gradient descent step into $B$ smaller steps:
\begin{equation}
    \label{eq:gradientdescentupdate}
    w_{t}
    =
    w_{t-1}
    -
    \eta\, \nabla_w \ell_{I_t}(w_{t-1})
\end{equation}
where we again assume some starting point $w_0 \in W$.
When we reach $t=B$ and have exhausted all the batches we say we have complete an \emph{epoch}.
Subsequently, we generate a new set of random batches and repeat the process.
This process is called \emph{stochastic gradient descent}, or SGD for short, the stochastic referring the the random selecting of batches.

If the batch $I$ is uniformly drawn then
\begin{equation*}
    \mathbb{E} \left[
        \nabla_w \ell_I(w)
    \right]
    =
    \nabla_w \ell_{\text{total}}(w),
\end{equation*}
i.e. the gradient of a uniformly drawn random batch is an unbiased estimator of the gradient of the full dataset.

\begin{remark}[Higher order methods]
Gradient descent (stochastic or not) is a first order method since it relies only on the first order derivatives.
Higher order methods are rarely used with neural networks because of the large amount of parameters.
A model with $N$ parameters has $N$ first order derivatives but $N^2$ second order derivatives, since neural network commonly have millions to billions of parameters calculating second order derivatives is simply not feasible.
\end{remark}

A simple but important modification to the (stochastic) gradient descent algorithm is \emph{momentum}.
Instead of taking each step based only on the current gradient we take into account the direction we were moving in in previous steps.
The modified gradient descent step is given by
\begin{equation*}
\begin{split}
    v_t &= \mu \, v_{t-1} - \eta\,\nabla_w \ell_{I_t}(w_{t-1}),
    \\
    w_t &= w_{t-1} + v_t,
\end{split}
\end{equation*}
where we initialize with $v_0 = 0$ and $\mu\in[0,1)$ is called the \emph{momentum coefficient}.
The variable $v_t$ can be interpreted as the current velocity vector along which we are moving in the parameter space.
At each iteration our new velocity takes a fraction of the previous velocity (controlled by $\mu$) and updates it with the current gradient.
Observe that for $\mu=0$, i.e. no momentum, we revert to \eqref{eq:gradientdescentupdate}. 
The larger we take $\mu$ the more influence the history of the gradients has on our current step.

\section{Training}

In practice the \emph{training process} involves more than applying SGD on the whole dataset.
One thing we need to keep in mind is that the dataset is the only real information we have about the underlying function we are trying to approximate, so there is no way to judge how good our approximation is outside of using the dataset.
Given that we know that neural networks are universal approximators we can always find a neural networks that perfectly fits any given dataset; making overfitting a given.
What we are really after is \emph{generalization}: the ability of our neural network to give the correct output for inputs it has not seen before.

We accomplish this by splitting the dataset and only using part of it to train the network, the remaining part of the dataset (that the network has never seen) we use to test how well our network has generalized.
The exact split depends on the situation but using $80\%$ of the dataset for training and $20\%$ for testing is a good rule-of-thumb.

Let us denote our training and testing datasets as
\begin{equation*}
    \mathcal{D}_{\text{train}}
    ,
    \mathcal{D}_{\text{test}}
    \subset
    \mathcal{D}
    .
\end{equation*}
We then perform SGD on the training dataset, i.e. try to minimize the training loss:
\begin{equation}
    \label{eq:trainingloss}
    \ell_{\text{train}}(w)
    :=
    \frac{1}{|\mathcal{D}_{\text{train}}|}
    \sum_{(x,y) \in \mathcal{D_{\text{train}}}}
    \ell(F(x;w),y)
    .
\end{equation}
But we judge the performance of our network by the testing loss:
\begin{equation}
    \label{eq:testingloss}
    \ell_{\text{test}}(w)
    :=
    \frac{1}{|\mathcal{D}_{\text{test}}|}
    \sum_{(x,y) \in \mathcal{D_{\text{test}}}}
    \ell(F(x;w),y)
    .
\end{equation}

\begin{figure}[ht!]
\centering
\begin{minipage}[c]{0.67\textwidth}
\begin{center}
\resizebox{\linewidth}{!}{
\begin{tikzpicture}
    \begin{scope}[domain=0.1:5]
      \draw[-latex] (-0.1,0) -- (5.2,0) node[below left] {epochs};
      \draw[-latex] (0,-0.1) -- (0,4) node[above] {loss};
      \draw[color=RoyalBlue,variable=\x,style=very thick,samples=500] plot ({\x},{1/(\x+0.2)}) node[above left] {train};
      \draw[color=Plum,variable=\x,style=very thick,samples=500] plot ({\x},{0.25+exp(\x/2.4)*1/(\x+0.2)}) node[above left] {test};
      
      \draw[thick,dashed] (1.8,1) ellipse (0.45 and 0.8) node[align=center,above,yshift=0.9cm]{stop\\training\\here};
\end{scope}

\begin{scope}[domain=0.1:5, shift={(-6,0)}]
    \draw[-latex] (-0.1,0) -- (5.2,0) node[below left] {epochs};
    \draw[-latex] (0,-0.1) -- (0,4) node[above] {loss};
    \draw[color=RoyalBlue,variable=\x,style=very thick,samples=500] plot ({\x},{1/(\x+0.2)}) node[above left] {train};
    \draw[color=Plum,variable=\x,style=very thick,samples=500] plot ({\x},{0.8+1/(1.4*\x+0.2)}) node[above left] {test};
\end{scope}	
\end{tikzpicture}
}
\end{center}
\end{minipage}
    ~\hfill~
    \begin{minipage}[c]{0.3\textwidth}
    %\vspace{0pt}
    \captionsetup{width=\linewidth, format=plain}
    \caption{
    Typical progression of the training and testing loss.
    The training loss will generally converge to some very low value.
    The testing loss either behaves in a similar fashion and will converge on some higher value, as is illustrated on the left.
    The testing loss could also start to increase again at some point, as on the right, this indicates overfitting and tells you when to stop training.
    }
    \label{fig:trainigprogress}
    \end{minipage}
\end{figure}
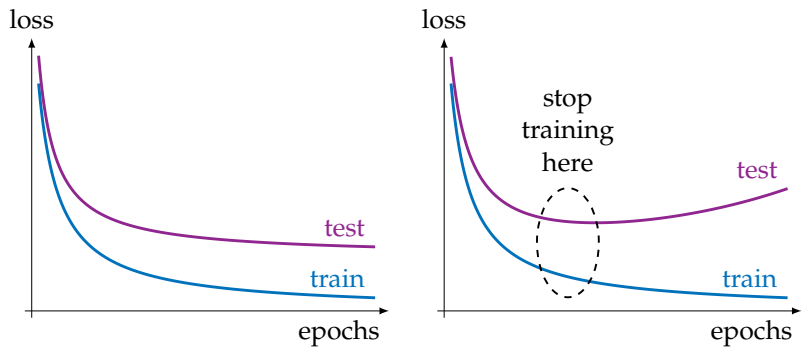

By monitoring the testing loss during training we are able to judge for how long we should train, how well our network generalizes and when overfitting sets in.
Figure~\ref{fig:trainigprogress} illustrates the typical behaviour of loss curves that are encountered during training.

When we are designing a network for a given application we usually do repeated training while we play with the \emph{hyperparameters} to try and improve the testing loss.
This can lead us to overfit the hyperparameters to the given testing dataset.
There are two things we can do to mitigate this:
\begin{enumerate}
    \item split the dataset in three parts: training/testing/validation,
    \item redo the split randomly every time we retrain.
\end{enumerate}
In the author's opinion the second method is preferable since it is easier to implement and even if the first method is used it is still necessary to redo the split if we decide to make some changes after having used the validation dataset (since if we use it more than once we are back where we started).

\begin{remark}[Hyperparameters]
    The hyperparameters are all parameters that we ourselves select and are not trainable parts of the model. Examples are: batch size, learning rate, size of the model, choice of activation function, ratio of the dataset split, etc.
\end{remark}

%%%%%%%%%%%%%%%%%%%%%%%%%%%%%%%%%%%%%%%%%%%%%%%%%%%%%%%%%%%%%%%%%%%%%%%%%%%%%%%%%
%%%
%%%
%%%
%%%
%%%
%%%%%%%%%%%%%%%%%%%%%%%%%%%%%%%%%%%%%%%%%%%%%%%%%%%%%%%%%%%%%%%%%%%%%%%%%%%%%%%%%
\chapter{Deep Learning}
\label{ch:deeplearning}

The recent advances in machine learning were made largely with so called deep neural networks.
The \emph{deep} in this context refers to neural networks where the input will pass through many neurons for processing, an arrangement not dissimilar to the brain and in contrast to our previous shallow networks.
Once the potential of these networks was demonstrated the term \emph{deep learning} was coined as a synonym for doing machine learning with deep neural networks.
Deep neural networks are a very broad class of models that contains a host of various architectures for different applications, see \textcite{wikipedia2021typesofnetworks} for an overview. We will only cover two main architectures but much of what we will discuss is transferable to other types of networks. 

While deep networks can be very powerful there are challenges in getting them to work properly and many aspects of these networks are not yet well understood. 
This chapter will cover what deep neural networks are and the techniques and algorithms that are necessary to make them work.

%%%%%%%%%%%%%%%%%%%%%%%%%%%%%%%%%%%%%%%%%%%%%%%%%%%%%%%%%%%%%%%%%%%%%%%%%%%%%%%%%
\section{Deep Neural Networks}

\subsection{Feed Forward Networks}

The quintessential example of a deep neural network is the \emph{feed forward} neural network wherein connections between nodes do not form cycles.
A feed forward network simply takes a set of shallow networks and concatenates them, feeding the output of one layer of neurons as input into the next layer of neurons.
Let us formalize this construction.

Let $L \in \mathbb{N}$ and $N_0,N_1,\ldots,N_{L+1} \in \mathbb{N}$.
Let $\sigma_1,\ldots,\sigma_L$ be activation functions, which we will assume to be scalar functions that we apply pointwise for the moment.
Let $F_i : \mathbb{R}^{N_{i-1}} \to \mathbb{R}^{N_i}$ be affine transforms given by
\begin{equation*}
    F_i (\bm{x}) := A_i \bm{x} + \bm{b}_i
\end{equation*}
where $A_i \in \mathbb{R}^{N_i \times N_{i-1}}$ and $\bm{b}_i \in \mathbb{R}^{N_i}$ for $i \in \{ 1, \ldots, L+1\}$.
Then we call $\mathcal{N} : \mathbb{R}^{N_0} \to \mathbb{R}^{N_{L+1}}$ given by
\begin{equation}
    \label{eq:feedforward}
    \mathcal{N}
    :=
    F_{L+1}
    \circ 
    \sigma_L
    \circ
    F_L
    \circ
    \cdots
    \circ
    \sigma_1
    \circ
    F_1
\end{equation}
a \emph{feed forward neural network}. 
This network is said to have $L$ \emph{(hidden) layers}, $N_i$ is said to be the \emph{width} of the layer $i$ and the maximum of all the layer's widths, i.e. $\max_{i=1}^L \{ N_i \}$, is also called the width of the network.

We usually label inputs with $\bm{x}$, outputs with $\bm{y}$ and if necessary the intermediate results with $\bm{z}^{(i)} \in \mathbb{R}^{N_i}$ as follows:
\begin{equation*}
    \bm{z}^{(0)} := \bm{x}
    ,
    \qquad
    \bm{z}^{(i)} := \sigma_i \left( F_i(\bm{z}^{(i-1)}) \right)
    ,
\end{equation*}
for $i \in \{ 1,\ldots, L+1\}$.

\begin{figure}[ht!]
\centering
\begin{minipage}[c]{0.65\textwidth}
%\vspace{0pt}
\def\layersep{2.6cm}
\def\depthsep{1.4cm}
\resizebox{\linewidth}{!}{
\begin{tikzpicture}[
   shorten >=1pt
   , shorten <=1pt
   ,-latex,
   draw=black!50,
    node distance=\layersep,
    every pin edge/.style={<-,shorten <=1pt},
    neuron/.style={circle,fill=black!25,minimum size=24pt,inner sep=0.5pt},
    input neuron/.style={neuron, fill=RoyalBlue!50},
    output neuron/.style={neuron, fill=Plum!50},
    hidden neuron/.style={neuron, fill=PineGreen!50},
    annot/.style={text width=4em, text centered}
]

% Draw the input layer nodes
\foreach \name / \y in {1,...,4}
    \node[input neuron] (I-\name) at (0,-\y*\depthsep) {$x_{\y}$};

\foreach \name / \y in {1,...,3}
    \node[hidden neuron] (H1-\name) at (1*\layersep,-\y*\depthsep-0.5*\depthsep) {$z^{(1)}_{\y}$};
    
\foreach \name / \y in {1,...,5}
    \node[hidden neuron] (H2-\name) at (2*\layersep,-\y*\depthsep+0.5*\depthsep) {$z^{(2)}_{\y}$};
    
\foreach \name / \y in {1,...,4}
    \node[hidden neuron] (H3-\name) at (3*\layersep,-\y*\depthsep) {$z^{(3)}_{\y}$};

% Draw the output layer node
\foreach \name / \y in {1,...,2}
    \node[output neuron] (O-\name) 
    at (4*\layersep,-\y*\depthsep-1*\depthsep) {$y_{\y}$};

\foreach \source in {1,...,4}
    \foreach \dest in {1,...,3}
       \path (I-\source) edge (H1-\dest);

\foreach \source in {1,...,3}
    \foreach \dest in {1,...,5}
       \path (H1-\source) edge (H2-\dest);
       
\foreach \source in {1,...,5}
    \foreach \dest in {1,...,4}
       \path (H2-\source) edge (H3-\dest);

\foreach \source in {1,...,4}
    \foreach \dest in {1,...,2}
       \path (H3-\source) edge (O-\dest);

\path (H1-3) edge[bend right=80] node[midway,below, text=black!75] {\textsf{skip connection}} (H3-4);

\draw[|<->|, thick, draw=black!75] 
(4.6*\layersep,-0.2*\depthsep) -- (4.6*\layersep, -4.8*\depthsep) node[midway, fill=white, text=black!75] {\textsf{width}};

\draw[|<->| ,thick,draw=black!75] 
(0, -5.8*\depthsep) -- (4*\layersep, -5.8*\depthsep) node[midway, fill=white, text=black!75] {\textsf{depth}};

\end{tikzpicture}
}
\end{minipage}
~\hfill~
\begin{minipage}[c]{0.3\textwidth}
%\vspace{0pt}
\captionsetup{width=\linewidth, format=plain}
\caption{Graph representation of a feed forward neural network with 3 hidden layers. Each node is computed as the activation function applied to an affine combination of its inputs. Passing signals to deeper layers rather than the next layer is also a common feature of these types of networks and is called a skip connection.}
\label{fig:feedforward}
\end{minipage}
\end{figure}
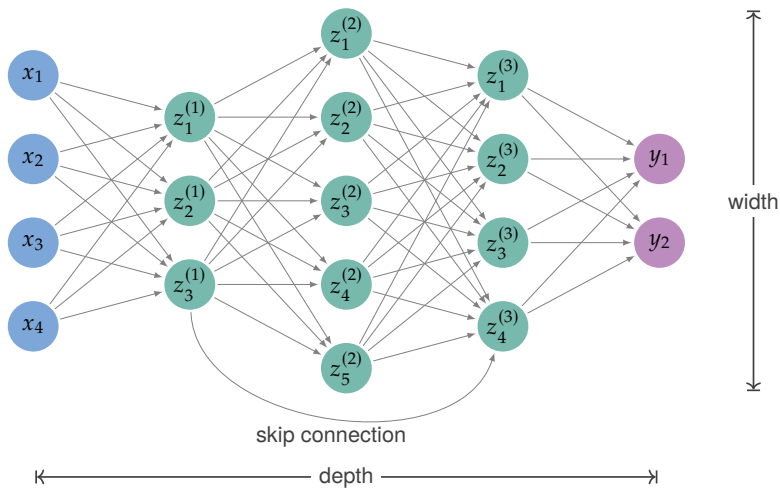

Variants to this architecture are still called feed forward networks.
For example we could include multi-variate activation functions such as soft-max or max pooling, this would require us to specify the input and output widths of the layers differently but we could still write the network down as in \eqref{eq:feedforward}.
Another common feature is a \emph{skip connection}: some of the outputs of a layer are passed to layers deeper down rather than (or in addition) to the next layer.
An example of a feed forward network is illustrated in graph form in Figure~\ref{fig:feedforward}.

\begin{remark}[Recurrent neural networks]
    The other major class of neural networks besides feed forward are the \emph{recurrent neural networks}, these networks allow for cycles to be present in their graphs.
    Recurrent neural networks are mainly used for processing sequential data such as in speech and natural language applications.
    See \textcite{wikipedia2021typesofnetworks} for a survey.
\end{remark}

Previously we saw how a shallow neural network with ReLU activation functions results in a piecewise linear function.
If we want to model a more complex function we needed to increase the width of the network, the number of linear pieces would scale linearly with the amount of neurons.

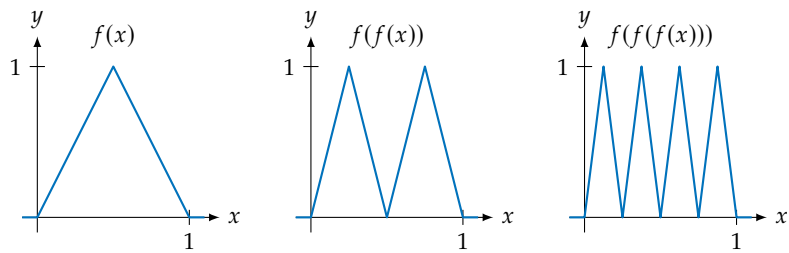
\begin{figure}[ht!]
\centering
\begin{minipage}[c]{0.65\textwidth}
\begin{tikzpicture}[scale=2.0, every node/.style={scale=0.8}]
    \begin{scope}
        \draw[-latex] (-0.1,0) -- (1.2,0) node[right] {$x$};
        \draw[-latex] (0,-0.1) -- (0,1.2) node[above] {$y$};
        \draw[-] (1,0.05) -- (1,-0.05) node[below] {$1$};
        \draw[-] (0.05,1) -- (-0.05,1) node[left] {$1$};
        
        \draw[RoyalBlue, thick, line join=round] 
        (-0.1,0) -- (0,0) -- (0.5,1) -- (1,0) -- (1.1,0);
        
        \node[] at (0.5,1.2) {$f(x)$};
    \end{scope}
    
    \begin{scope}[xshift=1.8cm]
        \draw[-latex] (-0.1,0) -- (1.2,0) node[right] {$x$};
        \draw[-latex] (0,-0.1) -- (0,1.2) node[above] {$y$};
        \draw[-] (1,0.05) -- (1,-0.05) node[below] {$1$};
        \draw[-] (0.05,1) -- (-0.05,1) node[left] {$1$};
        
        \draw[RoyalBlue, thick, line join=round] 
        (-0.1,0) -- (0,0) -- (0.25,1) -- (0.5,0) -- (0.75,1) -- (1,0) -- (1.1,0);
        
        \node[] at (0.5,1.2) {$f(f(x))$};
    \end{scope}
    
    \begin{scope}[xshift=3.6cm]
        \draw[-latex] (-0.1,0) -- (1.2,0) node[right] {$x$};
        \draw[-latex] (0,-0.1) -- (0,1.2) node[above] {$y$};
        \draw[-] (1,0.05) -- (1,-0.05) node[below] {$1$};
        \draw[-] (0.05,1) -- (-0.05,1) node[left] {$1$};
        
        \draw[RoyalBlue, thick, line join=round]
        (-0.1,0) -- (0,0) -- (0.125,1) -- (0.25,0) -- (0.375,1) -- (0.5,0)
        -- (0.625,1) -- (0.75,0) -- (0.875,1) -- (1,0) -- (1.1,0);
        
        \node[] at (0.5,1.2) {$f(f(f(x)))$};
    \end{scope}
\end{tikzpicture}
\end{minipage}
~\hfill~
\begin{minipage}[c]{0.3\textwidth}
    \captionsetup{width=\linewidth, format=plain}
    \caption{Iterative application of the function $f$ from \eqref{eq:sawtooth} causes an exponential increase in the complexity of the result as measured by the amount of linear pieces.}
    \label{fig:sawtooth}
\end{minipage}
\end{figure}

Consider a sawtooth function as in Figure~\ref{fig:sawtooth}, if we wanted to model a sawtooth with $n$ teeth with a shallow ReLU network we would need $2 n$ neurons.
But consider the network for $1$ tooth:
\begin{equation}
    \label{eq:sawtooth}
    f(x) := \relu(2 x) - \relu(4 x - 2) + \relu(2 x - 2)
    ,
\end{equation}
and concatenate it multiple times: we would also get a more and more complex sawtooth for every concatenation as is shown in Figure~\ref{fig:sawtooth}.

In fact we would get a doubling of the number of teeth every time we reapply $f$ and so an exponential increase in the amount of linear pieces in the output. 
This is the major benefit of depth: the complexity of the functions a network can model grows faster with depth than with width.

This is not to say that we should design our networks with maximum depth and minimum width.
As is usually the case in engineering there are tradeoffs to consider and any real network strikes a balance between width and depth.

\begin{remark}
    While in a shallow ReLU network each linear piece is independent in a deep ReLU network this is not the case.
    This can be understood by considering that the number of parameters in a deep ReLU network scales linearly with the number of layers while the number of pieces of the output scales exponentially, at some point there will not be enough parameters to describe any piecewise linear function with that amount of pieces.
    With an equivalent shallow ReLU network the output space is exactly the space of all piecewise linear functions with that amount of pieces.
\end{remark}

\subsection{Vanishing and Exploding Gradients}

One difficulty that arises with deep networks is in training.
A parameter in an early layer of the network, it has to pass through many layers before finally contributing to the loss.

\begin{figure}[ht!]
\centering
\begin{minipage}[c]{0.65\textwidth}
\centering
\begin{tikzpicture}[scale=3.0, every node/.style={scale=1.0}]
        \draw[-latex] (-1.2,0) -- (1.3,0) node[right] {$a$};
        \draw[-latex] (0,-0.1) -- (0,1.2) node[above] {};
        \draw[-] (1/4,0.05) -- (1/4,-0.05) node[below] {$1$};
        \draw[-] (0.05,1) -- (-0.05,1) node[left] {$1$};
        
        \draw[RoyalBlue, thick, smooth, samples=100 ,domain=-1.2:1.2] plot({\x},{1/(1+exp(-4*\x))});

        \draw[Plum, thick, smooth, samples=100 ,domain=-1.2:1.2] plot({\x},{(1/(1+exp(-4*\x)))*(1-(1/(1+exp(-4*\x))))});
        
        \node[] at (0.6,1.05) {$\sigma(a)$};
        \node[] at (0.6,0.2) {$\sigma'(a)$};
        
        \begin{scope}[on background layer]
            \fill [right color=black!60, left color=white, fill opacity=0.3] (0.8, -0.1) rectangle (1.2, 1.2);
            \fill [right color=white, left color=black!60, fill opacity=0.3] (-0.8, -0.1) rectangle (-1.2, 1.2);
            \node [white, rotate=90, font=\sffamily] at (1.0, 0.5) {saturation};
            \node [white, rotate=90, font=\sffamily] at (-1.0, 0.5) {saturation};
        \end{scope}

\end{tikzpicture}
\end{minipage}
~\hfill~
\begin{minipage}[c]{0.3\textwidth}
    \captionsetup{width=\linewidth, format=plain}
    \caption{The sigmoid activation function and its derivative. Observe that the derivative is at most $\sfrac{1}{4}$ and goes to zero quickly as $a \to \pm\infty$, this is called saturation.}
    \label{fig:vanishing}
\end{minipage}
\end{figure}
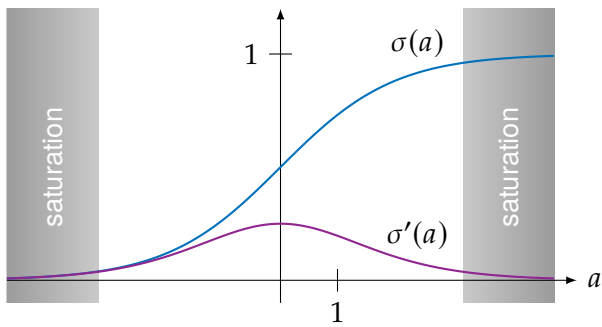

Let us disregard the affine transforms of a network for the moment and focus on the activation functions.
Let $\sigma$ be the sigmoid activation function given by $\sigma(a):=\frac{1}{1+e^{-a}}$, then define
\begin{equation*}
    \sigma^N := \underbrace{\sigma \circ \sigma \circ \cdots \circ \sigma}_{N}.
\end{equation*}
For computing the derivative of $\sigma^N$ we apply the chain rule to find the recursion
\begin{equation*}
    \frac{\partial}{\partial a} \sigma^N (x)
    =
    \sigma' \left( \sigma^{N-1}(a) \right)
    \,
    \frac{\partial}{\partial x} \sigma^{N-1}(a).
\end{equation*}
From Figure~\ref{fig:vanishing} we deduce that $0 < \sigma'(a) \leq \sfrac{1}{4}$ so
\begin{equation*}
    \left|
        \frac{\partial}{\partial a}
        \sigma^N(a)
    \right|
    \leq
    \left( \frac{1}{4} \right)^N
    .
\end{equation*}
Consequently performing a gradient descent step of the form
\begin{equation*}
    a_{i+1} = a_i - \eta \,
    \frac{\partial}{\partial a} \sigma^N (a_i)
    ,
\end{equation*}
would not change the parameter very much, a problem which becomes worse with every additional layer.
Worse is that $\sigma'(a)$ goes to zero quickly for large absolute values of $a$, as can be seen in Figure~\ref{fig:vanishing}.
In this regime, when $\sigma(a)$ is close to $0$ or $1$, we say the sigmoid is \emph{saturated}.

Theoretically we could train for longer to compensate for very small gradients but in practice this also does not work because of how floating point math works.
Specifically when working with floating point numbers we have
\begin{equation*}
    a + b = a
    \quad
    \text{if}
    \quad
    |b| \ll |a|,
\end{equation*}
hence a small enough update to a parameter does not actually change the parameter.

This behaviour of gradients becoming smaller and smaller the deeper the network becomes is called the \emph{vanishing gradient problem}.
The opposite phenomenon can also occur where the gradients become larger and larger leading to unstable training, this is called the \emph{exploding gradient problem}.

We can contrast the behaviour of the sigmoid with the rectified linear unit in the same situation. 
Recall that the ReLU is defined as $\relu(x):=\max\{0,x\}$, then we define
\begin{equation*}
    \relu^N := \underbrace{\relu \circ \relu \circ \cdots \circ \relu}_{N}.
\end{equation*}
Calculating the derivative we get:
\begin{equation}
    \label{eq:relun_deriv}
    \frac{\partial}{\partial a} \relu^N (a)
    =
    \begin{cases}
        \ 1 \quad &\text{if } a > 0,
        \\
        \ 0 &\text{else},
    \end{cases}
\end{equation}
which does not depend on the number of layers $N$.
Conceptually \eqref{eq:relun_deriv} can be interpreted as a ReLU allowing for a $1$ derivative for parameters that are currently affecting the loss (regardless of which layer the parameter resides in) and giving a zero derivative for parameters that do not.
The ReLU (at least partially) sidestepping the vanishing/exploding gradient problem accounts for their popularity in deep neural networks.

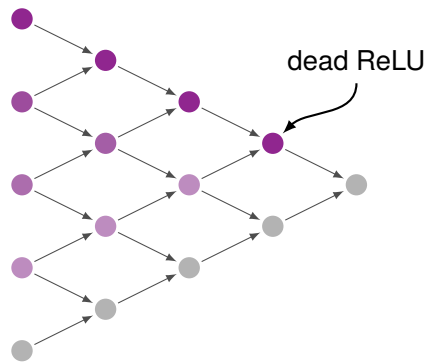
\begin{figure}[ht!]
\centering
\begin{minipage}[c]{0.65\textwidth}
\centering
\begin{tikzpicture}[
    scale=1.1
    ,shorten >=1pt
    ,shorten <=1pt
    ,-latex
    ,draw=black!70
    ,neuron/.style={circle,fill=black!30, minimum size=8pt, inner sep=0.5pt}
]

\node[neuron] (L1-0) at (0,0) {};
\node[neuron, fill=Plum!50] (L1-1) at (0,1) {};
\node[neuron, fill=Plum!67] (L1-2) at (0,2) {};
\node[neuron, fill=Plum!84] (L1-3) at (0,3) {};
\node[neuron, fill=Plum!100] (L1-4) at (0,4) {};

\node[neuron] (L2-0) at (1,0.5) {};
\node[neuron, fill=Plum!50] (L2-1) at (1,1.5) {};
\node[neuron, fill=Plum!75] (L2-2) at (1,2.5) {};
\node[neuron, fill=Plum!100] (L2-3) at (1,3.5) {};

\node[neuron] (L3-0) at (2,1) {};
\node[neuron, fill=Plum!50] (L3-1) at (2,2) {};
\node[neuron, fill=Plum!100] (L3-2) at (2,3) {};

\node[neuron] (L4-0) at (3,1.5) {};
\node[neuron, fill=Plum!100] (L4-1) at (3,2.5) {};

\node[neuron] (L5-0) at (4,2) {};

\path (L1-0) edge (L2-0);
\path (L1-1) edge (L2-0);
\path (L1-1) edge (L2-1);
\path (L1-2) edge (L2-1);
\path (L1-2) edge (L2-2);
\path (L1-3) edge (L2-2);
\path (L1-3) edge (L2-3);
\path (L1-4) edge (L2-3);

\path (L2-0) edge (L3-0);
\path (L2-1) edge (L3-0);
\path (L2-1) edge (L3-1);
\path (L2-2) edge (L3-1);
\path (L2-2) edge (L3-2);
\path (L2-3) edge (L3-2);

\path (L3-0) edge (L4-0);
\path (L3-1) edge (L4-0);
\path (L3-1) edge (L4-1);
\path (L3-2) edge (L4-1);

\path (L4-0) edge (L5-0);
\path (L4-1) edge (L5-0);

\node [font=\sffamily] (label1) at (4,3.5) {dead ReLU};
\draw[-latex, black, thick] (label1) to[in=45, out=-90] (L4-1);

%\draw[smooth cycle, black!20, tension=0.5] 
%plot coordinates{(-0.3, 0.7)(2.3, 1.7)(2.3, 3.3)(-0.3,4.3)};

\end{tikzpicture}
\end{minipage}
~\hfill~
\begin{minipage}[c]{0.3\textwidth}
%\vspace{0pt}
\captionsetup{width=\linewidth, format=plain}
\caption{A dead ReLU (i.e. with negative argument and so zero derivative) will cause the gradients of the upstream neurons to also become zero, this effect will cascade all the way through to the start of the network. Neurons that feed into non-dead ReLUs will still have a gradient via those live connections. The purple shading of the nodes indicates the degree to which each node is affected by the dead ReLU.}
\label{fig:relu_shadow}
\end{minipage}
\end{figure}

This is not to say that ReLUs are without issues.
If an input to a ReLU is negative it will have a gradient of zero, we say the ReLU or the neuron is \emph{dead}.
Consequently all neurons in the previous layer that only (or predominantly) feed into that neuron will also have a gradient of zero (or close to zero).
You could visualize this phenomenon as a dead neuron casting a shadow on the lower layers as is illustrated in Figure~\ref{fig:relu_shadow}.

During training, which neurons are dead and where the shadow is cast changes from batch to batch.
As long as most neurons are not perpetually in the shadow we should be able to properly train the network.
However, if we happen to get an unlucky initialization it is quite possible that a large part of our network is permanently stuck with zero gradients, this phenomenon is called the \emph{dying ReLU problem}.

Of course, the choice of activation function is not the only thing we need to consider.
The gradients in a real network also depend on the current values of the parameters of the affine transforms.
So we still need to initialize those parameters to values that do not cause vanishing or exploding gradient problem right from the start.
We will get back to parameter initialization later on.

\subsection{Scaling to High Dimensional Data}

If we consider all components of a matrix $A_i$ from \eqref{eq:feedforward} to be trainable parameters then we say layer $i$ is \emph{fully connected} or \emph{dense}, meaning all outputs of the layer depend on all inputs.
When all layers are fully connected we say that the network is fully connected.
Having all components of the matrices $A_i$ and vectors $\bm{b}_i$ as trainable parameters is of course not required.
In fact the general fully connected feed forward network from \eqref{eq:feedforward} is not often used in practice, but many architectures are specializations of this type adapted to specific applications.
Specialization in this context primarily means choosing which components of the $A_i$'s and $\bm{b}_i$'s are trainable and which are fixed to some chosen constant.

The primary reason fully connected network are not used is that they simply do not scale to high-dimensional data.
Consider an application where we want to apply a transform on a $1000 \times 1000$ color image.
The input and output would be elements of the space $\mathbb{R}^{3 \times 1000 \times 1000}$, a linear transform in that space would be given by a matrix
\begin{equation*}
    A \in \left( \mathbb{R}^{3 \times 1000 \times 1000} \right)^2
    .
\end{equation*}
This matrix has $9 \cdot 10^{12}$ components.
Storing that many numbers in 32-bit floating point format would take a whopping 36 terabytes, and that is just a single matrix.
Clearly this rules out fully connected networks for high dimensional applications such as imaging.

There are 3 strategies employed to deal with this problem:
\begin{enumerate}[label=\arabic*)]
    \item sparsity,
    \item weight sharing,
    \item parameterization.
\end{enumerate}

\emph{Sparsity} mean employing sparse matrices for the linear operations. 
When we fix most entries of a matrix to zero we do not need to store those entries and we simplify the calculation that we need to perform since we already know the result of the zeroed out parts.

With \emph{weight sharing} the same parameter is reused at multiple locations in the matrix.
In this case we just have to store the unique parameters and not the whole matrix.
Weight sharing is technically a special case of the last strategy.

With \emph{parameterization} we let the components of the matrix depend on a (small) number of parameter. In this case we store the parameters and compute the matrix components when required.

The following matrices are an example comparing the fully connected case with the three proposed strategies:
\begin{equation*}
    \underset{
        \strut
        \textsf{(fully connected)}
    }
    {
    \begin{bmatrix}
        a & b & c
        \\
        d & e & f
        \\
        g & h & i
    \end{bmatrix}
    }
    ,
    \qquad
    \underset{
        \strut
        \textsf{(sparse)}
    }
    {
    \begin{bmatrix}
        a &  & 
        \\
         &  & 
        \\
         & b & \hphantom{b}
    \end{bmatrix}
    }
    ,
    \qquad
    \underset{
        \strut
        \textsf{(shared weights)}
    }
    {
    \begin{bmatrix}
        a & b & b
        \\
        b & a & b
        \\
        b & b & a
    \end{bmatrix}
    }
    ,
    \qquad
    \underset{
        \strut
        \textsf{(parametrized)}
    }
    {
    \begin{bmatrix}
        f_{1}(a,b) & f_{2}(a,b) & f_{3}(a,b)
        \\
        f_{4}(a,b) & f_{5}(a,b) & f_{6}(a,b)
        \\
        f_{7}(a,b) & f_{8}(a,b) & f_{9}(a,b)
    \end{bmatrix}
    }
    ,
\end{equation*}
we need to store 9 numbers for the fully connected matrix but only 2 numbers for each of the other matrices.

These 3 strategies are not mutually exclusive and we will look at an architecture that uses a combination of sparsity and weight sharing later on.

%%%%%%%%%%%%%%%%%%%%%%%%%%%%%%%%%%%%%%%%%%%%%%%%%%%%%%%%%%%%%%%%%%%%%%%%%%%%%%%%%
\section{Initialization}

For deep network, the initial values of the parameters make a significant difference to the functioning of the SGD algorithms.
Not in the least because initial parameters being too small/large in magnitude would cause vanishing/exploding gradient problems and cause immediate issues for training.
In this section we will develop some stochastic initialization schemes that provide a functional starting point for a network to train from.
We generally use stochastic methods for initialization since we need initial parameter values in a layer to be sufficiently different.
Imagine if the values were the same, then their gradients would also be the same and they would never diverge from each other, locking the network in an untrainable state.
The most straightforward method of achieving this is drawing values from a probability distribution, which is currently the common practice.
Usually the distributions that are used are either the normal $\mathcal{N}(0,\sigma^2)$ or the uniform $\rident{Unif}[-a,a]$, where we need to pick $\sigma^2$ or $a$ to avoid training issues such as vanishing/exploding gradient.
Each group of parameters (the linear coefficients and the biases in each layer) typically get assigned their own distribution.
We will look at how these distributions are currently chosen.

\begin{remark}[Deterministic initialization scheme]
    If you recall, in the tutorial notebook {\tt 4\_FunctionApproximationIn1D.ipynb} we developed a deterministic initialization scheme that outperformed the default stochastic scheme.
    We did have to use our insight about the problem as a whole (beyond what the data provided) to do it.
    It is quite conceivable that for a given application a deterministic scheme can be developed that gives a much better starting point for training than the current crop of stochastic schemes.
\end{remark}

\begin{remark}[On the importance of good initialization]
    A good initialization scheme allows the network to reach a higher performance level in less time.
    This can have important consequences for large production networks.
    One such network is GPT-3 \parencite[see][]{brown2020language}, which is used for natural language processing, it has 175 billion parameters and training it has reportedly cost 4 million \$.
    Hence better initialization schemes can be of substantial economic value.
\end{remark}

\subsection{Stochastic Initialization}

To develop a suitable probability distribution to initialize parameters with we start by looking at the input to a neuron as a vector valued random variable $X \in \mathbb{R}^n$.
The output of a neuron is then a random variable $Y \in \mathbb{R}^m$ per
\begin{equation*}
    Y = \sigma \left(  A X +\bm{b} \right)
    ,
\end{equation*}
for some constant matrix $A \in \mathbb{R}^{m \times n}$, bias vector $\bm{b} \in \mathbb{m}$ and activation function $\sigma$.
Or alternatively written out per component:
\begin{equation}
    \label{eq:Y_i}
    Y_i = \sigma \left( \sum_{j=1}^n A_{ij} X_j + b_i \right)
    .
\end{equation}
The idea is to then initialize $A$ and $\bm{b}$ so that the variance of the signal does not change too much from layer to layer:
\begin{equation*}
    \sum_{i=1}^m \Var(Y_i)^2
    \approx
    \sum_{j=1}^n \Var(X_j)^2
    .
\end{equation*}
Controlling the $L^2$ norm of the signal variances is not necessarily the only possibility here, but it is the choice we will proceed with.

Calculating variances of functions of random variables is difficult in general.
The schemes we will be looking at depend on the following approximation.

\begin{lemma}
    \label{lem:variance_of_function}
    Let $X$ be a real valued random variable and let $f:\mathbb{R}\to\mathbb{R}$ be a differentiable function then
    \begin{equation*}
        \Var\left( f(X) \right)
        \approx
        f' \left( \mathbb{E}[X] \right)^2 \cdot \Var(X),
    \end{equation*}
    assuming the variance of $X$ is finite and $f'$ is differentiable.
\end{lemma}

\begin{proof}
    Let $m := \mathbb{E}[X]$ and approximate $f$ by its linearization $f(X) \approx f(m) + f'(m) (X - m)$.
    Then we find
    \begin{align*}
        \Var\left( f(X) \right)
        &=
        \mathbb{E}\left[ \left( f(X) - \mathbb{E}[f(X)] \right)^2 \right]
        \\
        &\approx
        \mathbb{E}\left[ \left( f(m) + f'(m) (X - m) - \mathbb{E}[f(m) + f'(m) (X - m)] \right)^2 \right]
        \\
        &=
        \mathbb{E}\left[ \left( f(m) + f'(m) (X - m) - f(m) - f'(m) \,\mathbb{E}[(X - m)] \right)^2 \right]
        \\
        &=
        \mathbb{E}\left[ \left( f'(m) (X - m) \right)^2 \right]
        \\
        &=
        f'(m)^2 \, \mathbb{E}\left[ (X-m)^2 \right]
        \\
        &=
        f' \left( \mathbb{E}[X] \right)^2 \cdot \Var(X).
    \end{align*}
\end{proof}

Intuitively, the approximation in Lemma~\ref{lem:variance_of_function} can be read as: if the variance of $X$ is small and $f'$ is reasonably bounded then the variance of $f(X)$ will also be small.

Applying Lemma~\ref{lem:variance_of_function} to \eqref{eq:Y_i} yields
\begin{equation*}
    \Var(Y_i)
    \approx
    \sigma'  \left( \sum_{j=1}^n A_{i j} \, \mathbb{E}[X_j] + b_i \right)^2
    \ 
    \left( \sum_{j=1}^n A_{i j}^2 \ \Var(X_j) \right)
    .
\end{equation*}
To make progress let us assume $\mathbb{E}[X_1]=\ldots=\mathbb{E}[X_n]$ and $\Var(X_1)=\ldots=\Var(X_n)$, then
\begin{equation*}
    \Var(Y_i)
    \approx
    \underbrace{
    \sigma'  \left( \mathbb{E}[X_1] \sum_{j=1}^n A_{i j} + b_i \right)^2
    \ 
    \left( \sum_{j=1}^n A_{i j}^2  \right)}_{\text{ideally}\ \approx 1} \ \Var(X_1)
    .
\end{equation*}
When the bracketed term is approximately 1 then the variances of the outputs $Y_i$ are about the same as those of the inputs $X_j$.

Let us now turn the $A_{i j}$'s and $b_i$'s into random variables: $A_{i j} \sim \mu_1$ and $b_i \sim \mu_2$ for all $i$ and $j$ in their respective ranges and where $\mu_1$ and $\mu_2$ are some choice of scalar probability distributions. 
Ideally we would choose $\mu_1$ and $\mu_2$ so that
\begin{equation}
    \label{eq:E=1}
    \mathbb{E}\left[ 
        \sigma'  \left( \mathbb{E}[X_1] \sum_{j=1}^n A_{i j} + b_i \right)^2
        \ 
        \left( \sum_{j=1}^n A_{i j}^2  \right)
    \right]
    =
    1.
\end{equation}
This expression allows us to put a condition on our choice of probability distributions that ensures that the variances of the signals between layers stay under control (at least at the start of training).
The following examples show how \eqref{eq:E=1} is utilized.

\begin{example}[Sigmoid with balanced inputs]
    Like before assume $\mathbb{E}[X_1]=\ldots=\mathbb{E}[X_n]$ and $\Var(X_1)=\ldots=\Var(X_n)$ and our goal is choosing a probability distribution for the linear coefficients and one for the biases.
    Say $\sigma$ is the sigmoid activation function and we have $\mathbb{E}[X_i]=0$, i.e. we have balanced inputs.
    Additionally we want balanced parameter initialization, i.e. $\mathbb{E}[b_i]=0$ and $\mathbb{E}[A_{i j}]=0$ for all $i,j$ in their respective ranges.
    Since $\mathbb{E}[X_i]=0$ \eqref{eq:E=1} reduces to:
    \begin{equation*}
        \mathbb{E}\left[ 
            \sigma'  \left( b_i \right)^2
            \ 
            \left( \sum_{j=1}^n A_{i j}^2  \right)
        \right]
        =
        \mathbb{E}\left[ \sigma'  \left( b_i \right)^2  \right]
        \,
        \mathbb{E}\left[ 
             \sum_{j=1}^n A_{i j}^2 
        \right],
    \end{equation*}
    since the $b_i$'s and $A_{i j}$'s are independent.
    We know that $0 < \sigma'(x) \leq \sfrac{1}{4}$ and that the maximum is achieved at $x=0$.
    Hence for that first factor to not become too small we need the variance of $b_i$ to be small since we already decided on setting $\mathbb{E}[b_i]=0$.
    Of course the smallest possible variance is zero, so let us be uncompromising and fix $b_i=0$ for all $i$.
    Thus the previous expression becomes
    \begin{equation*}
        \sigma'(0)^2 \ n \, \mathbb{E} \left[ A_{i j}^2 \right]
        =
        \frac{n}{16} \, \mathbb{E} \left[ (A_{i j} - 0)^2 \right]
        =
        \frac{n}{16} \, \mathbb{E} \left[ (A_{i j} - \mathbb{E}[A_{i j}])^2 \right]
        =
        \frac{n}{16} \, \Var(A_{i j}),
    \end{equation*}
    which equals 1 if
    \begin{equation*}
        \Var(A_{i j}) = \frac{16}{n}.
    \end{equation*}
    So we could choose our probability distribution for the linear coefficients to be the normal distribution $\mathcal{N}(0,\sfrac{16}{n})$ or the uniform distribution $\rident{Unif}\left[ -\sfrac{4\sqrt{3}}{n},  \sfrac{4\sqrt{3}}{n}\right]$.
    Of course the choice of the type of distribution is free as long as the expected value is zero and the variance is $\sfrac{16}{n}$, but in practice you will usually only encounter normal or uniform distributions.
    In any case this choice of expected values and variances will provide some assurance that the signals will not explode or die out as they travel through the network (at least at the start of training).
\end{example}

\begin{example}[ReLU with balanced inputs]
    Again assume $\mathbb{E}[X_1]=\ldots=\mathbb{E}[X_n]$ and $\Var(X_1)=\ldots=\Var(X_n)$.
    This time we use the ReLU activation function and we are going to initialize both our linear coefficients and biases with a uniform distribution $\rident{Unif}[-a,a]$, our goal is choosing $a>0$ in a suitable manner.
    Assume again that the inputs are balanced, i.e. $\mathbb{E}[X_i]=0$, then the expression from \eqref{eq:E=1} simplifies to
    \begin{align*}
        \mathbb{E} \left[
            \relu'\left( b_i\right)^2
            \ 
            \left( \sum_{j=1}^n A_{i j}^2 \right)
        \right]
        &=
        \mathbb{E} \left[
            \mathbb{1}_{b_i>0}
        \right]
        \ 
        \mathbb{E}\left[
        \left( \sum_{j=1}^n A_{i j}^2 \right)
        \right]
        \\
        &=
        \mathbb{P} \left( b_i > 0 \right)
        \ 
        n \, \Var(A_{i j})
        \\
        &=
        \frac{n}{2} \, \Var(A_{i j})
        \\
        &= \frac{n a^2}{6},
    \end{align*}
    which equals 1 if $a=\sqrt{\sfrac{6}{n}}$ so we would draw our $A_{i j}$'s and $b_i$'s from $\rident{Unif}\left[ - \sqrt{\sfrac{6}{n}}, \sqrt{\sfrac{6}{n}} \right]$.
\end{example}

\subsection{Xavier Initialization}

The initialization schemes from the previous section focused on controlling the variance of the signals going forward through the network.
While this does help in controlling the vanishing/exploding gradient problem we can also look at gradients directly as they backpropagate through the network, this is the approach taken by \textcite{glorot2010understanding}. The first author's name is Xavier Glorot and for that reason the scheme we will be seeing is commonly referred to as Glorot or Xavier initialization (as it is in PyTorch for \href{https://pytorch.org/docs/stable/nn.init.html#torch.nn.init.xavier_uniform_}{example}).

The idea is to treat the partial derivatives of the loss function with regards to the linear coefficients and biases as random variables as well.
Consider a setting with a linear activation function and no bias:
\begin{equation*}
    Y_i = \sum_{j=1}^n A_{i j} X_j,
\end{equation*}
where we assume all inputs $X_j$ are distributed i.i.d. with zero mean.
We additionally want to initialize our coefficients $A_{i j}$ with mean zero as well.
Since the $A_{i j}$ and $X_j$'s are independent the variance distributes over the sum.
Additionally we have that
\begin{equation*}
    \Var(A_{i j} X_j)
    =
    \mathbb{E}[X_j]^2 \Var(A_{i j})
    + \mathbb{E}[A_{i j}]^2 \Var(X_j)
    + \Var(A_{i j}) \Var(X_j)
    =
    \Var(A_{i j}) \Var(X_j)
\end{equation*}
since $\mathbb{E}[X_j]=\mathbb{E}[A_{i j}]=0$.
So we work out that
\begin{equation*}
    \Var(Y_i) = \sum_{j=1}^n \Var(A_{i j}) \Var(X_j) = n \Var(A_{i j}) \Var(X_j).
\end{equation*}
Hence for forward signal propagation we have $\Var(Y_i)=\Var(X_j)$ if 
\begin{equation}
\label{eq:var_forward}
\Var(A_{i j})=\frac{1}{n}.
\end{equation}

But we can look at the backward gradient propagation as well.
Let $\ell$ be a loss function at the end of the network, then we can look at the partial derivatives of $\ell$ with respect to the inputs and outputs as random variables as well.
Call these random variables $\frac{\partial \ell}{\partial X_i}$ and $\frac{\partial \ell}{\partial Y_i}$, applying the chain rule gives us
\begin{equation*}
    \frac{\partial \ell}{\partial X_j}
    =
    \sum_{i=1}^m \frac{\partial \ell}{\partial Y_i} \frac{\partial Y_i}{\partial X_j} 
    =
    \sum_{i=1}^m \frac{\partial \ell}{\partial Y_i} A_{i j}
    .
\end{equation*}
Now we make the same assumption about the backward gradients as we did about the forward signals, namely that the partial derivatives $\frac{\partial \ell}{\partial Y_i}$ are i.i.d. with zero mean.
Then we can do the same calculation as before and find
\begin{equation*}
    \Var\left( \frac{\partial \ell}{\partial X_j} \right)
    =
    m \Var\left( A_{i j}  \right) \Var \left( \frac{\partial \ell}{\partial Y_i} \right).
\end{equation*}
Hence if we want to have $\Var\left( \frac{\partial \ell}{\partial X_j} \right)=\Var \left( \frac{\partial \ell}{\partial Y_i} \right)$ for backward gradient propagation we need to set
\begin{equation}
    \label{eq:var_backward}
    \Var(A_{i j}) = \frac{1}{m}.
\end{equation}
Now unless $n=m$ we cannot satisfy \eqref{eq:var_forward} and \eqref{eq:var_backward} at the same time, but we can compromise and set
\begin{equation}
    \label{eq:var_forward_backward}
    \Var(A_{i j}) = \frac{2}{n+m}.
\end{equation}
Under this choice we can use the normal distribution $\mathcal{N}(0,\frac{2}{n+m})$ or the uniform distribution $\rident{Unif}\left[ -\sqrt{\frac{6}{n+m}},  \sqrt{\frac{6}{n+m}} \right]$ to draw our coefficients $A_{i j}$ from.

Of course in reality we never use the linear activation function.
The original Xavier initialization scheme has been expanded to include specific activation functions. 
For example for the ReLU by \textcite{he2015delving}, they arrive at
\begin{equation*}
    \Var(A_{i j}) = \frac{4}{n+m}.
\end{equation*}
Which intuitively makes sense: since the ReLU is zero on half its domain the variance of the coefficients needs to be increased to keep the variances of the signals/gradients constant.

Other choices of activation function lead to other multipliers being introduced to the same basic formula \eqref{eq:var_forward_backward}:
\begin{equation*}
    \Var(A_{i j}) = \alpha^2 \frac{2}{n+m},
\end{equation*}
where $\alpha$ is called the gain and depends on the choice of activation function. 

\begin{remark}
See \href{https://pytorch.org/docs/stable/nn.init.html#torch.nn.init.xavier\_uniform\_}{torch.nn.init.xavier\_uniform\_} and \href{https://pytorch.org/docs/stable/nn.init.html#torch.nn.init.xavier\_normal\_}{torch.nn.init.xavier\_normal\_} for \mbox{PyTorch's} implementation of these initialization schemes.
\end{remark}

\subsection{Those are a lot of Assumptions}

In the last two sections we made a lot of assumptions to arrive at simple formulas.
Some of the assumptions are even verifiably incorrect in the networks we employ.
In spite of the coarse and inelegant way these initialization schemes were derived they are widely used for the simple reason that they work.
They do not totally solve the vanishing/exploding gradient problem but they still significantly improve the performance of the gradient descent algorithms.

%%%%%%%%%%%%%%%%%%%%%%%%%%%%%%%%%%%%%%%%%%%%%%%%%%%%%%%%%%%%%%%%%%%%%%%%%%%%%%%%%
\section{Convolutional Neural Networks}

\(\rightarrow\)  Visualizations of convolutions from \textcite{dumoulin2018guide}, also available at \url{https://github.com/vdumoulin/conv_arithmetic}.

\(\rightarrow\)  Handy reference that provides an overview of CNNs:
\href{https://stanford.edu/~shervine/teaching/cs-230/cheatsheet-convolutional-neural-networks}{CNN cheat-sheet}.

We previously saw that using fully connected networks for high dimensional data such as images is a non-starter due to the memory and computational requirements involved.
The proposed solution was reducing the effective amount of parameters by a combination of using a sparse matrix and weight sharing/parameterization.
How exactly we should sparsify the network and which weights should be shared or parametrized had to be determined  application by application.

In this section we will look at a combination of sparsity and weight sharing that is suited to data that has a natural spatial structure, think about signals in time (1D), images (2D) or volumetric data (3D).
What these types of spatial data have in common is that we treat each `part' of the input data in the same way, e.g. we do not process the left side of an image in another way than the right side.

The type of network that exploits this spacial structure is called a \emph{Convolutional Neural Networks}, or CNN for short. As the name suggest these networks employ the convolution operation as well as the closely related pooling operation. We will look at how discrete convolution and pooling are defined and how they are used to construct a deep CNN.

\subsection{Discrete Convolution}

Recall that in the familiar continuous setting the convolution of two functions $f,g : \mathbb{R} \to \mathbb{R}$ is defined as:
\begin{equation}
    \label{eq:classic_convolution}
    (f * g)(x) := \int_{\mathbb{R}} f(x-y) \, g(y) \, \d y
    ,
\end{equation}
which can be interpreted as a filter (or kernel) $f$ being translated over the data $g$ and at each translated location the $L^2$ inner product is taken.

To switch to the discrete setting we will use notation that is more in line with programming languages.
When we have $f \in \mathbb{R}^n$ we will use square brackets and zero-based indexing to access its components, so $f[0] \in \mathbb{R}$ is $f$'s first component and $f[n-1] \in \mathbb{R}$ its last.
We use this array-based notation since we will need to do some computations on indices and $f[i-j+1]$ is easier to read than $f_{i-j+1}$.

\begin{example}
    Let $x \in \mathbb{R}^n$, $y \in \mathbb{R}^m$ and $A \in \mathbb{R}^{m \times n}$ then the familiar matrix product $y=Ax$ can be written in array notation as
    \begin{equation*}
        y[i] = \sum_{j=0}^{n-1} A[i,j] \, x[j]
    \end{equation*}
    for each $i \in \{ 0,\ldots,m-1 \}$.
\end{example}

\begin{definition}[Discrete cross-correlation and convolution in 1D]
\label{def:discrete_convolution}
Let $f \in \mathbb{R}^n$ (the input) and $k \in \mathbb{R}^m$ (the kernel) then their discrete cross-correlation $(k \star f) \in \mathbb{R}^{n-m+1}$ is given by:
\begin{equation*}
    (k \star f)[i] 
    := 
    \sum_{j=0}^{m-1}
    k[j] \, f[i+j]
\end{equation*}
for $i \in \{ 0, \ldots, n-m \}$.

Discrete convolution is defined similarly but with one of the inputs reversed:
\begin{equation*}
    (k * f)[i] 
    := 
    \sum_{j=0}^{m-1}
    k[m-1-j] \, f[i+j]
\end{equation*}
for $i \in \{ 0, \ldots, n-m \}$.
\end{definition}
Making an index substitution in the definition of discrete convolution makes the relation to the continuous convolution \eqref{eq:classic_convolution} more apparent:
\begin{equation*}
    (k * f)[i]
    =
    \sum_{j=i}^{i+m-1}
    k[i-j + (m-1)]
    \,
    f[j]
    .
\end{equation*}

The idea here is to let the components of the kernel $k \in \mathbb{R}^m$ be trainable parameters.
In that context it does not matter whether the kernel is reflected or not and there is no real reason to distinguish convolution from cross-correlation.
Consequently in the deep learning field it is usual to refer to both operations as convolution.
Most `convolution' operators in deep learning software are in fact implemented as cross-correlations, as in \href{https://pytorch.org/docs/stable/generated/torch.nn.Conv1d.html?highlight=conv1d#torch.nn.Conv1d}{PyTorch} for example.
We will adopt the same convention and talk about such subjects as convolution layers and convolutional neural networks but the actual underlying operation we will use is cross-correlation.

As in the continuous case, discrete convolution can be generalized to higher dimensions.
For this course we will only look at the 2 dimensional case as extending to higher dimensions is straightforward.
To make things slightly simpler we will restrict ourselves to square kernels, in practice kernels are almost always chosen to be square anyway.

\begin{definition}(Discrete cross-correlation in 2D)
    \label{def:discrete_convolution_2D}
    Let $f \in \mathbb{R}^{h \times w}$ (the input, read $h$ as height and $w$ as width) and $k \in \mathbb{R}^{m \times m}$ (the kernel) then their discrete cross-correlation $(k \star f) \in \mathbb{R}^{(h-m+1) \times (w-m+1)}$ is given by
    \begin{equation*}
        (k \star f)[i_1, i_2]
        :=
        \sum_{j_1,j_2=0}^{m-1}
        k[j_1,j_2] \, f[i_1 + j_1, i_2 + j_2]
    \end{equation*}
    for $i_1 \in \{ 0,\ldots, h-m \}$ and $i_2 \in \{ 0, \ldots, w-m \}$.
\end{definition}
The convolution operator ``*'' can be extended to 2 dimensions similarly but we will only be using cross-correlation for the remainder of our discussion of convolutional neural networks.

\begin{figure}[ht!]
\centering
\begin{minipage}[c]{0.65\textwidth}
\centering
    \def\nx{4}
    \def\ny{4}
    \def\padx{0}
    \def\pady{0}
    \def\kx{3}
    \def\ky{3}
    \def\stridex{1}
    \def\stridey{1}
    \def\dilatex{1}
    \def\dilatey{1}
    \def\operator{$k$}
    \def\operatorscale{1.3}
    \def\zoffsetscale{.7}
    \include{tikz/makeconv.tex}
\begin{tikzpicture}[scale=1.0]
    \def\lowerlabel{$f$}
    \def\upperlabel{$k \star f$}
    \begin{scope}
    \makeconv{0}{0}
    \end{scope}
    \def\lowerlabel{}
    \def\upperlabel{}
    
    \begin{scope}[xshift=4.2cm]
    \makeconv{0}{1}
    \end{scope}
    
    \begin{scope}[yshift=-4cm]
    \makeconv{1}{0}
    \end{scope}
    
    \begin{scope}[xshift=4.2cm, yshift=-4cm]
    \makeconv{1}{1}
    \end{scope}
\end{tikzpicture}
\end{minipage}
~\hfill~
\begin{minipage}[c]{0.3\textwidth}
    \captionsetup{width=\linewidth, format=plain}
    \caption{An illustration of convolution (or cross-correlation) in 2D. Here the input $f \in \mathbb{R}^{4 \times 4}$ (colored blue) is convolved with the kernel $k \in \mathbb{R}^{3 \times 3}$, which yields an output $(k \star f) \in \mathbb{R}^{2 \times 2}$ (colored purple).}
    \label{fig:conv2d}
\end{minipage}
\end{figure}
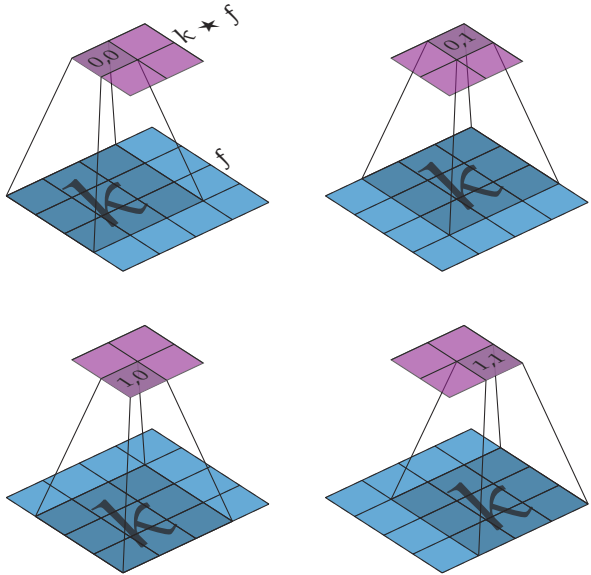

For a visualization of discrete convolution in 2D, see Figure~\ref{fig:conv2d}.

\FloatBarrier
\begin{remark}[Index interpretation conventions]
    \label{remark:indexing_conventions}
    In Definition~\ref{def:discrete_convolution} we called the first dimension of an array $f \in \mathbb{R}^{h \times w}$ the height and the second the width.
    Additionally, if you look at Figure~\ref{fig:conv2d} we place the origin point $(0,0)$ in the top left.
    This is different to the Cartesian convention of denoting points in the plane with $(x,y)$ (i.e. width first, height second) and having the origin in the bottom left.
    Both conventions are illustrated bellow for an element of $\mathbb{R}^{3 \times 4}$.
    \begin{center}
    \begin{tikzpicture}
        \begin{scope}
            \matrix[matrix of math nodes, nodes={fill=RoyalBlue!30,draw=black, minimum size=8mm}]{
                0,3 & 2,3 & 2,3
                \\
                0,2 & 2,2 & 2,2
                \\
                0,1 & 1,1 & 2,1
                \\
                0,0 & 1,0 & 2,0
                \\
            };
            \begin{scope}[xshift=-1.4cm, yshift=-1.75cm]
            \draw[-latex, thick] (0,0) to node[below] {$x$} (1,0);
            \draw[-latex, thick] (0,0) to node[left] {$y$} (0,1);
            \end{scope}
            \node[] at (0cm, 2.1cm) {\emph{Cartesian convention}};
        \end{scope}
        \begin{scope}[xshift=6cm]
            \begin{scope}[yshift=-1mm]
            \matrix[matrix of math nodes, nodes={fill=Plum!30,draw=black, minimum width=8mm, minimum height=8mm}]{
                0,0 & 0,1 & 0,2 & 0,3
                \\
                1,0 & 2,1 & 3,2 & 4,3
                \\
                2,0 & 2,1 & 2,2 & 2,3
                \\
            };
            \begin{scope}[xshift=-1.8cm, yshift=1.35cm]
            \draw[-latex, thick] (0,0) to node[left] {$i_1$} (0,-1);
            \draw[-latex, thick] (0,0) to node[above] {$i_2$} (1,0);
            \end{scope}
            \end{scope}
            \node[] at (0cm, 2.1cm) {\emph{Array convention}};
        \end{scope}
    \end{tikzpicture}
    \end{center}
    Of course choosing either convention does not change the underlying object, merely the interpretation of the indices and shape in colloquial terms.
    The array convention is almost universally adopted in software and it is used in PyTorch, for that reason we will adopt it when dealing with spatial data.
\end{remark}

\subsection{Padding}

The convolution operations we proposed so far have the property that the output is of smaller size than the input.
Depending on our goal this might or might not be desirable.
If shrinking output size is not desirable we can use padding to ensure the output size is the same as the input size.
The most common type of padding is zero padding, which we will look at for the 2 dimensional case.

\begin{definition}[Zero padding in 2D]
    Let $f \in \mathbb{R}^{h \times w}$ and let $p_t,p_b,p_l,p_r$, which we read as top, bottom, left and right padding respectively.
    Then we define $\iident{ZP}_{p_t,p_b,p_l,p_r} f \in \mathbb{R}^{(h+p_t+p_b)\times(w+p_l+p_r)}$ as
    \begin{equation*}
        \iident{ZP}_{p_t,p_b,p_l,p_r} f [i_1,i_2]
        :=
        \begin{cases}
            0 \qquad & \text{if } i_1 < p_t \text{ or } i_1 \geq h + p_t
            \\
            & \text{ or } i_2 < p_l \text{ or } i_2 \geq w + p_l,
            \\
            f[i_1-p_t,\,i_2-p_l]
            & \text{else,}
        \end{cases}
    \end{equation*}
    for all $i_1 \in \{ 0,\ldots,h+p_t+p_b-1 \}$ and $i_2 \in \{0,\ldots,w+p_l+p_r-1 \}$.
\end{definition}    

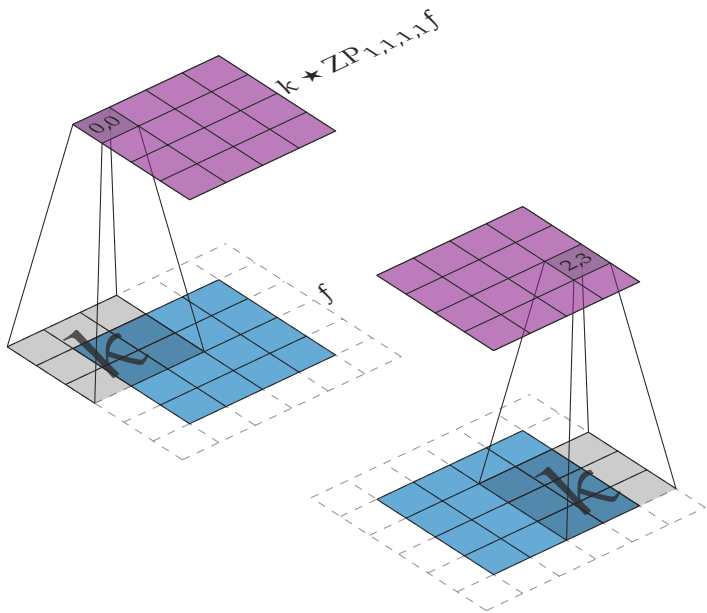
\begin{figure}[ht!]
\centering
\begin{minipage}[c]{0.65\textwidth}
\centering
    \def\nx{4}
    \def\ny{4}
    \def\padx{1}
    \def\pady{1}
    \def\kx{3}
    \def\ky{3}
    \def\stridex{1}
    \def\stridey{1}
    \def\dilatex{1}
    \def\dilatey{1}
    \def\operator{$k$}
    \def\operatorscale{1.3}
    \def\zoffsetscale{.9}
    \include{tikz/makeconv.tex}
\begin{tikzpicture}[scale=1.0]
    \def\lowerlabel{$f$}
    \def\upperlabel{$k \star \iident{ZP}_{1,1,1,1} f$}
    \begin{scope}
    \makeconv{0}{0}
    \end{scope}
    \def\lowerlabel{}
    \def\upperlabel{}
    
    \begin{scope}[xshift=4cm, yshift=-2cm]
    \makeconv{2}{3}
    \end{scope}
\end{tikzpicture}
\end{minipage}
~\hfill~
\begin{minipage}[c]{0.3\textwidth}
    \captionsetup{width=\linewidth, format=plain}
    \caption{Adding padding to the input allows the output of the convolution operation to retain the size of the original input. The most common type of padding is zero-padding, where each out-of-bounds value is assumed to be zero.}
    \label{fig:conv2d_padding}
\end{minipage}
\end{figure}

If we then have an $f \in \mathbb{R}^{h \times w}$ and we want to convolve with a kernel $\mathbb{R}^{m \times m}$ while keeping the shape of the output the same we can choose
\begin{equation*}
    p_t := \left\lfloor \frac{m-1}{2} \right\rfloor
    ,\quad 
    p_b := \left\lceil \frac{m-1}{2} \right\rceil
    ,\quad 
    p_l := \left\lfloor \frac{m-1}{2} \right\rfloor
    ,\quad 
    p_r := \left\lceil \frac{m-1}{2} \right\rceil
    ,
\end{equation*}
then $k \star \iident{ZP}_{p_t,p_b,p_l,p_r} f  \in \mathbb{R}^{h \times w}$. We leave verifying this claim as an exercise. An example of this technique is illustrated in Figure~\ref{fig:conv2d_padding}.

Many more padding techniques exist, they only vary in how the out-of-bounds values are chosen.
In PyTorch the available padding modes are listed in \href{https://pytorch.org/docs/stable/generated/torch.nn.functional.pad.html?highlight=padding}{pytorch.org/docs/stable/generated/torch.nn.functional.pad.html}.

\FloatBarrier
\subsection{Max Pooling}

The second operation commonly found in CNN is max pooling.
We already saw an activation function called max pooling previously, this is exactly what is used in a CNN but with a particular choice of subsets to take maxima over that plays well with the spatial structure of the data.

The idea is to have a `window' slide over the input data in the same way that a convolution kernel slides over the data and then take the maximum value in each window.
We will take a look at a particular type of 2D max pooling that is commonly found in CNNs used for image processing and classification applications.

\begin{definition}[$m \times m$ max pooling in 2D]
    Let $f \in \mathbb{R}^{h \times w}$ and let $m \in \mathbb{N}$.
    Then we define $\iident{MP}_{m,m} f \in \mathbb{R}^{\lfloor\sfrac{h}{m}\rfloor \times \lfloor\sfrac{w}{m}\rfloor}$ as
    \begin{equation*}
        \iident{MP}_{m,m} f [i_1,i_2]
        :=
        \max_{\substack{
            0 \leq j_1 < m
            \\
            0 \leq j_2 < m
            }}
        f[ i_1 m + j_1, i_2 m + j_2]
    \end{equation*}
    for all $i_1 \in \{ 0 ,\ldots, \lfloor\sfrac{h}{m}\rfloor \}$ and $i_2 \in \{ 0 ,\ldots, \lfloor\sfrac{w}{m}\rfloor \}$.
\end{definition}

Two things to note about this particular definition. 
First, if $h$ and/or $w$ is not divisible by $m$ then some values at the edges will be ignored entirely and not contribute to the output.
We could modify the definition to resolve this but in practice $m$ is usually set to $2$ and so in the worst case we lose a single row of values at the edge, and that is only if the input has odd dimensions.

\begin{figure}[ht!]
\centering
\begin{minipage}[c]{0.65\textwidth}
\centering
    \def\nx{4}
    \def\ny{4}
    \def\padx{0}
    \def\pady{0}
    \def\kx{2}
    \def\ky{2}
    \def\stridex{2}
    \def\stridey{2}
    \def\dilatex{1}
    \def\dilatey{1}
    \def\operator{$\max$}
    \def\operatorscale{0.7}
    \def\zoffsetscale{.9}
    \include{tikz/makeconv.tex}
\begin{tikzpicture}[scale=1.0]
    \def\lowerlabel{$f$}
    \def\upperlabel{$\iident{MP}_{2,2} f$}
    \begin{scope}
    \makeconv{0}{0}
    \end{scope}
    \def\lowerlabel{}
    \def\upperlabel{}
    
    \begin{scope}[xshift=4.2cm]
    \makeconv{0}{1}
    \end{scope}
    
    \begin{scope}[yshift=-4cm]
    \makeconv{1}{0}
    \end{scope}
    
    \begin{scope}[xshift=4.2cm, yshift=-4cm]
    \makeconv{1}{1}
    \end{scope}
\end{tikzpicture}
\end{minipage}
~\hfill~
\begin{minipage}[c]{0.3\textwidth}
    \captionsetup{width=\linewidth, format=plain}
    \caption{Max pooling is similar to convolution in that a window of a certain size slides over the input, instead of taking a weighted sum inside the window we take the maximum value. Usually the window is moved in strides equal to its size so that the outputs are the maxima from disjoint sets of the input. The pooling operation in the figure is usually just called $2 \times 2$ max pooling, referring to both the window size and stride used.}
    \label{fig:maxpool2x2}
\end{minipage}
\end{figure}
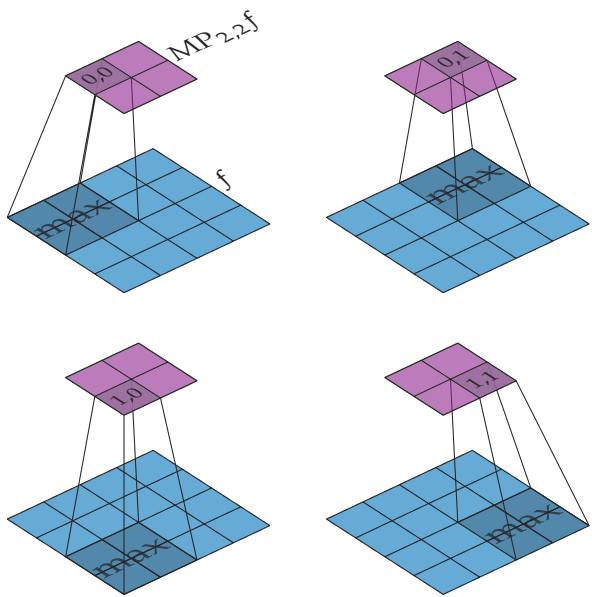

Second, we move the window in steps of $m$ in both directions instead of taking steps of $1$, this is called having a \emph{stride} of $m$.
Having a stride larger than $1$ allows max pooling to quickly reduce the dimensions of the data.
Figure~\ref{fig:maxpool2x2} shows an example of how $\iident{MP}_{2,2}$ works.

\FloatBarrier
\subsection{Convolutional Layers}

So how are convolutions/cross-correlations used in a neural network?
First of all we organize our inputs and outputs differently, instead of the inputs and outputs being element of $\mathbb{R}^n$ for some $n$ we want to keep the spatial structure.
In the 2D case it makes sense to talk about objects with height and width, so elements in $\mathbb{R}^{h \times w}$ for some choice of $h,w \in \mathbb{N}$.
We will call these objects 2D maps or just \emph{maps}.
We will want more than one map at any stage of the network, we say we want multiple \emph{channels}.

Concretely, say we have a deep network where we index the number of layers with $i$, then we denote the input of layer $i$ as an element in $\mathbb{R}^{C_{i-1} \times h_{i-1} \times w_{i-1}}$ and the output of layer $i$ as an element of $\mathbb{R}^{C_{i} \times h_{i} \times w_{i}}$.
We say layer $i$ has $C_{i-1}$ \emph{input channels} and $C_{i}$ \emph{output channels}.
We call $h_{i-1}\times w_{i-1}$ and $h_i \times w_i$ the shape of the input respectively output maps.

\begin{example}[Image inputs]
    If we design a neural network to process color 1080p images then $C_0=3$ (RGB images have 3 color channels), $h_0=1080$ and $w_0=1920$, i.e. the input to the first layer is an element in $\mathbb{R}^{3 \times 1080 \times 1920}$.
    For a monochrome image we would need just 1 channel.
\end{example}

A \emph{convolution layer} is just a specialized network layer, hence it follows the same pattern of first doing a linear transform, then adding a bias and finally applying an activation function.
The activation function in a CNN is typically a ReLU or max pooling function (or both).
The linear part will consist of taking convolutions of the input maps, but there are several ways to do that, we will look at two of them.

The first, and most straightforward one, is called \emph{single channel convolution} (alternatively know as \emph{depthwise convolution}).
With this method we assign a kernel of a certain size to each input channel and then perform the convolution of each input channel with each kernel.
This gives us a number of maps equal to the number of input channels, we subsequently take point-wise linear combinations of those maps to generate the desired number of output maps.

\begin{definition}[Single channel convolution]
    \label{def:singlechannel_conv}
    Let $f \in \mathbb{R}^{C \times h \times w}$ and let $k \in \mathbb{R}^{C \times m \times m}$.
    We call $C$ the number of input channels, $h \times w$ the input map shape and $m \times m$ the kernel shape.
    Let $A \in \mathbb{R}^{C' \times C}$, we call $C'$ the number of output channels.
    Then we define $SCC_{k,A}(f) \in \mathbb{R}^{C' \times (h-m+1) \times (w-m+1)}$ as:
    \begin{equation*}
        \iident{SCC}_{k,A}(f)[c',i_1,i_2]
        :=
        \sum_{c=0}^{C-1}
        A[c',c]
        \ 
        k[c,\cdot,\cdot] \star f[c,\cdot,\cdot]
        ,
    \end{equation*}
    for all $c' \in \{ 0, \ldots, C'-1 \}$, $i_1 \in \{ 0,\ldots,h-m+1 \}$ and $i_2 \in \{ 0,\ldots, w-m+1 \}$.
\end{definition}

\begin{figure}[ht!]
\centering
\begin{tikzpicture}

\newcommand{\makegrid}[3]{
	\coordinate (FROM) at (0,0);
	\coordinate (TO) at (#1,#2);

	\path[fill=#3, opacity=0.85] (FROM) rectangle (TO);
	\draw[step=10mm, opacity=0.85] (FROM) grid (TO);
}

\begin{scope}[
scale=.6
, on grid
,z = {(0,1)}
,x = {(0.8,0.38)}
,y={(-0.8,0.72)}
]

\begin{scope}[xshift=6cm]
	% kernel stack 
	\begin{scope}[canvas is xy plane at z=0.65, transform shape]
		\makegrid{2}{2}{PineGreen!70}
	\end{scope}

	\begin{scope}[canvas is xy plane at z=1.65, transform shape]
		\makegrid{2}{2}{Plum!70}
	\end{scope}

	\begin{scope}[canvas is xy plane at z=2.65, transform shape]
		\makegrid{2}{2}{RoyalBlue!70}
	\end{scope}
	
	\node[] at (0,0,6) {\textsf{kernel stack}};
	\node[] at (0,0,-1) {$\mathbb{R}^{3 \times 2 \times 2}$};
\end{scope}

\begin{scope}[xshift=-0.5cm]
	\node[] (in1) at (4,0,0) {};
	\node[] (in2) at (4,0,1) {};
	\node[] (in3) at (4,0,2) {};
\end{scope}

\begin{scope}[xshift=6cm]
	\node[] (kernel1) at (0,2,0.65) {};
	\node[] (kernel2) at (0,2,1.65) {};
	\node[] (kernel3) at (0,2,2.65) {};

	\node[] (kernel1out) at (2,0,0.65) {};
	\node[] (kernel2out) at (2,0,1.65) {};
	\node[] (kernel3out) at (2,0,2.65) {};
\end{scope}

\draw[latex-latex] (in1) to[bend left=10] node[scale=0.6,above] {$\star$} (kernel1);
\draw[latex-latex] (in2) to[bend left=10] node[scale=0.6,above] {$\star$} (kernel2);
\draw[latex-latex] (in3) to[bend left=10] node[scale=0.6,above] {$\star$} (kernel3);

% input stack
\begin{scope}[xshift=-0.5cm]
	\begin{scope}[canvas is xy plane at z=0, transform shape]
		\makegrid{4}{4}{Gray}
	\end{scope}
	
	\begin{scope}[canvas is xy plane at z=1, transform shape]
		\makegrid{4}{4}{Gray!70}
	\end{scope}
	
	\begin{scope}[canvas is xy plane at z=2, transform shape]
		\makegrid{4}{4}{Gray!40}
	\end{scope}

	\node[] at (0,0,6) {\textsf{input}};
	\node[] at (0 ,0,-1) {$\mathbb{R}^{3 \times 4 \times 4}$};
\end{scope}

\def\M{
$
\begin{pmatrix}
	0 & 1 & 1
	\\
	1 & 0 & 2
\end{pmatrix}
$
}

\begin{scope}[xshift=11cm]
	\node[rectangle, scale=0.9] (M) at (0,0,2.4) {\M};
	\node[] at (0,0,6) {\textsf{linear}};
	\node[] at (0,0,5.3) {\textsf{combination}};
	\node[] at (0 ,0,-1) {$\mathbb{R}^{2 \times 3}$};
\end{scope}

\draw[-latex] (kernel1out) to[bend right=10] (M.200);
\draw[-latex] (kernel2out) to[] (M.180);
\draw[-latex] (kernel3out) to[bend left=10] (M.160);

% output stack
\begin{scope}[xshift=16cm]
	\begin{scope}[canvas is xy plane at z=0.5, transform shape]
		\makegrid{3}{3}{PineGreen!65!RoyalBlue}
	\end{scope}

	\begin{scope}[canvas is xy plane at z=1.5, transform shape]
		\makegrid{3}{3}{Plum!50!RoyalBlue}
	\end{scope}

	\node[] (out2) at (0,1.5,0.5) {};
	\node[] (out1) at (1.5,3,1.5) {};

	\draw[-latex] (M.20) to[bend left=25] (out1);
	\draw[-latex] (M.-20) to[bend right=25] (out2);
	
	\node[] at (0,0,6) {\textsf{output}};
	\node[] at (0,0,-1) {$\mathbb{R}^{2 \times 3 \times 3}$};
\end{scope}

\end{scope}
\end{tikzpicture}
\caption{With \emph{single channel convolution}, also called \emph{depthwise convolution}, each input channel gets assigned a single kernel. After doing $C_{in}$ cross-correlations we take pointwise linear combinations of the resulting maps to generate the desired number of output maps. In this example that yields a total of 18 trainable parameters (or 20 if we include a bias per output channel).}
\label{fig:singlechannel_conv}
\end{figure}
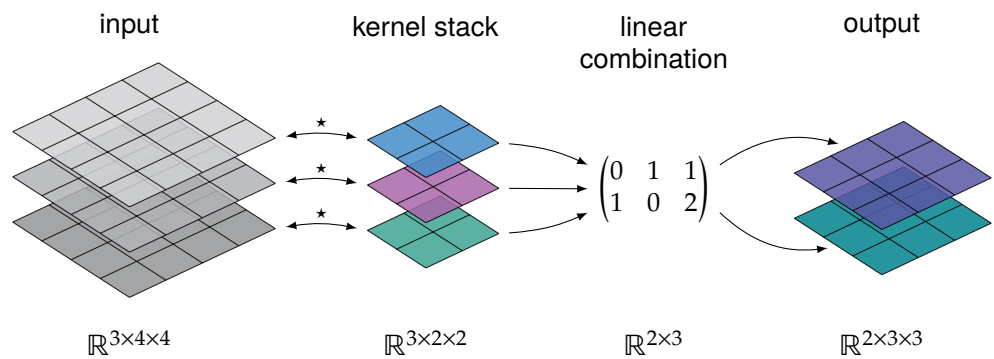

This operation allows the entries of the kernel stack $k$ and matrix $A$ to be trainable, hence the number of trainable parameters is $C\cdot m^2+C' \cdot C$.
Figure~\ref{fig:singlechannel_conv} gives a visualization of how single channel convolution works.

An alternative way of using convolution to build a linear operator is \emph{multi channel convolution} (alternatively known as \emph{multi channel multi kernel} (MCMK) convolution).
Instead of assigning a kernel to each input channel and then taking linear combinations we assign a kernel to each combination of input and output channels.

\begin{definition}[Multi channel convolution]
    \label{def:multichannel_conv}
    Let $f \in \mathbb{R}^{C \times h \times w}$ and let $k \in \mathbb{R}^{C' \times C \times m \times m}$.
    We call $C$ the number of input channels, $C'$ the number of output channels, $h \times w$ the input map shape and $m \times m$ the kernel shape.
    Then we define $MCC_{k}(f) \in \mathbb{R}^{C' \times (h-m+1) \times (w-m+1)}$ as:
    \begin{equation*}
        \iident{MCC}_{k}(f)[c',i_1,i_2]
        :=
        \sum_{c=0}^{C-1}
        k[c',c,\cdot,\cdot] \star f[c,\cdot,\cdot]
        ,
    \end{equation*}
    for all $c' \in \{ 0, \ldots, C'-1 \}$, $i_1 \in \{ 0,\ldots,h-m+1 \}$ and $i_2 \in \{ 0,\ldots,w-m+1 \}$.
\end{definition}

\begin{figure}[ht!]
\centering
\begin{tikzpicture}

\newcommand{\makegrid}[3]{
	\coordinate (FROM) at (0,0);
	\coordinate (TO) at (#1,#2);

	\path[fill=#3, opacity=0.85] (FROM) rectangle (TO);
	\draw[step=10mm, opacity=0.85] (FROM) grid (TO);
}

\begin{scope}[
scale=.6
, on grid
,z = {(0,1)}
,x = {(0.8,0.38)}
,y={(-0.8,0.72)}
]

\begin{scope}[xshift=6cm]
	% kernel stack 1
	\begin{scope}[canvas is xy plane at z=2.8, transform shape]
		\makegrid{2}{2}{RoyalBlue}
	\end{scope}
	
	\begin{scope}[canvas is xy plane at z=3.3, transform shape]
		\makegrid{2}{2}{RoyalBlue!70}
	\end{scope}

	\begin{scope}[canvas is xy plane at z=3.8, transform shape]
		\makegrid{2}{2}{RoyalBlue!40}
	\end{scope}

	% kernel stack 2
	\begin{scope}[canvas is xy plane at z=-0.3, transform shape]
		\makegrid{2}{2}{Plum}
	\end{scope}
	
	\begin{scope}[canvas is xy plane at z=0.2, transform shape]
		\makegrid{2}{2}{Plum!70}
	\end{scope}

	\begin{scope}[canvas is xy plane at z=0.7, transform shape]
		\makegrid{2}{2}{Plum!40}
	\end{scope}
	
	\node[] at (0,0,6) {\textsf{kernel stack}};
	\node[] at (0,0,-1) {$\mathbb{R}^{2 \times 3 \times 2 \times 2}$};
\end{scope}

\begin{scope}[xshift=-0.5cm]
	\node[] (in1a) at (4,2,0) {};
	\node[] (in1b) at (4,2,1) {};
	\node[] (in1c) at (4,2,2) {};

	\node[] (in2a) at (2,0,0) {};
	\node[] (in2b) at (2,0,1) {};
	\node[] (in2c) at (2,0,2) {};
\end{scope}

\begin{scope}[xshift=6cm]
	\node[] (kernel1a) at (0,2,2.8) {};
	\node[] (kernel1b) at (0,2,3.3) {};
	\node[] (kernel1c) at (0,2,3.8) {};

	\node[] (kernel2a) at (0,2,-0.3) {};
	\node[] (kernel2b) at (0,2,0.2) {};
	\node[] (kernel2c) at (0,2,0.7) {};

	\node[] (kernel3a) at (2,0,2.8) {};
	\node[] (kernel3b) at (2,0,3.3) {};
	\node[] (kernel3c) at (2,0,3.8) {};

	\node[] (kernel4a) at (2,0,-0.3) {};
	\node[] (kernel4b) at (2,0,0.2) {};
	\node[] (kernel4c) at (2,0,0.7) {};
\end{scope}

\begin{scope}[xshift=12cm]

\end{scope}

\draw[latex-latex] (in1a) to[bend left=20] node[scale=0.6,above] {$\star$} (kernel1a);
\draw[latex-latex] (in1b) to[bend left=20] node[scale=0.6,above] {$\star$} (kernel1b);
\draw[latex-latex] (in1c) to[bend left=20] node[scale=0.6,above] {$\star$} (kernel1c);

% input stack
\begin{scope}[xshift=-0.5cm]
	\begin{scope}[canvas is xy plane at z=0, transform shape]
		\makegrid{4}{4}{Gray}
	\end{scope}
	
	\begin{scope}[canvas is xy plane at z=1, transform shape]
		\makegrid{4}{4}{Gray!70}
	\end{scope}
	
	\begin{scope}[canvas is xy plane at z=2, transform shape]
		\makegrid{4}{4}{Gray!40}
	\end{scope}

	\node[] at (0,0,6) {\textsf{input}};
	\node[] at (0 ,0,-1) {$\mathbb{R}^{3 \times 4 \times 4}$};
\end{scope}

\draw[latex-latex] (in2a) to[bend left=20] node[scale=0.6,above] {$\star$} (kernel2a);
\draw[latex-latex] (in2b) to[bend left=20] node[scale=0.6,above] {$\star$} (kernel2b);
\draw[latex-latex] (in2c) to[bend left=20] node[scale=0.6,above] {$\star$} (kernel2c);

\begin{scope}[xshift=11cm]
	\node[] at (0,0,6) {\textsf{sum}};
	\node[circle, inner sep=1pt] (sum1) at (0,0,4) {$\sum$};
	\node[circle, inner sep=1pt] (sum2) at (0,0,1) {$\sum$};
\end{scope}

% output stack
\begin{scope}[xshift=16cm]
	\begin{scope}[canvas is xy plane at z=0.5, transform shape]
		\makegrid{3}{3}{Plum!70}
	\end{scope}

	\begin{scope}[canvas is xy plane at z=1.5, transform shape]
		\makegrid{3}{3}{RoyalBlue!70}
	\end{scope}

	\draw[-latex] (kernel3a) to[bend right=5] (sum1.200);
	\draw[-latex] (kernel3b) to[] (sum1.180);
	\draw[-latex] (kernel3c) to[bend left=5] (sum1.160);

	\draw[-latex] (kernel4a) to[bend right=5] (sum2.200);
	\draw[-latex] (kernel4b) to[] (sum2.180);
	\draw[-latex] (kernel4c) to[bend left=5] (sum2.160);

	\node[] (out2) at (0,1.5,0.5) {};
	\node[] (out1) at (1.5,3,1.5) {};

	\draw[-latex] (sum1) to[bend left=10] (out1);
	\draw[-latex] (sum2) to[bend right=10] (out2);
	
	\node[] at (0,0,6) {\textsf{output}};
	\node[] at (0,0,-1) {$\mathbb{R}^{2 \times 3 \times 3}$};
\end{scope}

\end{scope}
\end{tikzpicture}
\caption{With \emph{multi channel convolution} each output channel gets assigned a stack of kernels, where each stack has a kernel per input channel. This results in $C_{out} \cdot C_{in}$ kernels. The resulting outputs of the $C_{out} \cdot C_{in}$ cross-correlations are then summed up pointwise per output channel resulting in $C_{out}$ output channels. In this example that yields a total of 24 trainable parameters (or 26 if we include a bias per output channel).}
\label{fig:multichannel_conv}
\end{figure}
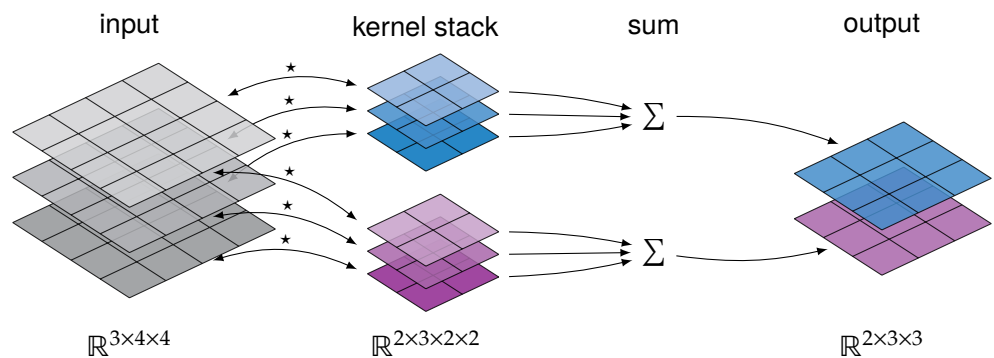

Under this construction the kernel components are the trainable parameters and so we have a total of $C'\cdot C \cdot m^2$ trainable parameters.
The multi channel convolution construction is illustrated in Figure~\ref{fig:multichannel_conv}.

Neural network frameworks like \href{https://pytorch.org/docs/stable/generated/torch.nn.Conv2d.html?highlight=conv2d#torch.nn.Conv2d}{\mbox{PyTorch}} implement the multi channel version. CNN's in literature are also generally formulated with multi channel convolutions.
So why did we also introduce single channel convolutions?

First of all, single and multi channel convolution are equivalent in the sense that given an instance of one I can always construct an instance of the second that does the exact same calculation.
Consequently we can work with whichever construction we prefer for whatever reason without losing anything.
The nice thing about single channel convolution is that there is a clear separation between the processing done inside a particular channel and the way the input channels are combined to create output channels.
These two processing steps are fused together in the multi channel convolution operation. 
In the author's opinion this makes the multi channel technique harder to reason about, having two distinct steps that do two distinct things seems more elegant.

All that is left now to create a full CNN layer is combining one of the convolution operations, add padding if desirable and pass the result through a max pooling operation and/or a scalar activation function such as a ReLU.

\FloatBarrier
\subsection{Classification Example: MNIST \& LeNet-5}

\begin{figure}[ht!]
\centering
\begin{minipage}[c]{0.65\textwidth}
\centering
\includegraphics[width=\linewidth]{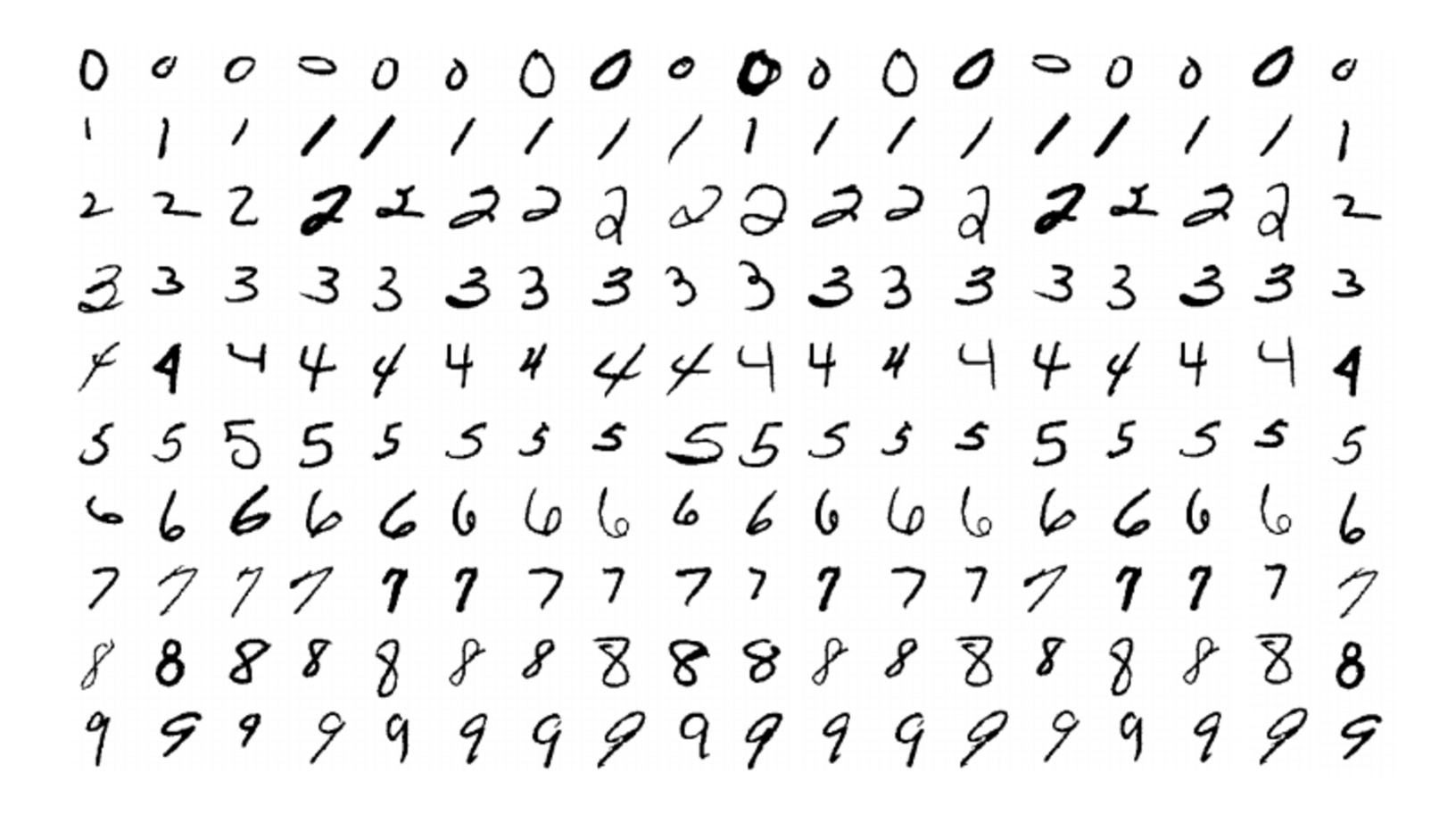}
\end{minipage}
~\hfill~
\begin{minipage}[c]{0.3\textwidth}
    \captionsetup{width=\linewidth, format=plain}
    \caption{The MNIST dataset (Modified National Institute of Standards and Technology dataset) consists of a large collection of $28\times 28$ grayscale images of hand drawn digits. The goal is assigning to each image the correct 0-9 label.}
    \label{fig:mnist}
\end{minipage}
\end{figure}

\begin{figure}[ht!]
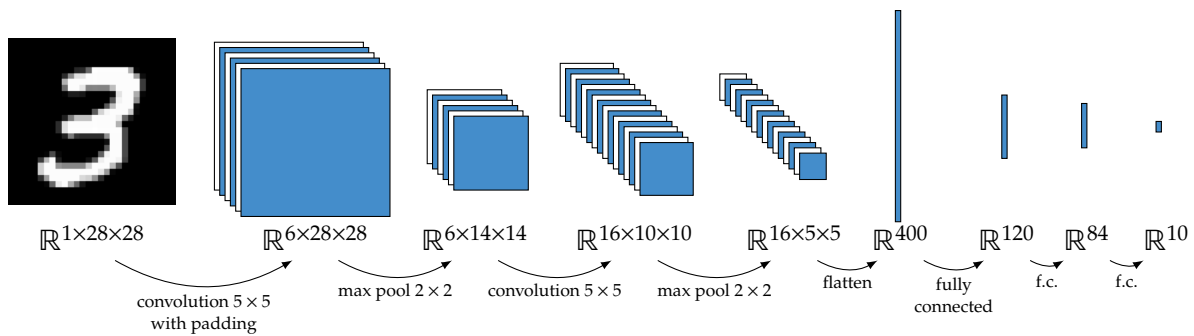

    \centering
    \include{tikz/lenet5.tex}
    \vspace{-1cm}
    \caption{A modernized LeNet-5 network, the classic LeNet-5 network from \textcite{lecun1998gradient} used sigmoidal activation functions, unusual sub-sampling layers and Gaussian connections. Replacing these with ReLUs, max pooling and fully connected layers yields a somewhat simpler network that works equally well.}
    \label{fig:letnet5}
\end{figure}

%%%%%%%%%%%%%%%%%%%%%%%%%%%%%%%%%%%%%%%%%%%%%%%%%%%%%%%%%%%%%%%%%%%%%%%%%%%%%%%%%
\section{Automatic Differentiation \& Backpropagation}

All our training algorithms are gradient based and so our ability to train a network rests on us computing those gradients.
The traditional approaches we could take are:
\begin{enumerate}[label=(\arabic*)]
    \item compute the derivative formulas ourselves and code them into the program,
    \item use a computer algebra system to perform symbolic differentiation and use the resulting formula,
    \item use numeric differentiation (i.e. finite differences).
\end{enumerate}
None of these three options are appealing. We definitely do not want to work out the derivatives ourselves for every network we define so option (1) is out. Symbolic differentiation with computer algebra systems like Mathematica usually produces complex and cryptic expressions in non-trivial cases. Subsequently evaluating these complex expressions is usually anything but efficient, this rules out option (2). For networks with millions if not billions of parameters computing finite differences would entail millions or billions of evaluations of the network, this makes option (3) impossible to use.

What we need is a fourth technique: \emph{automatic differentiation}, also called \emph{algorithmic differentiation} or simply \emph{autodiff}.
We will restrict ourselves to looking at automatic differentiation in the context of feed-forward neural networks but it is a very general technique, see \textcite{baydin2018automatic} for a broader survey of its uses in machine learning.
In introducing automatic differentiation it is perhaps best to emphasizing that it is \emph{not} symbolic differentiation nor numeric differentiation, even though it contains elements of both.

\begin{remark}[Autograd vs. autodiff]
    Autograd is a particular autodiff implementation in Python. 
    Autograd served as a prototype for a lot of autodiff implementations, including PyTorch's, for that reason in PyTorch the term \emph{autograd} is used as a synonym for autodiff.
\end{remark}

\subsection{Notation and an Example}

Let us say we have a network $F:\mathbb{R}\times \mathbb{R}^2 \to \mathbb{R}$ with two parameters $a$ and $b$ and loss function $\ell$.
We are interested in computing:
\begin{equation}
    \label{eq:a_b_derivatives}
    \left. \frac{\partial}{\partial a}\ell(F(x_0;a,b),y_0) \right\vert_{(a,b)}
    \quad
    \text{and}
    \quad
    \left. \frac{\partial}{\partial b}\ell(F(x_0;a,b),y_0) \right\vert_{(a,b)}
    ,
\end{equation}
i.e. the partial derivatives with respect to out parameters at their current values which are just real numbers.
Note how we use $a$ and $b$ first as dummy variables to indicate the input we want to differentiate over and then again as real valued function arguments.

Partial derivative notation can get cluttered quickly, as \eqref{eq:a_b_derivatives} shows.
To simplify things we are going to introduce some notational conventions.
Let $f:\mathbb{R}^2 \to \mathbb{R}$ be given by
\begin{equation}
    \label{eq:fghk}
    f(x,y) = g(h(x,y),k(x,y))
    ,
\end{equation}
where $g,h$ and $k$ are also functions from $\mathbb{R}^2$ to $\mathbb{R}$.
We evaluate $f$ at a fixed $(x,y)$ by calculating:
\begin{equation*}
    u = h(x,y)
    ,
    \quad
    v = k(x,y)
    ,
    \quad
    z = g(u,v) = f(x,y)
    ,
\end{equation*}
which are all just real numbers and not functions like $f,g,h$ and $k$ are.
Now we introduce the following notations for the intermediate results $u$ and $v$:
\begin{equation*}
    \begin{split}
    \frac{\partial u}{\partial x}
    &:=
    \left.
    \frac{\partial}{\partial x}
    h(x,y)
    \right|_{(x,y)}
    ,
    \qquad
     \frac{\partial u}{\partial y}
    :=
    \left.
    \frac{\partial}{\partial y}
    h(x,y)
    \right|_{(x,y)}
    ,
    \\
    \frac{\partial v}{\partial x}
    &:=
    \left.
    \frac{\partial}{\partial x}
    k(x,y)
    \right|_{(x,y)}
    ,
    \qquad
    \frac{\partial v}{\partial y}
    :=
    \left.
    \frac{\partial}{\partial y}
    k(x,y)
    \right|_{(x,y)}
    ,
    \end{split}
\end{equation*}
and for the final output $z$ we define:
\begin{equation*}
    \begin{split}
    \frac{\partial z}{\partial u}
    &:=
    \left.
    \frac{\partial}{\partial u} g(u,v)
    \right|_{(u,v)}
    ,
    \qquad
    \frac{\partial z}{\partial v}
    :=
    \left.
    \frac{\partial}{\partial v} g(u,v)
    \right|_{(u,v)}
    \\
    \frac{\partial z}{\partial x}
    &:=
    \left.
    \frac{\partial}{\partial x} f(x,y)
    \right|_{(x,y)}
    ,
    \qquad
    \frac{\partial z}{\partial y}
    :=
    \left.
    \frac{\partial}{\partial y} f(x,y)
    \right|_{(x,y)}
    ,
    \end{split}
\end{equation*}
which are all real numbers.
At first glance the partial derivative of one real number with respect to another real number is nonsensical,
but if we use one real number in the computation of a second one it does make sense to ask how sensitive the value of the second is with respect to the first if we changed it a little bit.
This sensitivity is of course nothing but the partial derivative of the function used in the computation of the second value evaluated at the first value.
But since we are not interested in the whole partial derivatives it simplifies things to keep the function and its partial derivatives implicit and use the simpler notation we just introduced.

Using this notation the chain rule applied to \eqref{eq:fghk} can be expressed very succinctly as
\begin{equation}
    \label{eq:numeric_chain_rule}
    \frac{\partial z}{\partial x}
    =
    \frac{\partial z}{\partial u} \frac{\partial u}{\partial x}
    +
    \frac{\partial z}{\partial v} \frac{\partial v}{\partial x}
    \quad
    \text{and}
    \quad
    \frac{\partial z}{\partial y}
    =
    \frac{\partial z}{\partial u} \frac{\partial u}{\partial y}
    +
    \frac{\partial z}{\partial v} \frac{\partial v}{\partial y}   
    .
\end{equation}

We are always looking to minimize our chosen loss function, hence we are mainly interested in the partial derivatives of the loss function.
We will use the following notation to further abbreviate the partial derivatives of the loss: let $\ell \in \mathbb{R}$ be the loss produced at the end of our calculation and let $\alpha \in \mathbb{R}$ be either a parameter or intermediate result used in the calculation of $\ell$ then we write
\begin{equation*}
    \overline{\alpha} := \frac{\partial \ell}{\partial \alpha}
    ,
\end{equation*}
which we refer to as a \emph{gradient} (technically the value of the gradient of the loss function with respect to $\alpha$ evaluated at the current value of $\alpha$, but that is quite the mouthful).
So in the case of the two-parameter network $F$ from before: for a given set of parameters $a,b \in \mathbb{R}$ we need to calculate the gradients $\overline{a},\overline{b} \in \mathbb{R}$, which are exactly the evaluated partial derivatives from \eqref{eq:a_b_derivatives}.

As an example to see how this notation works let us pick a concrete example and systematically work through differentiating it.
Let $F(x;a,b):=\sigma(a x + b)$ for some choice of (differentiable) activation function $\sigma:\mathbb{R} \to \mathbb{R}$.
Let $(x_0,y_0) \in \mathbb{R}^2$ be some data point and let us use the loss function $(y,y') \mapsto \frac{1}{2}(y-y')^2$.
After we pick our parameter values $a,b \in \mathbb{R}$ we can simply compute the corresponding loss, doing this is called the \emph{forward pass} and is shown on the left side in \eqref{eq:ad_example_1} below.
\begin{equation}
\label{eq:ad_example_1}
\begin{tikzpicture}[baseline]
\tikzstyle{every node}=[
rounded corners=1mm
, anchor=west
]
\matrix (m) [
matrix of math nodes
, column sep = 16pt
, row sep = 4pt
, nodes in empty cells
, ampersand replacement=\&
] 
{
\&
a,b,x_0,y_0 \in \mathbb{R}
\&
\&
\overline{a} = \frac{\partial \ell}{\partial z} \frac{\partial z}{\partial a} = \overline{z} \, x_0 \, , \ \ \overline{b}=\overline{z}
\&
\\
\&
\tikz[baseline]{
        \node[fill=LimeGreen!20,anchor=base] (1)
        {$z$};}=a x_0 + b
\&
\&
\overline{z}=\frac{\partial \ell}{\partial z}=\frac{\partial \ell}{\partial y} \frac{\partial y}{\partial z}=\overline{y}\, \sigma'(\tikz[baseline]{
        \node[fill=LimeGreen!20,anchor=base] (2)
        {$z$};})
\&
\\
\&
y=\sigma(z)
\&
\&
\overline{y}
=
\frac{\partial\ell}{\partial y}
=
\tikz[baseline]{
\node[fill=PineGreen!20,anchor=base] (3){$y-y_0$};}
\&
\\
\&
\ell = \frac{1}{2} (\tikz[baseline]{
        \node[fill=PineGreen!20,anchor=base] (4)
        {$y-y_0$};})^2
\&
\&
\overline{\ell}=\frac{\partial\ell}{\partial\ell}=1
\&
\\
};
\path[-latex, very thick, color=RoyalBlue!100]
(m-1-1) edge node [sloped, anchor=center, above, rotate=180] {\textsf{\small forward}} (m-4-1);
\path[-latex, very thick, color=Plum!100]
(m-4-5) edge node [sloped, anchor=center, above, rotate=180] {\textsf{\small backward}} (m-1-5);
\begin{scope}[node distance=1mm]
    \draw[black,-] 
    coordinate[above=of m-1-3] (a)  
    coordinate[below=of m-4-3] (b)
    (a) -- (b);
\end{scope}
\end{tikzpicture}
\end{equation}
After having evaluated the network we can start calculating the partial derivatives of the loss with respect the the parameters $a$ and $b$ by apply the chain rule from back to front, this is called the \emph{backward pass} and is shown on the right side in \eqref{eq:ad_example_1}.
Some thing to note about the schematic in \eqref{eq:ad_example_1}.
\begin{itemize}
    \item 
    Everything that we wrote down is a numeric computation and the whole schematic can be executed by a computer as is.
    
    \item 
    In writing down the backward pass we did use our symbolic knowledge of how the operations in the forward pass need to be differentiated if we look at them as functions.
    
    \item 
    Writing down the trivial $\overline{\ell}=1$ is of course redundant, but autodiff implementations such as PyTorch's do actually start the backward pass by creating a single element tensor containing the value $1$ (and it makes the schematic look symmetric). 
    
    \item 
    Some of the intermediate values computed during the forward pass are reused during the backward pass. 
    In the schematic \eqref{eq:ad_example_1} the reused values have been shaded in green.
\end{itemize}

\begin{remark}
    In PyTorch, the intermediate results computed during the forward pass that need to be retained for the backward pass are called \emph{saved tensors}.
    In a typical neural network many of the intermediate results have to be saved for the backward pass, this is the reason that training a neural network requires a lot of memory.
\end{remark}

Nothing we have done in the example in \eqref{eq:ad_example_1} is novel, we just did what we would normally do if asked to calculate these partial derivatives.
Only, we wrote it down systematically in a way that we can automate.

\subsection{Automation \& the Computational Graph}

The key to automating the type of calculation in \eqref{eq:ad_example_1} and \eqref{eq:ad_example_2} is splitting it into primitive operations and tracking how they are composed. Consider the network $F(x;a,b):=\relu(a x + b)$ with loss function $(y,y')\mapsto \frac{1}{2}(y-y')^2$ evaluated for some data point $(x_0,y_0) \in \mathbb{R}^2$ and parameter values $a,b \in \mathbb{R}$. Evaluating the network one primitive operation at a time looks as follows.

\begin{equation}
\label{eq:ad_example_2}
\begin{tikzpicture}[baseline]
\tikzstyle{every node}=[
rounded corners=1mm
, anchor=west
]
\matrix (m) [
matrix of math nodes
, column sep = 16pt
, row sep = 4pt
, nodes in empty cells
, ampersand replacement=\&
] 
{
\&
\text{data: } x_0, y_0 \in \mathbb{R}
\&
\&
\&
\\
\&
\text{parameters: } a,b \in \mathbb{R}
\&
\&
\overline{a} 
= \frac{\partial \ell}{\partial t_1} \frac{\partial t_1}{\partial a}
= \overline{t_1} \, x_0, 
\ \ 
\overline{b}
= \frac{\partial \ell}{\partial t_2} \frac{\partial t_2}{\partial b}
= \overline{t_2}
\&
\\
\&
t_1 = a x_0
\&
\&
\overline{t_1} 
= \frac{\partial \ell}{\partial t_2} \frac{\partial t_2}{\partial t_1}
= \overline{t_2}
\&
\\
\&
\tikz[baseline]{\node[fill=LimeGreen!20,anchor=base] (1){$t_2$};} 
= t_1 + b
\&
\&
\overline{t_2} 
= \frac{\partial \ell}{\partial t_3} \frac{\partial t_3}{\partial t_2}
= \overline{t_3} \, \mathbb{1}_{\tikz[baseline]{\node[fill=LimeGreen!20,anchor=base] (1){$t_2$};} \geq 0}
\&
\\
\&
t_3 = \relu(t_2)
\&
\&
\overline{t_3} 
= \frac{\partial \ell}{\partial t_4} \frac{\partial t_4}{t_3}
= \overline{t_4}
\&
\\
\&
\tikz[baseline]{\node[fill=PineGreen!20,anchor=base] (1){$t_4$};} = t_3-y_0
\&
\&
\overline{t_4} 
= \frac{\partial \ell}{\partial t_4}
= \frac{\partial \ell}{\partial t_5} \frac{\partial t_5}{\partial t_4}
= \overline{t_5} \cdot 2 t_4 
= \tikz[baseline]{\node[fill=PineGreen!20,anchor=base] (1){$t_4$};}
\&
\\
\&
t_5 = t_4^2
\&
\&
\overline{t_5} 
= \frac{\partial \ell}{\partial t_5}
= \frac{1}{2}
\&
\\
\&
\ell = \frac{1}{2} t_5
\&
\&
\overline{\ell} = 1
\&
\\
};
\path[-latex, very thick, color=RoyalBlue!100]
(m-1-1) edge node [sloped, anchor=center, above, rotate=180] {\textsf{\small forward}} (m-8-1);
\path[-latex, very thick, color=Plum!100]
(m-8-5) edge node [sloped, anchor=center, above, rotate=180] {\textsf{\small backward}} (m-1-5);
\begin{scope}[node distance=1mm]
    \draw[black,-] 
    coordinate[above=of m-1-3] (a)  
    coordinate[below=of m-8-3] (b)
    (a) -- (b);
\end{scope}
\end{tikzpicture}
\end{equation}

The backward pass consists again of numeric computations but at every step we need to know how the value computed at that line, say $\alpha$, is used so we are able to compute the correct partial derivative $\overline{\alpha}$.
In the case of \eqref{eq:ad_example_2} that is fairly straightforward as every intermediate value is only used in the next step i.e. $t_i$ only depend on $t_{i-1}$ and the parameters.
Consequently $\overline{t_i}$ only depends on $\overline{t_{i+1}}$ and $t_i$.
A more general example that has a slightly more complicated structure is the network $F(x;a,b):=\relu(a x + b) + (a x + b)$.
We can write this network out in primitive form as well, we omit the forward/backward arrows but add a dependency graph that shows how the intermediate results depend on each other.

\begin{equation}
\label{eq:ad_example_3}
\begin{tikzpicture}[baseline]
\tikzstyle{every node}=[
rounded corners=1mm
, anchor=west
]
\matrix (m) [
matrix of math nodes
, column sep = 24pt
, row sep = 4pt
, nodes in empty cells
, ampersand replacement=\&
] 
{
\&
\text{data: } x_0, y_0 \in \mathbb{R}
\&
\&
\&
\\
\&
\text{parameters: } a,b \in \mathbb{R}
\&
\&
\overline{a} = \overline{t_1} \, x_0, \ \ \overline{b}=\overline{t_2}
\&
\\
\&
t_1 = a x_0
\&
\&
\overline{t_1} = \overline{t_2}
\&
\\
\&
t_2
= t_1 + b
\&
\&
\overline{t_2} 
= \frac{\partial \ell}{\partial t_2}
= \frac{\partial \ell}{\partial t_3} \frac{\partial t_3}{\partial t_2} 
+ \frac{\partial \ell}{\partial t_4} \frac{\partial t_4}{\partial t_2}
= \overline{t_3} \, \mathbb{1}_{t_2 \geq 0}
+
\overline{t_4}
\&
\\
\&
t_3 = \relu(t_2)
\&
\&
\overline{t_3} = \overline{t_4}
\&
\\
\&
t_4 = t_2 + t_3
\&
\&
\overline{t_4}
=
\overline{t_5}
\&
\\
\&
t_5 = t_4-y_0
\&
\&
\overline{t_5} = \overline{t_6} \cdot 2 t_5 =  t_5
\&
\\
\&
t_6 = t_5^2
\&
\&
\overline{t_6} = \frac{1}{2}
\&
\\
\&
\ell = \frac{1}{2} t_6
\&
\&
\overline{\ell} = 1
\&
\\
};
%\path[-latex, very thick, color=RoyalBlue!100]
%(m-1-1) edge node [sloped, anchor=center, above, rotate=180] {\textsf{\small forward}} (m-8-1);
%\path[-latex, very thick, color=Plum!100]
%(m-8-5) edge node [sloped, anchor=center, above, rotate=180] {\textsf{\small backward}} (m-1-5);
\tikzstyle{every path}=[
-latex
, thick
, shorten <=2pt
, shorten >=2pt
]
\path[color=RoyalBlue!100]
(m-4-2.west) edge[bend left] (m-3-2.west)
(m-5-2.west) edge[bend left] (m-4-2.west)
(m-7-2.west) edge[bend left] (m-6-2.west)
(m-8-2.west) edge[bend left] (m-7-2.west)
(m-9-2.west) edge[bend left] (m-8-2.west);
\path[color=Plum!100]
(m-6-2.west) edge[bend left] (m-5-2.west)
(m-6-2.west) edge[bend left=80] (m-4-2.west);
%%%%%
\path[color=RoyalBlue!100]
(m-8-4.west) edge[bend right] (m-9-4.west)
(m-7-4.west) edge[bend right] (m-8-4.west)
(m-6-4.west) edge[bend right] (m-7-4.west)
(m-5-4.west) edge[bend right] (m-6-4.west)
(m-3-4.west) edge[bend right] (m-4-4.west);
\path[color=Plum!100]
(m-4-4.west) edge[bend right] (m-5-4.west)
(m-4-4.west) edge[bend right=80] (m-6-4.west);
\path[color=PineGreen!100]
(m-4-4.west) edge[bend right=20, dotted] (m-4-2.east)
(m-7-4.west) edge[bend right=20, dotted] (m-7-2.east);
\begin{scope}[node distance=1mm]
    \draw[black,-] 
    coordinate[above=of m-1-3] (a)  
    coordinate[below=of m-9-3] (b)
    (a) -- (b);
\end{scope}
\end{tikzpicture}
\end{equation}

The dependency graph of the gradients $\overline{t_i}$ in \eqref{eq:ad_example_3} is naturally the reverse of the dependency graph of the values $t_i$ augmented with dependencies on the results from the forward pass.
Both the values $t_3$ and $t_4$ depend on the value $t_2$ hence $\overline{t_2}$ depends on both $\overline{t_3}$ and $\overline{t_4}$ in addition to $t_2$ itself.
This double dependency causes the chain rule applied to $\overline{t_2}$ to have two terms, of course this generalizes to multiple dependencies.

Constructing this \emph{computational graph} is exactly how machine learning frameworks such as PyTorch implement gradient computation. 
Each time you perform an operation on one or more tensors a new node is added to the graph to record what operation was performed and on which inputs the output depends.
Then, when it becomes time to compute the gradient (i.e. \texttt{.backward()} is called in PyTorch) the graph is traversed back to front.

As mentioned, the graph records what operation was performed at each node.
This is necessary because what backward computation needs to be performed at each node depends on what the corresponding forward computation was, this is where our symbolic knowledge needs to be added.

\subsection{Implementing Operations}

In the previous section we saw how the evaluation of a neural network (or any computation for that matter) can be expressed as a computational graph where each node correspond to a (primitive) operation.
To be able to do the backward pass each node needs to know how to compute its own partial derivative(s).
Previous examples \eqref{eq:ad_example_1} \eqref{eq:ad_example_2} and \eqref{eq:ad_example_3} only used simple scalar operations, now will explore how to implement both the forward and backward computation for a general multivariate vector-valued function.

Let $F:\mathbb{R}^n \to \mathbb{R}^m$ be a differentiable map, let $\bm{x}=[x_1 \ \cdots \ x_n]^T \in \mathbb{R}^n$ and $\bm{y}=[y_1 \ \cdots \ y_m]^T \in \mathbb{R}^m$ so that $\bm{y}=F(\bm{x})$. 
Let $\ell \in \mathbb{R}$ be the final loss computed for a neural network that contains $F$ as one of its operations.
Then we can generalize our gradient notation from scalars to vector as
\begin{equation*}
    \overline{\bm{x}}
    :=
    \frac{\partial \ell}{\partial \bm{x}}
    :=
    \begin{bmatrix}
        \frac{\partial \ell}{\partial x_1}
        \\
        \vdots
        \\
        \frac{\partial \ell}{\partial x_n}
    \end{bmatrix}
    ,
    \qquad
    \overline{\bm{y}}
    :=
    \frac{\partial \ell}{\partial \bm{y}}
    :=
    \begin{bmatrix}
        \frac{\partial \ell}{\partial y_1}
        \\
        \vdots
        \\
        \frac{\partial \ell}{\partial y_m}
    \end{bmatrix}
    .
\end{equation*}
During the forward pass the task is: given $\bm{x}$ compute $\bm{y}=F(\bm{x})$.
During the the backward pass the task is: given $\bm{x}$ and $\overline{\bm{y}}$ compute $\overline{\bm{x}}$, we call this backward operation $\overline{F}:\mathbb{R}^n \times \mathbb{R}^m \to \mathbb{R}^n$. Let us see what it looks like
\begin{align*}
    \overline{\bm{x}}
    &=
    \begin{bmatrix}
        \frac{\partial \ell}{\partial x_1}
        \\
        \vdots
        \\
        \frac{\partial \ell}{\partial x_n}
    \end{bmatrix}
    \\
    &=
    \begin{bmatrix}
        \sum_{j=1}^m \frac{\partial \ell}{\partial y_j}
        \frac{\partial y_j}{\partial x_1}
        \\
        \vdots
        \\
        \sum_{j=1}^m \frac{\partial \ell}{\partial y_j}
        \frac{\partial y_j}{\partial x_n}
    \end{bmatrix}
    \\
    &=
    \begin{bmatrix}
        \sum_{j=1}^m \overline{y_j} \,
        \frac{\partial y_j}{\partial x_1}
        \\
        \vdots
        \\
        \sum_{j=1}^m \overline{y_j} \,
        \frac{\partial y_j}{\partial x_n}
    \end{bmatrix}
    \\
    &=
    \begin{bmatrix}
        \frac{\partial y_1}{\partial x_1} 
        & 
        \cdots 
        &
        \frac{\partial y_m}{\partial x_1}
        \\
        \vdots
        &
        &
        \vdots
        \\
        \frac{\partial y_1}{\partial x_n} 
        & 
        \cdots 
        &
        \frac{\partial y_m}{\partial x_n}
    \end{bmatrix}
    \begin{bmatrix}
        \overline{y_1}
        \vphantom{\frac{\partial y_1}{\partial x_1}}
        \\
        \vdots
        \\
        \overline{y_m}
        \vphantom{\frac{\partial y_1}{\partial x_1}}
    \end{bmatrix}
    \\
    &=
    J(\bm{x})^T \overline{\bm{y}}
    ,
\end{align*}
where $J(\bm{x})$ is the Jacobian matrix of $F$ evaluated in $\bm{x}$.
So not unexpectedly we end up with the general form of the chain rule.

\begin{example}[Pointwise addition]
    Let $F:\mathbb{R}^2 \times \mathbb{R}^2 \to \mathbb{R}^2$ be defined by $F(\bm{a},\bm{b}) := \bm{a} + \bm{b}$, with $\bm{a}=[a_1 \ a_2]^T \in \mathbb{R}^2$ and $\bm{b}=[b_1 \ b_2]^T$.
    We can equivalently define $F$ as $F:\mathbb{R}^4 \to \mathbb{R}^2$ as follows:
    \begin{equation*}
        \bm{c} = F(\bm{a},\bm{b})
        =
        \begin{bmatrix}
        1 & 0 & 1 & 0
        \\
        0 & 1 & 0 & 1
        \end{bmatrix}
        \begin{bmatrix}
        a_1
        \\
        a_2
        \\
        b_1
        \\
        b_2
        \end{bmatrix}
        =
        \begin{bmatrix}
        a_1 + b_1
        \\
        a_2 + b_2
        \end{bmatrix}
        .
    \end{equation*}
    
    Since this is a linear operator the Jacobian matrix is just the constant matrix above.
    The backward operation $\overline{F}:\mathbb{R}^4 \times \mathbb{R}^2 \to \mathbb{R}^4$ is then given by
    \begin{equation*}
        \begin{bmatrix}
            \overline{\bm{a}}
            \\
            \overline{\bm{b}}
        \end{bmatrix}
        =
        \overline{F}(\bm{a},\bm{b},\overline{\bm{c}})
        =
        \begin{bmatrix}
            1 & 0
            \\
            0 & 1
            \\
            1 & 0
            \\
            0 & 1
        \end{bmatrix}
        \begin{bmatrix}
        \overline{c_1}
        \\
        \overline{c_2}
        \end{bmatrix}
        =
        \begin{bmatrix}
        \overline{c_1}
        \\
        \overline{c_2}
        \\
        \overline{c_1}
        \\
        \overline{c_2}
        \end{bmatrix}
        =
        \begin{bmatrix}
            \overline{\bm{c}}
            \\
            \overline{\bm{c}}
        \end{bmatrix}
    \end{equation*}
    for a gradient $\overline{\bm{c}} = [c_1 \ c_2]^T \in \mathbb{R}^2$.
    
    So to implement this operation we do not have to retain the inputs $\bm{a}$ and $\bm{b}$, we can implement the backward calculation by simply copying the incoming gradient $\overline{\bm{c}}$ and passing the two copies up the graph.
\end{example}

\begin{example}[Copy]
    Let $F:\mathbb{R}^n \to \mathbb{R}^{2n}$ be given by
    \begin{equation*}
        \begin{bmatrix}
        \bm{b}
        \\[4pt]
        \bm{c}
        \end{bmatrix}
        = 
        F(\bm{a}) 
        = 
        \begin{bmatrix}
        a_1
        \\
        \vdots
        \\
        a_n
        \\[2pt]
        a_1
        \\
        \vdots
        \\
        a_n
        \end{bmatrix}
        ,
    \end{equation*}
    with $\bm{a},\bm{b},\bm{c} \in \mathbb{R}^n$. 
    This operation makes two copies of its input, it is clearly linear with Jacobian
    \begin{equation*}
        J = 
        \begin{bmatrix}
        I_n 
        \\[4pt]
        I_n
        \end{bmatrix}
        ,
    \end{equation*}
    where $I_n$ is a unit matrix of size $n \times n$.
    Given gradients $\overline{\bm{b}},\overline{\bm{c}} \in \mathbb{R}^n$ of the outputs the gradient $\overline{\bm{a}}$ of the input is calculated as:
    \begin{equation*}
        \overline{\bm{a}}
        =
        \overline{F}(\bm{a},\overline{\bm{b}},\overline{\bm{c}})
        =
        \begin{bmatrix}
            I_n & I_n
        \end{bmatrix}
        \begin{bmatrix}
        \overline{\bm{b}}
        \\[4pt]
        \overline{\bm{c}}
        \end{bmatrix}
        =
        \overline{\bm{b}} + \overline{\bm{c}}
        .
    \end{equation*}
\end{example}

\begin{example}[Inner product]
    In this example the Jacobian is not constant and we do have to retain the inputs to be able to do the backward pass.
    Let $F:\mathbb{R}^{2n} \to \mathbb{R}$ be given by
    \begin{equation*}
        c 
        = 
        F(\bm{a},\bm{b})
        =
        \sum_{i=1}^n a_i b_i
        ,
    \end{equation*}
    for $\bm{a}=[a_1 \ \cdots\ a_n]^T$ and $\bm{b}=[b_1 \ \cdots\  b_n]^T \in \mathbb{R}^n$.
    Then the Jacobian matrix evaluated at $[\bm{a} \ \bm{b}]^T$ is given by
    \begin{equation*}
        J(\bm{a},\bm{b})
        =
        \begin{bmatrix}
            b_1 & \cdots & b_n & a_1 & \cdots & a_n
        \end{bmatrix}
        =
        \begin{bmatrix}
            \bm{b}^T & \bm{a}^T
        \end{bmatrix}
        .
    \end{equation*}
    Given a (scalar) gradient $\overline{c} \in \mathbb{R}$ of the output we compute the gradients of $\bm{a}$ and $\bm{b}$ as follows:
    \begin{equation*}
        \begin{bmatrix}
        \overline{\bm{a}}
        \\[4pt]
        \overline{\bm{b}}
        \end{bmatrix}
        =
        \overline{F}(\bm{a},\bm{b},\overline{c})
        =
        \begin{bmatrix}
            \bm{b}^T & \bm{a}^T
        \end{bmatrix}^T
        \, \overline{c}
        =
        \overline{c} \,
        \begin{bmatrix}
            \bm{b} 
            \\[4pt] 
            \bm{a}
        \end{bmatrix}
        ,
    \end{equation*}
    which depends on the input values $\bm{a}$ and $\bm{b}$.
\end{example}

In \eqref{eq:ad_example_2} and \eqref{eq:ad_example_3} we deconstructed the loss function $(y,y') \mapsto \frac{1}{2}(y-y')^2$ into three primitive operations.
As a consequence the backward pass has a step where we first multiply with $2$ and then multiply with $\sfrac{1}{2}$, this is not efficient of course.
Hence deconstructing our calculation into the smallest possible operations is not always advisable.
In this example we will express the $L^2$ loss function as single operation instead.

\begin{example}[$L^2$ loss]
    Let $F:\mathbb{R}^{2n} \to \mathbb{R}$ be given by
    \begin{equation*}
        \ell 
        = 
        F(\bm{y},\bm{y}')
        =
        \frac{1}{2} \sum_{i=1}^{n} (y_i - y_i')^2
    \end{equation*}
    for $\bm{y}=[y_1 \ \cdots\ y_n]^T$ and $\bm{y}'=[y_1' \ \cdots\  y_n']^T \in \mathbb{R}^n$.
    Then the Jacobian matrix evaluated at $[\bm{y} \ \bm{y}']^T$ is given by
    \begin{equation*}
        J(\bm{y},\bm{y}')
        =
        \begin{bmatrix}
            y_1 - y_1' & \cdots & y_n - y_n' & y_1'-y_1 & \cdots & y_n'-y_n
        \end{bmatrix}
        =
        \begin{bmatrix}
            (\bm{y}-\bm{y}')^T & (\bm{y}'-\bm{y})^T
        \end{bmatrix}
        .
    \end{equation*}
    The backward operation has signature $\overline{F}:\mathbb{R}^{2n} \times \mathbb{R} \to \mathbb{R}^{2n}$. 
    The second argument of $\overline{F}$ is a given gradient $\overline{\ell} \in \mathbb{R}$, which is trivially $\overline{\ell}=1$ if the output $\ell$ is the final loss we are interested in minimizing.
    We then compute the gradients as follows:
    \begin{equation*}
        \begin{bmatrix}
        \overline{\bm{y}}
        \\[4pt]
        \overline{\bm{y}'}
        \end{bmatrix}
        =
        \overline{F}(\bm{y},\bm{y}',\overline{\ell}=1)
        =
        \begin{bmatrix}
            (\bm{y}-\bm{y}')^T & (\bm{y}'-\bm{y})^T
        \end{bmatrix}^T
        \, \overline{\ell}
        =
        \begin{bmatrix}
            \bm{y}-\bm{y}'
            \\[4pt]
            \bm{y}'-\bm{y}
        \end{bmatrix}
        .
    \end{equation*}
    If $\bm{y}$ is the output of the neural network and $\bm{y}'$ is the data point then we are only interested in $\overline{\bm{y}}$ and we would only compute $\bm{y}-\bm{y}'$.
    This is equivalent to the computation in \eqref{eq:ad_example_2} and \eqref{eq:ad_example_3} but avoids the redundant multiplications.
\end{example}

%%%%%%%%%%%%%%%%%%%%%%%%%%%%%%%%%%%%%%%%%%%%%%%%%%%%%%%%%%%%%%%%%%%%%%%%%%%%%%%%%
\section{Adaptive Learning Rate Algorithms}

The learning rate is crucial to the success of the training process.
In general the loss is highly sensitive to some parameters but insensitive to others, just think about parameters in different layers.
Momentum alleviates some of the issues but introduces problems of its own and adds another hyperparameter we have to tune.

Assigning each parameters its own learning rate (and continually adjusting it) is not feasible.
What we need are automatic methods.
This has lead to the development of \emph{adaptive learning (rate) algorithms}.
The idea is to set the learning rate dynamically per-parameter at each iteration based on the history of the gradients.
We will look at three of these methods: \emph{Adagrad}, \emph{RMSProp} and \emph{Adam}.

Let us recall the setting.
We have a parameter space $W = \mathbb{R}^N$ and some initial parameter value $w_0 \in W$. 
Let $I_t$ be the batch at iteration $t \in \mathbb{N}$ and $\ell_{I_t}$ its associated loss function. We will abbreviate the current batch's gradient as:
\begin{equation*}
    g_t := \nabla \ell_{I_t} (w_t).
\end{equation*}

The SGD update rule with learning rate $\eta>0$ is then:
\begin{equation*}
    w_{t+1}
    =
    w_t - \eta g_t
    .
\end{equation*}
With momentum the update rule becomes:
\begin{equation*}
    \begin{split}
        v_{t} &= \mu v_{t-1} - \eta g_t,
        \\
        w_{t+1} &= w_t + v_t,
    \end{split}
\end{equation*}
where $v_0 = 0$ and $\mu \in [0,1)$ is the momentum factor.

\subsection{Adagrad}

Adagrad (Adaptive Gradient Descent) was one of the first adaptive learning rate methods introduced by  \textcite{duchi2011adaptive}.

Let $t \in \mathbb{N}_0$, $I_t$ the batch index set at iteration $t$ and $\ell_{I_t}$ its associated loss. 
Let $w_0$ be the initial parameter values and abbreviate 
$g_t := \nabla \ell_{I_t}(w_t)$.

The Adagrad update is defined per-parameter as:
\begin{equation*}
    \left( w_{t+1} \right)_i
    =
    \left( w_t \right)_i
    -
    \frac{\eta}{\sqrt{\sum_{k=0}^t \left( g_k \right)_i^2} 
    + \varepsilon}
    \left( g_t \right)_i
    .
\end{equation*}
Or when we take all the operations on the vectors to mean component-wise operations we can write more concisely:
\begin{equation*}
    w_{t+1}
    =
    w_t
    -
    \frac{\eta}{\sqrt{\sum_{k=0}^t  g_k^2 } 
    + \varepsilon}
    g_t
    .
\end{equation*}

At each iterations we slow down the effective learning rate of each individual parameter by the $L^2$ norm of the history of the partial derivatives with respect to that parameter. As $(g_t)_i$ may be very small or zero we add a small positive $\varepsilon$ (say $10^{-8}$ or so) for numerical stability.

The benefit of this method is that choosing $\eta$ becomes less important, the step size will eventually decrease to the point that it can settle into a local minimum.

On the other hand, the denominator $\sqrt{\sum_{k=0}^t \left( g_k \right)_i^2}$ is monotonically decreasing and will eventually bring the training process to a halt whether a local minimum has been reach or not.

\subsection{RMSProp}

Root Mean Square Propagation or RMSProp is an adaptive learning rate algorithm introduced by  \textcite{tieleman2012lecture}.

RMSProp uses the same basic idea as Adagrad, but instead of accumulating the squared gradients it uses a weighted average of the historic gradients. Let $\alpha \in (0,1)$:
\begin{equation*}
    \begin{split}
    v_{t} &= \alpha v_{t-1} + (1-\alpha) g_t^2
    \\
    w_{t+1} &= w_t - \frac{\eta}{\sqrt{v_t} + \varepsilon} g_t,
    \end{split}
\end{equation*}
where again all the arithmetic operations are applied component-wise. The $\alpha$ factor is called the \emph{forgetting factor}, \emph{decay rate} or \emph{smoothing factor}. The suggested hyperparameter values to start with are $\eta=0.001$ and $\alpha=0.9$.

\subsection{Adam}

The Adaptive Moment Estimation method, or Adam, was introduced by \textcite{kingma2017adam}.
In addition to storing an exponentially decaying average of past square gradients $v_t$ like RMSProp, Adam also keeps a running average of past gradients $m_t$, similar to momentum.

Let $\beta_1, \beta_1 \in (0,1)$,
\begin{equation*}
    \begin{split}
    m_{t} &= \beta_1 m_{t-1} + (1-\beta_1) g_t ,
    \\
    v_{t} &= \beta_2 v_{t-1} + (1-\beta_2) (g_t)^2 ,
    \end{split}
\end{equation*}
with $m_0 = v_0 = 0$.
The vectors $m_t$ and $v_t$ are estimates of the first moment (the mean) and the second moment (the uncentered variance), hence the name of the method.

As we start with $m_0 = v_0 =0$ the estimates are biased towards zero. Let us see how biased and whether we can correct that bias. By induction we have:
\begin{equation*}
    \begin{split}
        m_t &= (1-\beta_1) \sum_{i=1}^t \beta_1^{t-i} g_i,
        \\
        v_t &= (1-\beta_2) \sum_{i=1}^t \beta_2^{t-i} (g_i)^2,
    \end{split}
\end{equation*}
and so:
\begin{equation*}
    \begin{split}
        \mathbb{E}[m_t] &= (1-\beta_1) \sum_{i=1}^t \beta_1^{t-i}  \mathbb{E}[g_i],
        \\
         \mathbb{E}[v_t] &= (1-\beta_2) \sum_{i=1}^t \beta_2^{t-i}  \mathbb{E}[(g_i)^2].
    \end{split}
\end{equation*}

Assuming $\mathbb{E}[g_i]$ and $ \mathbb{E}[(g_i)^2]$ are stationary (i.e. do not depend on $i$) we get:
\begin{equation*}
    \begin{split}
        \mathbb{E}[m_t] &= \left((1-\beta_1) \sum_{i=1}^t \beta_1^{t-i}\right)  \mathbb{E}[g_t],
        \\
         \mathbb{E}[v_t] &= \left((1-\beta_2) \sum_{i=1}^t \beta_2^{t-i}  \right) \mathbb{E}[(g_t)^2].
    \end{split}
\end{equation*}
Which simplifies to:
\begin{equation*}
    \begin{split}
        \mathbb{E}[m_t] &= (1-\beta_1^t)  \mathbb{E}[g_t],
        \\
         \mathbb{E}[v_t] &= (1-\beta_2^t) \mathbb{E}[(g_t)^2].
    \end{split}
\end{equation*}
Therefore, the bias-corrected moments are:
\begin{equation*}
    \hat{m}_t = \frac{m_t}{1-\beta_1^t}
    ,
    \qquad 
    \hat{v}_t = \frac{v_t}{1-\beta_2^t}
    .
\end{equation*}
The update rule of Adam is then given by:
\begin{equation*}
    w_{t+1} = w_t - \eta \frac{\hat{m}_t}{\sqrt{\hat{v}_t} + \varepsilon},
\end{equation*}
where again all arithmetic is applied component-wise.

The default hyperparameter values suggested by  \textcite{kingma2017adam} are $\eta=0.001$, $\beta_1=0.9$, $\beta_2=0.99$ and $\varepsilon=10^{-8}$.

\subsection{Which variant to use?}

Which of these algorithms should you use? 
If your data is sparse (very high dimensional data usually is) then the adaptive learning rate methods usually outperform plain SGD (with or without momentum).
\textcite{kingma2017adam} showed that due to its bias correction Adam performs marginally better late in the training process. 
In that regard Adam is a safe option to try first.
Nonetheless trying different algorithms to see which one works best for any given problem can be worthwhile.

\begin{figure}[ht!]
    \centering
    \includegraphics[width=0.8\linewidth]{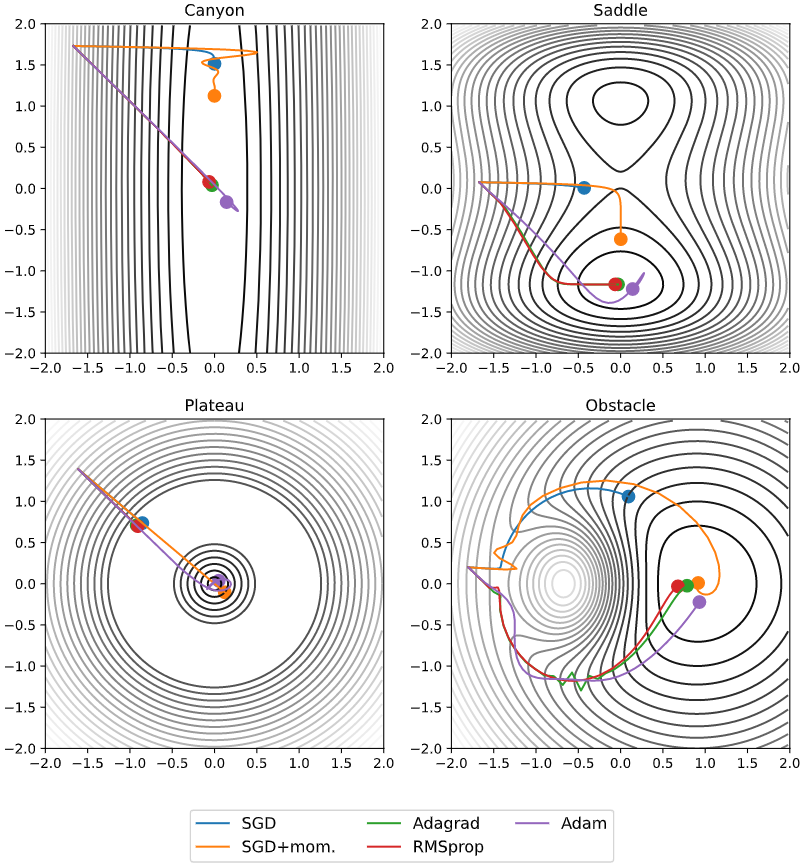}
    \caption{
        Comparing 5 SGD variants in 4 different situations for a number of steps.
        In the \textit{canyon} situation we have one parameters that has a large effect on the loss and one parameter that has little effect on the loss.
        The \textit{saddle} situation has a prominent saddle point.
        The \textit{plateau} has a large flat section with a vanishing gradient that needs to be traversed to get to the minimum.
        The final situation has an \textit{obstacle} the algorithm has to go around to reach the minimum.
        Dark level sets indicate lower function values.
        This figure was generated with \texttt{SGDDemonstration.ipynb}.
    }
    \label{fig:sgd_variants}
\end{figure}

Fig.~\ref{fig:sgd_variants} illustrates the different behaviour of the methods we have discussed in some artificial loss landscapes.
These landscapes model some of the problems the algorithms may encounter and show some of the strengths and weaknesses of each method.

There are many more SGD variants we have not discussed, you can look at the \texttt{toch.optim} namespace in the PyTorch documentation to see the available algorithms.

%%%%%%%%%%%%%%%%%%%%%%%%%%%%%%%%%%%%%%%%%%%%%%%%%%%%%%%%%%%%%%%%%%%%%%%%%%%%%%%%%
%%%
%%%
%%%
%%%
%%%
%%%%%%%%%%%%%%%%%%%%%%%%%%%%%%%%%%%%%%%%%%%%%%%%%%%%%%%%%%%%%%%%%%%%%%%%%%%%%%%%%
\chapter{Equivariance}
\label{ch:equivariance}

\begin{figure}[H]
    \centering
    \includegraphics[width=0.7\linewidth]{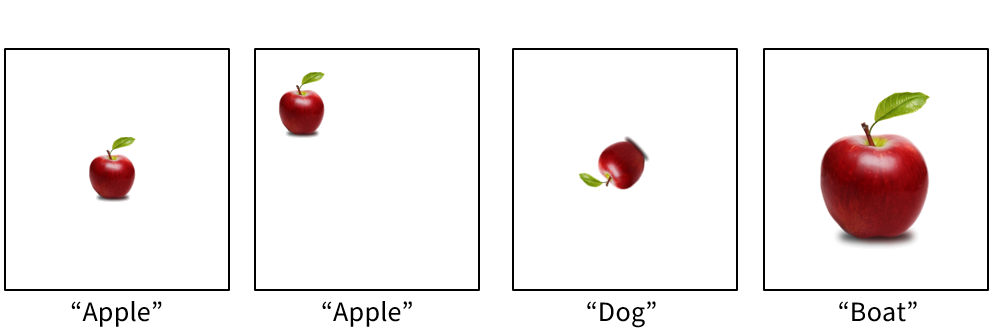}
    \caption{
    We would like a classification network to be invariant under translation, rotation and scaling.
    We could train our network with translated, rotated and scaled versions of our original data and hope the invariance gets encoded that way.
    But this would drastically increase the training time and gives no guarantee of success.
    Preferable would be designing the network in such a way that it is intrinsically invariant.
    }
    \label{fig:apples}
\end{figure}

There are many applications where we want the neural network to have certain symmetries, such as in Fig.~\ref{fig:apples}.
Most applications come with natural symmetries. 
It might be rotation-translation invariance for a medical diagnosis application that detects tumors in X-ray images or time invariance for a weather forecasting system.

The desired symmetry might not be an actual invariance in that the output is not expected to remain the same but instead transform in some manner similar to the input. 
For example: in an image enhancement application the output image is expected to rotate and translate along with the input image.
Some authors use \emph{invariance} to denote both cases, we will use the term \emph{equivariance} (as in: \textit{transforms with}) to refer to both and use invariance for the special case where the output should stay the same.

Expressing many of these symmetries on discrete domains is awkward, indeed rotations and translations of an image are not even well defined unless the translation is on grid or the rotation is in $90^\circ$ increments.
So instead we will be working in the continuous setting and only discretizing when it becomes time to implement our ideas.
For images, instead of representing them as elements of $\mathbb{R}^{H \times W}$ we represent them as elements of $C_c^\infty(\mathbb{R}^2)$, i.e. smooth functions with compact support.

Our eventual practical goal for this chapter is building a type of CNN that is not only translation equivariant but also rotation equivariant.
But being mathematicians we want to be general and develop a theory that allows for other transformations as well. 
This brings us to the theory of Lie groups, which are essentially continuous transformation groups, something we will make precise in this chapter.

Lie group theory plays an important role in many disparate fields of study such as geometric image analysis, computer vision, particle physics, financial mathematics, robotics, etc.
In the next sections we will build up a general equivariance framework based on Lie groups.
The payoff of this theoretical work will be a general recipe for building a neural network that is equivariant with respect to an arbitrary transformation group: a so called Group Equivariant CNN \parencite{cohen2016group,cohen2020general}, or \emph{G-CNN} for short.

%%%%%%%%%%%%%%%%%%%%%%%%%%%%%%%%%%%%%%%%%%%%%%%%%%%%%%%%%%%%%%%%%%%%%%%%%%%%%%%%%
\section{Manifolds}

We start with the basic object we will be working with: the manifold.
%Intuitively the manifold is the answer to the question: what is the minimum amount of structure a set has to have to be able to do analysis on it.
We are accustomed to doing analysis on $\mathbb{R}^n$ but we also know that there are non-Euclidean spaces of importance. 
Classic example is the unit circle $S^1$, which is distinctly non-Euclidean but still admits derivatives, integrals, PDEs, etc.
When working with an object such as $S^1$ we usually do it indirectly by parameterizing it some way, such as with an angle $\theta \in [0,2 \pi)$, so that at least locally it resembles $\mathbb{R}$.
We can generalize this and consider all spaces that we can, at least locally, identify with a subset of $\mathbb{R}^n$.

\subsection{Characterization}

The proper way to introduce manifolds is starting with a set and then adding several layers of structure, as is done in \textcite{lee2010topological,lee2013smooth}.
After going through that laborious process we would then see a characterization that is a more practical tool for constructing manifolds.
In this course we will skip straight to that characterization in the form of the following lemma.

\begin{lemma}[Smooth manifold chart lemma]
\label{lemma:smooth_manifold_chart}
Let $M$ be a set and suppose we are given a collection of subsets $\{ U_\alpha \}_{\alpha \in I}$ of $M$ for some index set $I$, together with maps $\varphi_\alpha:U_\alpha \to \mathbb{R}^n$. Then $M$ together with $\left\{ \left( U_\alpha, \varphi_\alpha \right) \right\}_{\alpha \in I}$ gives a smooth $n$-dimensional manifold if the following conditions are satisfied.
\begin{enumerate}[label=(\roman*)]
    \item For all $\alpha \in I$, $\varphi_\alpha$ is a bijection between $U_\alpha$ and an open subset of $\mathbb{R}^n$.
    
    \item For each $\alpha,\beta \in I$, the sets $\varphi_\alpha \left( U_\alpha \cap U_\beta \right)$ and $\varphi_\beta \left( U_\alpha \cap U_\beta \right)$ are open in $\mathbb{R}^n$.
    
    \item When $U_\alpha \cap U_\beta \neq \varnothing$ then the map $\tau_{\alpha,\beta} := \varphi_\beta \circ \varphi_\alpha^{-1} : \varphi_\alpha(U_\alpha \cap U_\beta) \to \varphi_\beta(U_\alpha \cap U_\beta)$ is smooth.
    
    \item There exists an (at most) countably infinite $J \subset I$ so that $\bigcup_{\alpha \in J} U_\alpha = M$, i.e. there exists a countable cover of $M$.
    
    \item Whenever $p_1$ and $p_2$ are distinct point in $M$, there exist $U_\alpha$ and $U_\beta$ (not necessarily distinct) so that there exist disjoint sets $V \subset U_\alpha$ and $W \subset U_\beta$, with $\varphi_\alpha(V)$ and $\varphi_\beta(W)$ open in $\mathbb{R}^n$, with $p_1 \in V$ and $p_2 \in W$.
\end{enumerate}
\end{lemma}

\begin{figure}[H]
    \centering
    \begin{tikzpicture}
    %\draw[smooth cycle,tension=3.0] plot coordinates{(1,0) (1,1) (2,2) (3,3) (6,0)} node [label=right:$M$];

%\def\colora{RoyalBlue!80}
%\def\colorb{Red!80}
%\def\colorab{Plum!80}

\def\colora{RoyalBlue!70}
\def\colorb{Plum!70}
\def\colorab{RoyalBlue!50!Plum!80}

% Manifold
\draw[smooth cycle, tension=1.0] 
plot coordinates{(0,0.1) (2.3,0.3) (4.5,0.2) (4.6,2.8) (2.6,2.9) (0.1,3.1)};
\node [] at (0.4, 2.8) {$M$};

\def\Ua{
plot [smooth cycle, tension=0.6] 
coordinates{(1.1, 1.8) (2.2, 1.0) (3.2, 1.6) (2.0, 2.7)}
}

\def\Ub{
plot [smooth cycle, tension=0.6] 
coordinates{(2.2, 0.9) (3.6, 0.8) (3.9, 1.9) (2.4, 2.1)}
}

\begin{scope}[]
	\clip \Ua;
	\fill [\colora, even odd rule] \Ua \Ub;
\end{scope}

\begin{scope}[]
	\clip \Ub;
	\fill [\colorb, even odd rule] \Ub \Ua;
\end{scope}

\begin{scope}[]
	\clip \Ub;
	\fill [\colorab] \Ua;
\end{scope}

\draw \Ua;
\node [] at (1.5, 1.8) {$U_{\alpha}$};

\draw \Ub;
\node [] at (3.6, 1.6) {$U_{\beta}$};

\coordinate (p1) at (1.8,2.3);
\coordinate (p3) at (3.4,1.0);

\begin{scope}[xshift=7.5cm, yshift=2.5cm]
	
	\def\Ua1{
	plot [smooth cycle, tension=0.4] 
	coordinates {(0.4, 0.5) (1.8, 0.4) (1.9, 1.9) (0.5, 2.0)}
	}

	\def\Ub1{
	plot [smooth cycle, tension=0.9] 
	coordinates {(0.3, 0.6) (1.9, 1.4) (1.9, 0.1) (0.5, 0.1)}
	}

	\draw[-latex] (-0.2,0) -- (2.5,0) node[above] {$\mathbb{R}^n$};
	\draw[-latex] (0,-0.2) -- (0,2.5);
	
	\begin{scope}
		\clip \Ua1;
		\fill [\colora, even odd rule] \Ua1 \Ub1;
	\end{scope}

	\begin{scope}[]
		\clip \Ub1;
		\fill [\colorab] \Ua1;
	\end{scope}

	\draw \Ua1;
	\clip \Ua1;
	\draw \Ub1;

	\coordinate (p2) at (0.8, 1.6);
	\coordinate (p5) at (1.4, 0.7);
	\coordinate (p7) at (1.0, 0.7);
\end{scope}

\begin{scope}[xshift=7.5cm, yshift=-1.5cm]

	\def\Ub2{
	plot [smooth cycle, tension=0.5] 
	coordinates {(0.4, 0.5) (1.9, 0.6) (1.9, 2.0) (0.4, 1.8)}
	}

	\def\Ua2{
	plot [smooth cycle, tension=0.9] 
	coordinates {(0.3, 2.2) (1.9, 2.4) (1.9, 1.8) (0.8, 1.2)}
	}

	\draw[-latex] (-0.2,0) -- (2.5,0) node[above] {$\mathbb{R}^n$};
	\draw[-latex] (0,-0.2) -- (0,2.5);

	\begin{scope}
		\clip \Ub2;
		\fill [\colorb, even odd rule] \Ub2 \Ua2;
	\end{scope}

	\begin{scope}[]
		\clip \Ua2;
		\fill [\colorab] \Ub2;
	\end{scope}

	\draw \Ub2;
	\clip \Ub2;
	\draw \Ua2;

	\coordinate (p4) at (0.7, 0.7);
	\coordinate (p6) at (1.4, 1.7);
	\coordinate (p8) at (1.0, 1.7);
\end{scope}

\begin{scope}[]
	\path [{Circle[length=0.8mm]}-{Latex[length=2mm]}] 
	(p1) edge[bend left=20] node[anchor=center, above] {$\varphi_{\alpha}$} (p2);
	
	\path [{Circle[length=0.8mm]}-{Latex[length=2mm]}] 
	(p3) edge[bend right=20] node[anchor=center, below] {$\varphi_{\beta}$} (p4);

	\path [{Circle[length=0.8mm]}-{Latex[length=2mm]}] 
	(p5) 
	edge[bend left=20] node[anchor=center, right] 
	{$\tau_{\alpha,\beta}=\varphi_\beta \circ \varphi_\alpha^{-1}$} 
	(p6);

	\path [{Circle[length=0.8mm]}-{Latex[length=2mm]}] 
	(p8) 
	edge[bend left=20] node[anchor=center, left] 
	{$\tau_{\beta,\alpha}=\varphi_\alpha \circ \varphi_\beta^{-1}$} 
	(p7);
\end{scope}

\end{tikzpicture}
    \caption{A manifold $M$ with two charts $\left(U_\alpha, \varphi_\alpha\right)$ and $\left( U_\beta, \varphi_\beta\right)$ and their transitions maps $\tau_{\alpha,\beta}$ and $\tau_{\beta,\alpha}$.}
    \label{fig:manifold_charts}
\end{figure}
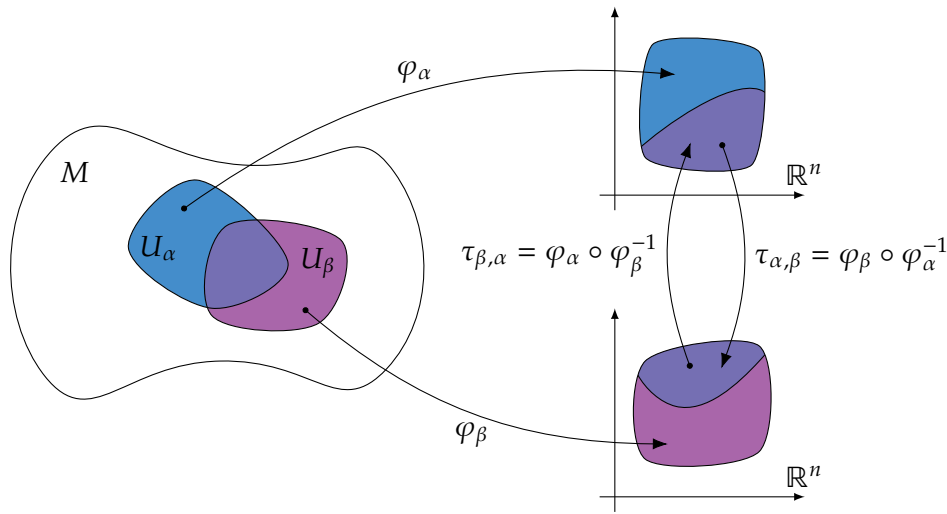

We say that each pair $\left( U_\alpha, \varphi_\alpha \right)$ is a (smooth) \textbf{chart} and that the set $\left\{ \left( U_\alpha, \varphi_\alpha \right) \right\}_{\alpha \in I}$ is a (smooth) \textbf{atlas} of $M$.
The maps $\tau_{\alpha,\beta} := \varphi_\beta \circ \varphi_\alpha^{-1}$ are called the atlas' \textbf{transition maps}. 
Since the transition maps are from $\mathbb{R}^n$ to $\mathbb{R}^n$ we know exactly what it means for these maps to be smooth.
Charts that have smooth transition maps both ways are said to be \emph{compatible}.
This construction is illustrated in Fig.~\ref{fig:manifold_charts}, for details and proof see \textcite[Ch. 1]{lee2013smooth}.

%\begin{figure}[H]
%    \centering
%    \includegraphics[width=0.7\textwidth]{figures/manifold-charts.pdf}
%    \caption{A manifold $M$ with two charts $\left(U_\alpha, \varphi_\alpha\right)$ and $\left( U_\beta, \varphi_\beta\right)$ and their transitions maps $\tau_{\alpha,\beta}$ and $\tau_{\beta,\alpha}$.}
%    \label{fig:manifold_charts}
%\end{figure}

\begin{remark}[The manifold is not the atlas]
    In Lemma~\ref{lemma:smooth_manifold_chart} we say that the set $M$ along with an atlas $\left\{ \left( U_\alpha, \varphi_\alpha \right) \right\}_{\alpha \in I}$ \emph{gives} a smooth manifold rather than stating that it \emph{is} a smooth manifold.
    This is because we could construct another atlas compatible with the first but the set together with the new atlas would still describe the same manifold.
    In example \ref{example:unit_circle} this distinction will be clear as we can see that we can construct any number of atlases describing the same manifold.
    
    We say two atlases are compatible if the transition maps between charts of the two atlases are also smooth, if two atlases are compatible they describe the same manifold.
    Formally we could define a smooth manifold as the equivalence class of all atlases per Lemma~\ref{lemma:smooth_manifold_chart} under the equivalence relation of the atlases being compatible.
    
    We could also consider the union of all possible compatible atlases, called the \emph{maximal atlas} as defining the manifold.
\end{remark}

\begin{remark}[Hausdorff space]
    Item \textit{(v)} of Lemma~\ref{lemma:smooth_manifold_chart}  guarantees that a manifold is a Hausdorff space, this simply means that for any two distinct points $p_1$ and $p_2$ we can find neighborhoods of both that are disjoint (a neighborhood in the manifold being a preimage of a neighborhood in a chart).
    This may seem redundant but one can construct counterexamples that satisfy \textit{(i)-(iv)} but not \textit{(v)}.
    The Hausdorff property is needed for limits to be unique and manifolds are required to be Hausdorff for that reason.
\end{remark}

We define what the \emph{open subsets} of the manifold are by looking at their images in $\mathbb{R}^n$. 
A subset $V \subset M$ is open if $\varphi_\alpha \left( V \cap U_\alpha \right)$ is open in $\mathbb{R}^n$ for all $\alpha \in I$. In other words: we let the atlas and the standard topology on $\mathbb{R}^n$ define the topology of the manifold. 
With this definition we could rephrase item \textit{(v)} of Lemma~\ref{lemma:smooth_manifold_chart} as: for any two distinct points of $M$ there exist \emph{neighborhoods} (i.e. open subsets containing the point in question) of said two points that are disjoint.

Of course if $\left( U_\alpha, \varphi_\alpha \right)$ is a chart then if $V \subset U_\alpha$ is open then $\left( V, \varphi_\alpha\vert_V \right)$ is also a chart.

\begin{remark}[Open sets and continuous charts]
    Defining the open sets of $M$ as the preimages of open subsets of $\mathbb{R}^n$ makes the charts continuous maps by definition.
    It might be the case that when we are constructing a manifold we do not start with $M$ being just a set.
    For example, in many cases we start with $M$ being a subset of $\mathbb{R}^n$, in that case we already know what the open subsets of $M$ are (i.e. the topology of $M$ is already known).
    If we already decided what the open subsets of $M$ are going to be then we need to make sure that the two notions of `open' coincide.
    This requires that the charts are continuous maps by construction instead of them being continuous by definition.
\end{remark}

Lemma \ref{lemma:smooth_manifold_chart} is a characterization so it works the other way around as well. If $M$ is a smooth manifold then it is always possible to produce a smooth atlas, i.e. a set $\left\{ \left( U_\alpha, \varphi_\alpha \right) \right\}_{\alpha \in I}$ that satisfies all the conditions of the lemma.

\begin{example}[The unit circle]
\label{example:unit_circle}
Let $S^1$ be the unit circle in $\mathbb{R}^2$, i.e. $S^1 = \left\{ (x,y) \ \middle\vert\ x^2+y^2=1 \right\}$. Then we can construct two smooth charts that cover $S^1$:
\begin{align*}
    & U_1 = S^1 \setminus \left\{ \left(-1,0\right) \right\}, \qquad  \varphi_1(x,y): U_1 \to (-\pi,\pi),
    \\
    & U_2 = S^1 \setminus \left\{ \left( 1,0 \right) \right\}, \qquad \varphi_2(x,y): U_2 \to (0,2\pi)
    .
\end{align*}
Where $\varphi_1$ gives the angle between the $x$-axis and $(x,y)$ measured from $-\pi$ to $\pi$ and $\varphi_2$ gives that same angle measured from $0$ to $2 \pi$, as illustrated in Figure~\ref{fig:s1_charts}.

Clearly $U_1 \cup U_2 = S^1$ and we see that $\varphi_1(U_1)=\left(-\pi,\pi\right)$ and $\varphi_2(U_2)=\left(0,2\pi \right)$, so these two charts form an atlas of $S^1$. Now we have to check the transition maps, note that $U_1 \cap U_2 = S^1 \setminus \left\{ (-1,0), (1,0) \right\}$ which consists of two disjoint components, so the transition map from the first to the second chart would have as domain $(-\pi,0) \cup (0,\pi)$ and be defined as:
\begin{equation*}
    \tau_{1,2} (\theta) = \varphi_2 \circ \varphi_1^{-1} (\theta)
    =
    \begin{cases}
        \theta & \text{ if } \theta \in (0,\pi),
        \\
        \theta + 2 \pi & \text{ if } \theta \in (-\pi,0),
    \end{cases}
\end{equation*}
which is smooth on its domain (the discontinuity at $\theta=0$ is not part of the domain). The inverted transition map is similarly smooth. The whole construction is visualized in Fig. \ref{fig:s1_charts}.
%\begin{figure}[H]
%    \centering
%    \includegraphics[width=0.7\linewidth]{S1_charts.pdf}
%    \caption{A smooth atlas on $S^1$ with the transition maps illustrated in grey.}
%    \label{fig:s1_charts}
%\end{figure}
\begin{figure}[H]
    \begin{minipage}[c]{0.6\textwidth}
    \centering
    \begin{tikzpicture}[scale=1.2]

\def\colora{RoyalBlue!80}
\def\colorb{Plum!80}
\def\colorab{Plum!80}

\draw[-latex] (-1.6, 0) -- (1.6, 0) node[right] {$x$}; 
\draw[-latex] (0, -1.6) -- (0, 1.6) node[above] {$y$}; 
\draw[thick] circle [radius=1];

\draw[\colora, thick] (0, 0) circle [radius=1.1];
\draw[\colorb, thick] (0, 0) circle [radius=0.9];

\draw[\colora, fill=white] (-1.1, 0) circle [radius=0.06];
\draw[\colorb, fill=white] (0.9, 0) circle [radius=0.06];

\node[] at (-1,1) {$S^1$};
\node[\colora] at (1,1) {$U_1$};
\node[\colorb] at (-0.4, 0.4) {$U_2$};

\coordinate (p1) at (1, 0.5);
\coordinate (p3) at (0.76,-0.5);

\begin{scope}[xshift=4.5cm, yshift=1.2cm]
	
	\draw[-latex] (-1.3, 0) -- (2.5, 0) node[above] {$\mathbb{R}$};
	\draw[-] (-1,0.1) -- (-1,-0.1) node[below] {$-\pi$};
	\draw[-] (0,0.1) -- (0,-0.1) node[below] {$0$};
	\draw[-] (1,0.1) -- (1,-0.1) node[below] {$\pi$};
	\draw[-] (2,0.1) -- (2,-0.1) node[below] {$2\pi$};

	\draw[very thick, \colora] (-1,0) -- (1,0);
	\draw[\colora, fill=white] (-1,0) circle [radius=0.05];
	%\draw[\colora, fill] (0,0) circle [radius=0.05];
	\draw[\colora, fill=white] (1,0) circle [radius=0.05];
	
	\coordinate (p2) at (-1.5, 0);
	\coordinate (q1) at (-1, 0);
	\coordinate (q2) at (0, 0);
	\coordinate (q3) at (1, 0);
\end{scope}

\begin{scope}[xshift=4.5cm, yshift=-1.2cm]

	\draw[-latex] (-1.3, 0) -- (2.5, 0) node[above] {$\mathbb{R}$};
	\draw[-] (-1,0.1) -- (-1,-0.1) node[below] {$-\pi$};
	\draw[-] (0,0.1) -- (0,-0.1) node[below] {$0$};
	\draw[-] (1,0.1) -- (1,-0.1) node[below] {$\pi$};
	\draw[-] (2,0.1) -- (2,-0.1) node[below] {$2\pi$};

	\draw[very thick, \colorb] (0, 0) -- (2, 0);
	\draw[\colorb, fill=white] (0, 0) circle [radius=0.05];
	%\draw[\colorb, fill] (1, 0) circle [radius=0.05];
	\draw[\colorb, fill=white] (2, 0) circle [radius=0.05];
	
	\coordinate (p4) at (-1.5, 0);
	\coordinate (r1) at (0, 0);
	\coordinate (r2) at (1, 0);
	\coordinate (r3) at (2, 0);
\end{scope}

\begin{scope}
	\path [-{Latex[length=2mm]}] 
	(p1) 
	edge[bend left=20] node[anchor=center, above] 
	{$\varphi_1$} 
	(p2);

	\path [-{Latex[length=2mm]}] 
	(p3) 
	edge[bend right=20] node[anchor=center, below] 
	{$\varphi_2$} 
	(p4);
\end{scope}

\begin{scope}[on background layer]
	\fill[black!50!white, fill opacity=0.5] (q1) to[out=-50, in=80] (r2) -- (r3) to[out=100, in=-120] (q2);
	\fill[black!50!white, fill opacity=0.5] (q2) to[out=-60, in=60] (r1) -- (r2) to[out=120, in=-120] (q3);
\end{scope}

\end{tikzpicture}
    \end{minipage}
    ~\hfill~
    \begin{minipage}[c]{0.3\textwidth}
    \captionsetup{width=\linewidth, format=plain}
    \caption{A smooth atlas on $S^1$ with the transition maps illustrated in grey.}
    \label{fig:s1_charts}
    \end{minipage}
\end{figure}
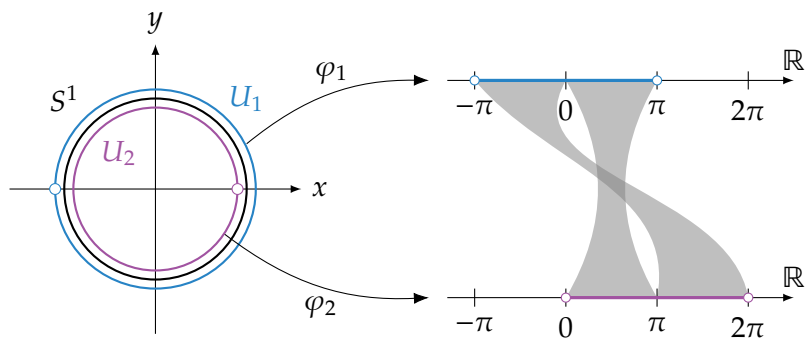
\end{example}

\subsection{Smooth Maps}

An important reason for introducing smooth manifolds is being able to talk about smooth maps. While the terms \emph{map} and \emph{function} are technically interchangeable we will use the term \textbf{function} for maps whose co-domain is $\mathbb{R}$ or $\mathbb{R}^k$ and use \textbf{map} for more general maps between manifolds.

\begin{definition}[Smooth map]
    Let $M$ and $N$ be two smooth manifolds and let $F:M \to N$ be any map. We say $F$ is a \textbf{smooth map} if for every $p \in M$ there exists a smooth chart $\left( U, \varphi \right)$ of $M$ that contains $p$ and a smooth chart $\left( V, \psi \right)$ of $N$ that contains $F(p)$ so that $F(U) \subseteq V$ and the map $\psi \circ F \circ \varphi^{-1}$ is smooth from $\varphi(U)$ to $\psi(F(U)) \subseteq \psi(V)$.
\end{definition}

\begin{figure}[ht!]
    \centering
    \begin{tikzpicture}[scale=1.2]

\def\colora{RoyalBlue!70}
\def\colorb{Plum!70}
\def\colorab{RoyalBlue!50!Plum!80}

% Manifold M
\begin{scope}
\draw[smooth cycle, tension=0.9] 
plot coordinates{(0,0.1) (2.1,0.2) (3.4,0.6) (3.4,2.3) (1.8,2.3) (0.1,2.1)};
\node [] at (0.4, 1.9) {$M$};

\def\U{
plot [smooth cycle, tension=0.7] 
coordinates{(1.0, 1.4) (2.0, 0.7) (2.7, 1.4) (1.8, 2.1)}
}

\fill [\colora] \U;
\draw \U;

\node [] at (1.8, 1.8) {$U$};
\draw [fill] (1.9, 1.3) circle[radius=0.03] node[right, font=\small] {$p$};

\coordinate (M1) at (2.0,0.9);
\coordinate (M2) at (3.2,1.8);
\end{scope}

% Manifold N
\begin{scope}[xshift=6cm]
\draw[smooth cycle, tension=0.9] 
plot coordinates{(0,0.3) (2.0,0.1) (3.5,0.4) (3.6,2.0) (1.8,2.4) (0.0,2.1)};
\node [] at (3.3, 1.9) {$N$};

\def\V{
plot [smooth cycle, tension=0.7] 
coordinates{(0.5, 1.2) (1.7, 0.4) (2.9, 0.7) (3.0, 1.7) (2.1, 2.2)(1.0, 2.1)}
}

\def\FU{
plot [smooth cycle, tension=0.7] 
coordinates{(1.0, 1.4) (2.0, 0.7) (2.7, 1.4) (1.8, 2.1)}
}

\fill [\colorb] \V;
\draw \V;

\fill [\colora] \FU;
\draw \FU;

\node [] at (0.7, 1.4) {$V$};
\node [] at (1.8, 1.8) {$F(U)$};
\draw [fill] (1.9, 1.3) circle[radius=0.03] node[right, font=\small] {$F(p)$};

\coordinate (N1) at (1.8,0.6);
\coordinate (N2) at (0.2,1.8);
\end{scope}

% Chart A
\begin{scope}[xshift=0.5cm, yshift=-4cm]
	
	\def\AU{
	plot [smooth cycle, tension=0.5] 
	coordinates {(0.6, 0.6) (1.8, 0.5) (1.8, 1.8) (0.6, 1.9)}
	}

	\draw[-latex] (-0.2,0) -- (2.5,0) node[above] {$\mathbb{R}^n$};
	\draw[-latex] (0,-0.2) -- (0,2.5);

	\fill [\colora] \AU;
	\draw \AU;
	
	\draw [fill] (1.1, 0.9) circle [radius=0.03] node [right] {$\varphi(p)$};

	\coordinate (A1) at (1.3, 1.6);
	\coordinate (A2) at (1.6, 1.3);
\end{scope}

% Chart B
\begin{scope}[xshift=6.5cm, yshift=-4cm]

	\def\BV{
	plot [smooth cycle, tension=0.5] 
	coordinates {(0.3, 0.3) (1.9, 0.4) (1.9, 2.0) (0.4, 1.9)}
	}

	\def\BU{
	plot [smooth cycle, tension=0.5] 
	coordinates {(0.4, 0.5)(0.9, 0.3) (1.7, 0.5) (1.7, 1.5) (0.6, 1.5)}
	}

	\draw[-latex] (-0.2,0) -- (2.5,0) node[above] {$\mathbb{R}^n$};
	\draw[-latex] (0,-0.2) -- (0,2.5);

	\fill [\colorb] \BV;
	\draw \BV;

	\fill [\colora] \BU;
	\draw \BU;
	
	\draw [fill] (1.1, 0.9) circle [radius=0.03] node [below] {$\psi(F(p))$};

	\coordinate (B1) at (1.3, 1.8);
	\coordinate (B2) at (0.8, 1.3);
\end{scope}

\begin{scope}[]
	\path [{Circle[length=0.8mm]}-{Latex[length=2mm]}] 
	(M1) edge[bend right=20] node[anchor=center, left] {$\varphi$} (A1);

	\path [{Circle[length=0.8mm]}-{Latex[length=2mm]}] 
	(N1) edge[bend left=20] node[anchor=center, right] {$\psi$} (B1);
	
	\path [{Circle[length=0.8mm]}-{Latex[length=2mm]}] 
	(M2) edge[bend left=20] node[anchor=center, above] {$F$} (N2);

	\path [{Circle[length=0.8mm]}-{Latex[length=2mm]}] 
	(A2) edge[bend right=20] node[anchor=center, below] {$\psi\circ F \circ \varphi^{-1}$} (B2);
\end{scope}

\end{tikzpicture}
    \caption{A smooth map $F$ between two manifolds and its representation $\psi \circ F \circ \varphi^{-1}$ between chart co-domains.}
    \label{fig:smooth-map}
\end{figure}
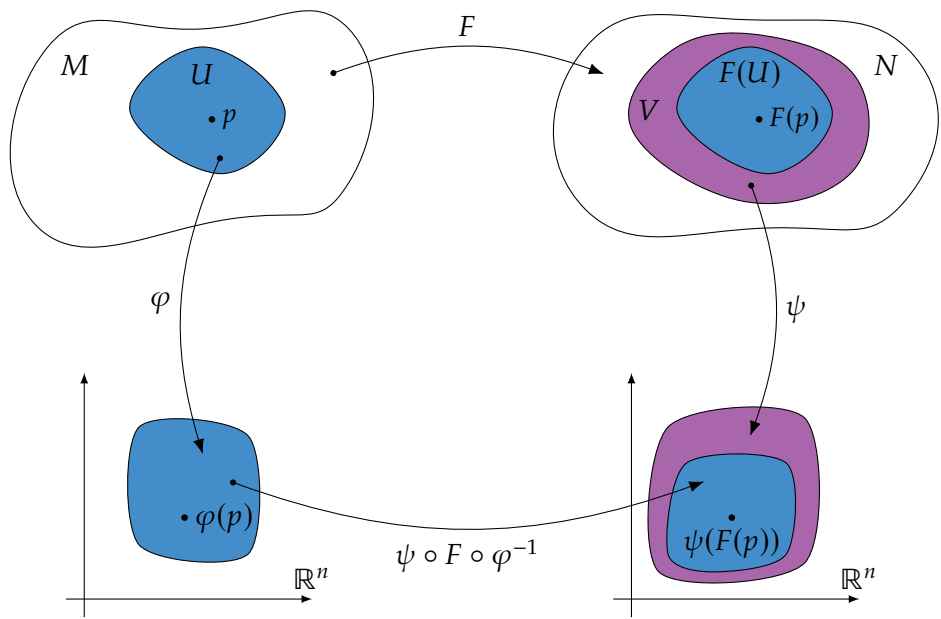

Due to the smooth structure (i.e. all of its possible atlases) that a smooth manifold is equipped with this definition is independent of the choice of charts, the smoothness of transition maps ensures this key property. The definition makes it clear that we can only talk about the smoothness of maps if the manifolds in question have smooth structures, so when we say $F:M\to N$ is a smooth map we imply that $M$ and $N$ are smooth manifolds even if we do not specify that fact explicitly. The construction from the definition is illustrated in Fig.~\ref{fig:smooth-map}.

%\begin{figure}[ht!]
%    \centering
%    \includegraphics[width=0.7\linewidth]{smooth-map.pdf}
%    \caption{A smooth map $F$ between two manifolds and its representation $\psi \circ F \circ \varphi^{-1}$ between chart co-domains.}
%    \label{fig:smooth-map}
%\end{figure}

With \textbf{smooth function} we now just mean a smooth map from $M$ to $\mathbb{R}$ or $\mathbb{R}^k$. Another special case for which we reserve its own term is that of a \textbf{smooth curve}: a smooth map from $\mathbb{R}$ (or some interval of $\mathbb{R}$) to another smooth manifold.

\begin{example}
    The map $F:\mathbb{R} \to S^1$ given by $F(\theta):=\left( \cos\theta,\, \sin\theta  \right)$ is a smooth map from the manifold $\mathbb{R}$ to the manifold $S^1$.
\end{example}

\begin{definition}[Diffeomorphism]
    Let $M$ and $N$ be smooth manifolds, a smooth bijective map from $M$ to $N$ that has a smooth inverse is called a \textbf{diffeomorphism}. Two manifolds between which a diffeomorphism exists are said to be \textbf{diffeomorphic}.
\end{definition}

\begin{example}
    The unit circle $S^1$ is not diffeomorphic with $\mathbb{R}$ since there exists no continuous bijection between the two.
    
    The unit circle is however diffeomorphic with the group $\iident{SO}(2)$ (i.e. the group of orthogonal $2 \times 2$ matrices with determinant 1) via the identification
    \begin{equation*}
        S^1 \ni
        (x,y) 
        \leftrightarrow
        \begin{bmatrix}
            x & -y
            \\
            y & x
        \end{bmatrix}
        \in
        \iident{SO}(2)
        ,
    \end{equation*}
    usually parametrized with $\theta \in \mathbb{R}/(2\pi\mathbb{Z})$ as
    \begin{equation*}
        S^1 \ni
        (\cos\theta,\sin\theta) 
        \leftrightarrow
        \begin{bmatrix}
            \cos\theta & -\sin\theta
            \\
            \sin\theta & \cos\theta
        \end{bmatrix}
        \in
        \iident{SO}(2)
        .
    \end{equation*}
\end{example}

%%%%%%%%%%%%%%%%%%%%%%%%%%%%%%%%%%%%%%%%%%%%%%%%%%%%%%%%%%%%%%%%%%%%%%%%%%%%%%%%%
\section{Lie Groups}

\subsection{Basic Definitions}

We assume that the transformations we are interested in form Lie groups.
Classical groups such as the general and special linear groups (the groups of matrices that are invertible resp. have determinant 1), the orthogonal group (the group of orthogonal matrices), etc. are examples of Lie groups.

\begin{definition}
A \emph{Lie group} G is a smooth manifold so that G is also an algebraic group given by two smooth maps, one for the group product (also called multiplication):
\begin{equation*}
    G \times G \to G, \quad (g_1,g_2) \mapsto g_1 g_2,
\end{equation*}
and one for inversion:
\begin{equation*}
    G \to G, \qquad\quad g \mapsto g^{-1}.
\end{equation*}
\end{definition}

Recall that being a group, a Lie group has the following properties.
\begin{itemize}
    \item \textbf{Closure}: $\forall g_1, g_2 \in G: g_1 g_2 \in G.$
    \item \textbf{Associativity}: $\forall g_1,g_2,g_3 \in G: (g_1 g_2) g_3=g_1 (g_2 g_3)$.
    \item \textbf{Unit element}: $\exists e \in G, \forall g \in G: e g=g e=g$ and this element $e$ is unique. We use the traditional $e$ to denote the unit element, which derives from the German \emph{Einselement}.
    \item \textbf{Inverse}: $\forall g \in G \ \exists g^{-1} \in G: gg^{-1}=g^{-1}g=e$.
\end{itemize}
We should emphasize that a group need not be commutative, indeed the particular Lie groups we are most interested in are not commutative and have group elements $g_1,g_2$ for which $g_1 g_2 \neq g_2 g_1$.

For $g \in G$, we denote by $L_g:G \to G$, $L_g(h)=gh$ the \emph{left multiplication} by $g$ and by $R_g:G \to G$, $R_g(h)=hg$ the \emph{right multiplication} by $g$. 
Left and right multiplication are also sometimes called left and right translation.

\begin{example}
    $\mathbb{R}^n$ is a (commutative) Lie group under vector addition $(\bm{x},\bm{y})\mapsto \bm{x}+\bm{y}$ and negation $\bm{x} \mapsto -\bm{x}$.
\end{example}

\begin{example}[Multiplicative group of positive real numbers]
    $\mathbb{R}_{> 0}$ is a (commutative) Lie group under multiplication $(x,y) \mapsto x y$ and inversion $x \mapsto \sfrac{1}{x}$.
\end{example}

\begin{example}[General linear group]
    The \emph{general linear group} of degree $n$, $\iident{GL}(n)$, is the group of all invertible $n \times n$ matrices. The group product is the matrix product.
\end{example}

\begin{example}[Special orthogonal group]
    The \emph{special orthogonal group} of degree $n$, $\iident{SO}(n)$, is the subgroup of $\iident{GL}(n)$ of all matrices with determinant $1$. 
    It is the group of rotations in $n$ dimensions. 
    For $SO(2)$ we write the matrices in terms of the angle of the rotation $\theta \in \mathbb{R}/(2\pi\mathbb{Z})$ as
    \begin{equation}
        \label{eq:so2_matrix}
        R(\theta) = 
        \begin{bmatrix}
        \cos\theta & -\sin\theta
        \\
        \sin\theta & \cos\theta
        \end{bmatrix}
        ,
    \end{equation}
    with the group law given by $R(\theta_1) R(\theta_2)=R(\theta_1 + \theta_2)$.
\end{example}

\begin{example}[Special Euclidean group]
    \label{example:special_euclidean_group}
    The \emph{special Euclidean group} of degree $n$, $\iident{SE}(n)$, is the group of rotations and translations in $n$ dimensions. As a set it equals $\mathbb{R}^n \times \iident{SO}(n)$ with the group law given by:
    \begin{equation}
        \label{eq:se_law}
        \left( \bm{x}_1, R_1 \right)
        \left( \bm{x}_2, R_2 \right)
        =
        \left( \bm{x}_1 + R_1 \bm{x}_2,\, R_1 R_2 \right)
        .
    \end{equation}
    The group law is not the direct product (i.e. not $\left( \bm{x}_1 + \bm{x}_2,\, R_1 R_2 \right)$) but the second group affects the law of the first group. 
    We call this a \emph{semidirect product}, 
    to emphasize this difference we write $\iident{SE}(n)=\mathbb{R}^n \rtimes \iident{SO}(n)$.
    
    Since we want to design a neural network that takes in 2D images and is rotation-translation equivariant, the Lie group $\iident{SE}(2)$ is of special interest to us.
    In the $\iident{SE}(2)$ case we can either represent elements as $(\bm{x},R(\theta)) \in \mathbb{R}^2 \times \iident{SO}(2)$ with $R(\theta)$ the rotation matrix from \eqref{eq:so2_matrix}, or as $(\bm{x},\theta) \in \mathbb{R}^2 \times [0,2\pi)$. 
    In the latter case the group law can be written as
    \begin{equation*}
        (\bm{x}_1,\theta_1)
        (\bm{x}_2,\theta_2)
        =
        \left(
            \bm{x}_1 + R(\theta_1) \bm{x}_2
            ,\,
            \theta_1 + \theta_2 \bmod{2\pi}
        \right)
        .
    \end{equation*}
\end{example}

\subsection{Lie Subgroups}

Algebraic groups can have subgroups (think of the many subgroups of the general linear group).
Lie groups also have subgroups but those subgroups are not automatically Lie groups themselves.
Let $G$ be a Lie group, then $H \subset G$ is a \emph{Lie subgroup} of $G$ if
\begin{enumerate}[label=(\roman*)]
    \item $H$ is a subgroup of the group $G$,
    \item $H$ is an immersed submanifold of the manifold $G$,
    \item the group operations on $H$ are smooth.
\end{enumerate}
We have not seen what an immersed submanifold is and this is beyond the scope of this course. However the following theorem provides an easier way of identifying most Lie subgroups.

\begin{theorem}[Cartan’s closed subgroup theorem]
    Any subgroup of a Lie group that is closed (as a set) is a Lie subgroup.
    \label{thm:closed_subgroup}
\end{theorem}
See \textcite[Ch. 7]{lee2013smooth} for proof and details.

Not all Lie subgroups are closed but the ones we are interested in all are. 
Consequently Theorem~\ref{thm:closed_subgroup} is our go-to method for proving whether a subgroup is a Lie group.

\begin{example}
    Every linear subspace of $\mathbb{R}^n$ is a (closed) Lie subgroup under vector addition.
\end{example}

\begin{example}
    The groups $\mathbb{R}^n \times \{ 0 \}$ and $\{ \bm{0} \} \times \iident{SO}(n)$ are (closed) Lie subgroups of $\iident{SE}(n)$.
\end{example}

\subsection{Group Actions}

The most important use of Lie groups in manifold theory involves the action of a Lie group on a manifold.

\begin{definition}[Group action]
\label{def:group_action}
If $G$ is a group and $M$ a set then a \emph{left action} of $G$ on $M$ is a map $G \times M \to M$ written as $(g,p) \mapsto g \cdot p$ that satisfies:
\begin{equation}
\begin{alignedat}{2}
        g_2 \cdot (g_1 \cdot p)
        &=
        (g_2 g_1) \cdot p
        \qquad
        &&\forall g_1,g_2 \in G, p \in M,
        \\
        e \cdot p &= p
        &&
        \forall p \in M.
\end{alignedat}
\label{eq:left_action}
\end{equation}
A \emph{right action} is defined similarly as a map $M \times G \to M$ that satisfies:
\begin{equation*}
\begin{alignedat}{2}
        (p \cdot g_1) \cdot g_2
        &=
        p \cdot (g_1 g_2)
        \qquad
        &&\forall g_1,g_2 \in G, p \in M,
        \\
        p \cdot e &= p
        &&
        \forall p \in M.
\end{alignedat}
\end{equation*}
If both $G$ and $M$ are manifolds and the map is smooth in both inputs $G$ and $M$ then we say the action is a \emph{smooth action}.
\end{definition}

We will focus exclusively on left actions since their group law \eqref{eq:left_action} has the property that group multiplication corresponds to map composition.
In any case a right action can always be converted to a left action by defining $g \cdot p := p \cdot g^{-1}$, or vice versa for turning a left action into a right action.

In our setting $G$ is always a Lie group and $M$ always a manifold and we will only be considering smooth actions.

Sometimes it is convenient to label an action, say $\rho:G \times M \to M$. The action of a group element $g$ on a point $p$ can then be written equivalently as $g \cdot p \equiv  \rho(g,p) \equiv \rho_g(p)$.

If $\rho$ is a smooth action then for all $g \in G$, $\rho_g : M \to M$ is a diffeomorphism since $\rho_{g^{-1}}$ is a smooth inverse.

A group action on a manifold induces an action on any function space on that manifold in a straightforward manner.
Let $X$ be a function space on $M$, such as $C^k(M)$ or $L^p(M)$, then the mapping $\rho^X:G \times X \to X$, defined by
\begin{equation}
    \label{eq:induced_action}
    \left(\rho^X (g, f)\right)(p) 
    = 
    f(g^{-1}\cdot p)
    =
    f(\rho(g^{-1},p))
    ,
\end{equation}
for all $f \in X$ and $g \in G$ is a (left) action.
We can verify that with
\begin{align*}
    &
    \rho^X \left(
        g_2
        ,\,
        \rho^X (g_1,\, f)
    \right)(p)
    \\
    &=
    \rho^X(g_1,\,f)(g_2^{-1} \cdot p)
    \\
    &=
    f \left( g_1^{-1} \cdot (g_2^{-1} \cdot p) \right)
    \\
    &=
    f \left( (g_1^{-1} g_2^{-1}) \cdot p \right)
    \\
    &=
    f \left( (g_2 g_1)^{-1} \cdot p \right)
    \\
    &=
    \rho^X (g_2 g_1,\, f)(p)
    .
\end{align*}

This action on function spaces has the additional property that it is linear in the second argument. We call actions with this property representations of the Lie groups.

\begin{definition}[Lie group representation]
    Let $G$ be a Lie group and $V$ a vector space (finite dimensional or not) then
    $\nu: G \to \rident{Aut}(V)$ 
    is a \emph{representation} of $G$ if it is a smooth homomorphism, i.e. is smooth and
    \begin{equation*}
        \nu(g_1 g_2) = \nu(g_1) \circ \nu(g_2)
        \qquad
        \forall g_1,g_2 \in G
        .
    \end{equation*}
    Recall that the automorphism group Aut($V$) is the group of invertible linear transformations of $V$.
\end{definition}
It follows from the definition that a representation $\nu$ also has the following properties:
\begin{equation*}
    \nu(e) = \mathrm{id}_V 
    \quad\text{and}\quad 
    \nu(g^{-1}) = \nu(g)^{-1}.
\end{equation*}

Since we usually only have one group action per manifold, and so one corresponding representation on a given function space we can overload the meaning of the ``$\cdot$'' symbol and use the following equivalent notations:
\begin{equation*}
    g \cdot f := \rho^X_g(f) := \rho^X(g, f)
    .
\end{equation*}

This is how we are going to be modeling transformation acting on our data.
In our rotation-translation case $f$ would be an input image on $\mathbb{R}^2$ and $g \in \iident{SE}(2)$ would be a rotation-translation acting on the image.

\subsection{Equivariant Maps and Operators}

Suppose $G$ is a Lie group and $M$ and $N$ are smooth manifolds with smooth (left) actions $\rho^M$ and $\rho^N$.
Then we can consider maps $F:M \to N$ that are \emph{equivariant} with respect to those group actions, i.e.
\begin{equation*}
    F(\rho^M(g,p)) = \rho^N (g,F(p))
\end{equation*}
for all $g \in G$ and $p \in M$, or more concisely:
\begin{equation*}
    F(g \cdot p) = g \cdot F(p).
\end{equation*}
Equivalently, $F$ is equivariant if the following diagram commutes for each $g \in G$:
\[\begin{tikzcd}
	M & N \\
	M & N.
	\arrow["F", from=1-1, to=1-2]
	\arrow["F", from=2-1, to=2-2]
	\arrow["{\rho^M_g}"', from=1-1, to=2-1]
	\arrow["{\rho^N_g}", from=1-2, to=2-2]
\end{tikzcd}\]

This idea extends naturally to operators between function spaces on those manifolds. 
Let $X$ be a function space on $M$ and $Y$ a function space on $N$ equipped with the corresponding representations $\rho^X$ and $\rho^Y$ per \eqref{eq:induced_action}. 
Then an operator $A:X \to Y$ is equivariant if 
\begin{equation}
\label{eq:equivariant_operator}
A \circ \rho^X_g = \rho^Y_g \circ A, \qquad \forall g \in G.
\end{equation}
Or in words: for every group element doing the corresponding transform on the input space $X$ and then applying the operator $A$ gives the same results as first applying the operator $A$ and then performing the transform corresponding to the group element on the output space $Y$.

Our goal in the continuous setting is designing our neural network as an equivariant operator that satisfies \eqref{eq:equivariant_operator}.

%%%%%%%%%%%%%%%%%%%%%%%%%%%%%%%%%%%%%%%%%%%%%%%%%%%%%%%%%%%%%%%%%%%%%%%%%%%%%%%%%
\subsection{Homogeneous Spaces}

While Lie groups represent the transformations we are interested in, homogeneous spaces are the spaces our data will live on and on which the Lie groups will act.

\begin{definition}[Homogeneous space]
    A smooth manifold $M$ is a \emph{homogeneous space} of a Lie group $G$ if there exists a smooth (left) action $\rho:G \times M \to M$ that is \emph{transitive}, i.e.
    \begin{equation*}
    \forall p_1,p_2 \in M \ \exists g \in G: \rho(g,p_1)=g \cdot p_1=p_2.
    \end{equation*}
\end{definition}

The elements of $M$ are called the points of the homogeneous space and $G$ is called the motion group or the fundamental group of the homogeneous space. The transitive property can be reformulated as: for every point in $M$ there is a $g$ that takes us to any other point in $M$.

Observe that $G$ is a homogeneous space of itself, called the \emph{principal homogeneous space}. The group action is just the left multiplication, i.e. $\rho_g = L_g$.

On the other end we have the \emph{trivial homogeneous space} consisting of a single element $\{ 0 \}$, which is a homogeneous space of every Lie group under the identity action $\rho(g,0)=0$.

\begin{remark}[Zero-dimensional manifolds]
    You may wonder whether $\{0\}$ is a manifold. 
    In fact all (at most) countable sets $S$ are $0$-dimensional manifolds. 
    Assign each point $p \in S$ its own unique chart $\varphi_p:\{ p \} \to \mathbb{R}^0=\{ 0 \}$.
    Since these chart domains do not overlap the charts are trivially smoothly compatible and form a unique smooth atlas.
\end{remark}

For each $p \in M$, the \emph{stabilizer} or the \emph{isotropy group} of $p$ is the subset $G_p$ of $G$ (also denoted by $\rident{Stab}_G(p)$) that fixes $p$:
\begin{equation}
    \label{eq:stabilizer}
    G_p 
    :=
    \rident{Stab}_G(p) 
    :=
    \left\{
        g \in G \ \middle\vert\  g \cdot p = \rho(g,p) = p
    \right\}
    .
\end{equation}
If we have $g_1,g_2 \in G_p$ then $(g_1 g_2) \cdot p = g_1 \cdot (g_2 \cdot p)=g_1 \cdot p=p$. 
So $g_1 g_2 \in G_p$, meaning that $G_p$ is a subgroup of $G$.
Moreover since the group action is smooth it follows that if $\left( g_n \right)_{n\in\mathbb{N}}$ is a sequence in $G_p$ with $\lim_{n\to\infty} g_n = g \in G$ then
\begin{equation*}
    \rho(g,p)
    =
    \rho\left( \lim_{n\to\infty} g_n, p \right)
    =
    \lim_{n\to\infty} \rho(g_n,p)
    =
    p.
\end{equation*}
From which we conclude that $g \in G_p$ and so $G_p$ is closed and consequently by Theorem~\ref{thm:closed_subgroup}, $G_p$ is a Lie subgroup of $G$ for all $p \in M$.

When we pick a \emph{reference element} $p_0 \in M$, we can define the subset $G_{p_0,p} \subset G$ of all group elements that map $p_0$ to $p$:
\begin{equation}
    G_{p_0,p} := \left\{ g \in G \ \middle\vert \ g \cdot p_0 = p \right\}
    .
    \label{eq:homogeneous_set}
\end{equation}
Note that this is generally not a subgroup, except for $G_{p_0,p_0}=G_{p_0}$. If we have two group elements that map $p_0$ to the same $p$, i.e. $g_1,g_2 \in G_{p_0,p}$ then 
\begin{equation}
\label{eq:homogeneous_equivalence}
g_1 \cdot p_0 = g_2 \cdot p_0 
\quad \Leftrightarrow\quad 
g_1^{-1} g_2 \cdot p_0 = p_0
\quad \Leftrightarrow\quad 
g_1^{-1} g_2 \in G_{p_0}.
\end{equation}
This condition imposes an equivalence relation on $G$, which we can quotient out as follows:
\begin{equation*}
    G/G_{p_0} 
    := 
    \left\{ 
        s \subset G 
        \ \middle\vert\  
        \forall g_1, g_2 \in s: g_1^{-1} g_2 \in G_{p_0}
    \right\}
    =
    \left\{
        G_{p_0,p}
        \ \middle\vert\
        \forall p \in M
    \right\}
    .
\end{equation*}
There is a straightforward isomorphism between $M$ and $G/G_{p_0}$ given by $p \mapsto G_{p_0,p}$ and $G_{p_0,p} \mapsto G_{p_0,p} \cdot p_0$.

From which we can conclude that all homogeneous spaces are isomorphic to a Lie group quotient $G/H$ for some closed Lie subgroup $H$ of $G$.
For this reason many authors blur the line between a homogeneous space and its corresponding group quotient and effectively equate a point of the homogeneous space with its corresponding equivalence class in the group, i.e. $p \equiv G_{p_0,p}$ after fixing a $p_0 \in M$. This leads to concise notation such as $g \in p \Leftrightarrow g \cdot p_0 = p$ and the dropping of the `$\cdot$' notation since if $p$ is seen as a subset of $G$ then $g \cdot p \equiv g p$.

%%%%%%%%%%%%%%%%%%%%%%%%%%%%%%%%%%%%%%%%%%%%%%%%%%%%%%%%%%%%%%%%%%%%%%%%%%%%%%%%%
\section{Linear Operators}

Now let us look at how we can start putting together an artificial neuron in our new setting. 
We have an input manifold $M$ and an output manifold $N$ that are both homogeneous spaces of a Lie group $G$. 
Our input data is a function on $M$, say $f \in X = \iident{B}(M)$ and we are expected to output a function on $N$, say an element of $Y = \iident{B}(N)$.
Recall that the set of bounded functions $\iident{B}(M)$ is a Banach space under the supremum norm (a.k.a. the $\infty$-norm or the uniform norm) given by $\| f \|_\infty := \sup_{p \in M} |f(p)|$. 

The first part of a discrete artificial neuron was a linear operator $A:\mathbb{R}^n \to \mathbb{R}^m$, given by:
\begin{equation*}
        \left( A \bm{x} \right)_i 
        = 
        \sum_{j} (A)_{ij} x_i.
\end{equation*}
Its analogue in the continuous setting is an \emph{integral operator} $A:\mathbb{R}^M \to \mathbb{R}^N$ of the form:
\begin{equation}
    \label{eq:integral_operator}
        \left( A f \right)(q)
        =
        \int_M k_A (p,q) \, f(p) \, \d p
        ,
\end{equation}
where the function $k_A:M \times N \to \mathbb{R}$ is called the operator's \emph{kernel}.
%If $k_A \in \iident{B}(M \times N)$ and  $p \mapsto k_A(p,q)$ is compactly supported for all $q \in N$, then the integral exists for all $q \in N$ and so $A$ is a well defined linear operator from $\iident{B}(M)$ to $\iident{B}(N)$.
%Note that not all linear operators are of this type, but since we plan to discretize in the end anyway this is a sufficiently broad class of linear operators.

\begin{remark}[Measurable functions]
    Technically for the Lebesgue integral in \eqref{eq:integral_operator} to exist the integrand needs to be measurable. 
    We will not be dealing with non-measurable functions and you may assume that when we say function we mean measurable function. 
    If the concept of measurable functions is new to you, you may ignore the issue.
\end{remark}

In this framework, instead of training the matrix $A$, we will train the kernel $k_A$. In practice we cannot train a continuous function so training the kernel will come down to either training a discretization or training the parameters of some parameterization of $k_A$.

Now we still need to specify how we are going to integrate on a homogeneous space to make progress.

\subsection{Integration}

Integration on $\mathbb{R}^n$ has the desirable property that it is translation invariant: for all $\bm{y} \in \mathbb{R}^n$ and integrable functions $f:\mathbb{R}^n \to \mathbb{R}$ we have
\begin{equation}
    \label{eq:translation_invariant_integral}
    \int_{\mathbb{R}^n} f(\bm{x}-\bm{y}) \, \d\bm{x}
    =
    \int_{\mathbb{R}^n} f(\bm{x}) \, \d\bm{x}
    .
\end{equation}
Ideally we would want integration on a homogeneous space $M$ of a Lie group $G$ to behave similarly, namely for all $g \in G$ we would like:
\begin{equation}
    \label{eq:group_invariant_integral}
    \int_M \left( g \cdot f \right)(p) \, \d\mu_M(p)
    :=
    \int_M f(g^{-1} \cdot p) \, \d\mu_M(p)
    =
    \int_M f(p) \, \d\mu_M(p),
\end{equation}
for some Radon measure $\mu_M$ on $M$.

\begin{remark}[Measures]
    Recall that measures are the generalization of concepts such as length, volume, mass, probability etc. A measure assigns a non-negative real number to subsets of a space in such a way that it behaves similarly to the aforementioned concepts.
    A Radon measure on a Hausdorff topological space is a measure that plays well with the topology of the space (defined for open and closed sets, finite on compact sets, etc.).
    The Lebesgue measure is the translation invariant Radon measure on $\mathbb{R}^n$ and coincides with our less general notion of the length/area/volume of subsets of $\mathbb{R}^n$.
    Integration on $\mathbb{R}^n$ such as in \eqref{eq:translation_invariant_integral} implicitly uses the Lebesgue measure and so is translation invariant.
    For a comprehensive introduction to measure theory see \textcite{tao2011introduction}. For the purpose of this course it is sufficient to think about a measure as measuring the volume of a subset.
\end{remark}

This imposes a condition on the measure $\mu_M$, namely: for all measurable subsets $S$ of $M$ and $g \in G$ we require 
\begin{equation}
\label{eq:invariant_measure}
\mu_M(g\cdot S)=\mu_M(S). 
\end{equation}
In other words we would need a (non-zero) group invariant measure to get the desired integral.
These G-invariant measures, or just \emph{invariant measures}, do not always exist. In some cases we can still obtain a \emph{covariant measure}, which is a measure that satisfies 
\begin{equation}
\label{eq:covariant_measure}
\mu(g \cdot S)=\chi(g)\,  \mu(S)
,
\end{equation}
where $\chi:G \to \mathbb{R}^+$ is a character of $G$. 

\begin{definition}[Character]
\label{def:character}
A multiplicative character or linear character or simply \emph{character} of a Lie group $G$ is a continuous homomorphism from the group to the multiplicative group of positive real numbers, i.e. $\chi:G \to \mathbb{R}_{>0}$ so that:
\begin{equation*}
    \chi(g_1 g_2) = \chi(g_1)\, \chi(g_2) \qquad \forall g_1,g_2 \in G.
\end{equation*}
\end{definition}
The function $\chi$ needs to be a character since by \eqref{eq:covariant_measure} we have:
\begin{equation*}
    \chi(g_1 g_2)\, \mu(S)
    =
    \mu(g_1 g_2 \cdot S)
    =
    \mu(g_1 \cdot (g_2 \cdot S))
    =
    \chi(g_1) \, \mu(g_2 \cdot S)
    =
    \chi(g_1) \, \chi(g_2) \, \mu(S)
    ,
\end{equation*}
for all $g_1,g_2 \in G$ and all measurable $S \subset M$.

If we integrate with respect to a G-invariant measure we say we have a G-invariant integral, or just an \emph{invariant integral}, if the measure is covariant with a character $\chi$ we say we have a $\chi$-covariant integral or just \emph{covariant integral}.

\begin{definition}[Covariant integral]
    \label{def:covariant_integral}
    Let $M$ be a homogeneous space of a Lie group $G$, we say the integral $\int_M \ldots \d p$ (using some Radon measure on $M$) is \emph{covariant} with respect to $G$ if there exists a character $\chi_M$ of $G$ so that 
    \begin{equation*}
    \int_M \left( g \cdot f \right) (p) \, \d p
    =
    \chi_M(g) \, \int_M f(p) \,  \d p
    \end{equation*}
    for all $g \in G$ and all $f:M \to \mathbb{R}$ for which the integral exists.
    In the special case that $\chi_M \equiv 1$ we say the integral is \emph{invariant}.
\end{definition}

\begin{remark}[Abuse of notation]
Integration is always with respect to some measure.
If we are integrating with respect to the measure $\mu$ then for the sake of completeness we should write
\begin{equation*}
    \int_M \ldots\ \d \mu(p)
    .
\end{equation*}
But since we only ever consider one measure per space we integrate over and for the sake of brevity we abbreviate $\d p \equiv \d\mu(p)$.
\end{remark}

If the homogeneous space is $G$ itself then an invariant measure is called the (left) \emph{Haar measure} on $G$ (named after the Hungarian mathematician Alfréd Haar). We can say \textit{the} Haar measure since Haar measures are unique up to multiplication with a constant and always exist \parencite[see][Ch. 2.7]{federer2014geometric}. 
Hence when integrating on the group itself we can always have a Haar measure $\mu_G$ so that the following equality holds
\begin{equation}
    \label{eq:invariant_group_integral}
    \int_G \left( h \cdot f \right)(g) \, \d g 
    =
    \int_G f(g) \, \d g
    \qquad
    \forall h \in G
    ,
\end{equation}
where we abbreviated $\d g := \d\mu_G(g)$.
We also call this invariant integral on the group the (left) \emph{Haar integral}.

Not all homogeneous spaces admit a covariant integral but those in which we are interested all do.
Going forward we will assume that all homogeneous spaces that we consider admit a covariant integral and that we can always use the equality from Definition~\ref{def:covariant_integral}.

\begin{example}[$G=\iident{SE}(2)$ and $M=\mathbb{R}^2$]
In the case we are most interested in, namely $G=\iident{SE}(2)$ and $M=\mathbb{R}^2$, we are fortunate that the Lebesgue measure on $\mathbb{R}^2$ is invariant with respect to $G$.
This is intuitively easy to understand: the area of a subset of $\mathbb{R}^2$ is invariant under both translation and rotation. 
\end{example}

\begin{example}[Haar measure on $\iident{SE}(2)$]
The Haar measure on $\iident{SE}(2)$ also conveniently coincides with the Lebesgue measure on $\mathbb{R}^2 \times [0,2\pi)$ when using the parameterization from Example~\ref{example:special_euclidean_group}.
Indeed, let $g=(\bm{x}_1,\theta_1)$ and $h=(\bm{x}_2,\theta_2)$ then:
\begin{equation*}
    \int_{\mathbb{R}^2} \int_{0}^{2\pi} 
    \left( (\bm{x}_1,\theta_1) \cdot f \right) (\bm{x}_2,\theta_2)
    \, \d\theta_2 \,\d\bm{x}_2
    =
    \int_{\mathbb{R}^2} \int_{0}^{2\pi} 
    f \left( (\bm{x}_1,\theta_1)^{-1}  (\bm{x}_2,\theta_2) \right)
    \, \d\theta_2 \,\d\bm{x}_2 
    .
\end{equation*}
When we change variables to $(\bm{x}_3,\theta_3)=(\bm{x}_1,\theta_1)^{-1}  (\bm{x}_2,\theta_2)$ we obtain the following Jacobian matrix:
\begin{equation*}
    \frac{\partial(x_2^1,x_2^2,\theta_2)}{\partial(x_3^1,x_3^2,\theta_3)}
    =
    \begin{pmatrix}
        \cos\theta_1 & -\sin\theta_1 & 0
        \\
        \sin\theta_1 & \cos\theta_1 & 0
        \\
        0 & 0 & 1
    \end{pmatrix}
    ,
\end{equation*}
which has determinant $1$.
Consequently, the Haar integral (up to a multiplicative constant) on $\iident{SE}(2)$ can be calculated as:
\begin{equation}
    \int_{\iident{SE}(2)} f(g) \, \d g
    =
    \int_{\mathbb{R}^2}
    \int_0^{2\pi} 
    f(\bm{x},\theta)
    \,
    \d\theta
    \d\bm{x}
    .
\end{equation}
\end{example}

\subsection{Equivariant Linear Operators}
\label{subsection:equivariant_linear_operators}

Of course the objective of this chapter is building equivariant operators, so when is an integral operator \eqref{eq:integral_operator} equivariant? 
Equivariance means that
\begin{equation*}
    A (g \cdot f) = g\cdot(A f) 
\end{equation*}
for all $g \in G$ and $f \in \iident{B}(M)$ or equivalently 
\begin{equation}
    \label{eq:equivariant_A_2}
    g^{-1} \cdot A (g \cdot f) = A f
    .
\end{equation}
This extra condition on $A$ will naturally impose some restrictions on the kernel of the operator as the following lemma shows.

\begin{lemma}[Equivariant linear operators]
    \label{lem:equivariant_integral_operator}
    Let $M$ and $N$ be homogeneous spaces of a Lie group $G$ so that $M$ admits a covariant integral with character $\chi_M$.
    Let $A$ be an integral operator \eqref{eq:integral_operator} from $\iident{C}(M) \cap \iident{B}(M)$ to $\iident{C}(N) \cap \iident{B}(N)$ with a kernel $k_A \in \iident{C}(M \times N)$.
    Then 
    \begin{equation*}
    A(g\cdot f) = g \cdot (A f)
    \end{equation*}
    for all $g \in G$ and $f \in \iident{C}(M) \cap \iident{B}(M)$ if and only if
    \begin{equation}
    \label{eq:equivariant_kernel_symmetry}
        \chi_M(g) \, k_A(g \cdot p,g \cdot q)
         =
         k_A(p,q)
    \end{equation}
    for all $g \in G$, $p \in M$ and $q \in N$.
    
    Moreover $A$ is bounded (and so continuous) in the supremum norm if
    \begin{equation}
        \label{eq:kernel_boundedness_requirement}
        \sup_{q \in N} \int_{M} |k_A(p,q)| \d p < \infty
        .
    \end{equation} 
\end{lemma}

\begin{proof}
\ \\[5pt]
``$\Rightarrow$''

Assuming $A$ to be equivariant, take an arbitrary $g \in G$ and $f \in \iident{C}(M) \cap \iident{B}(M)$ and substitute the definition of the group representation and $A$ in \eqref{eq:equivariant_A_2} to find
\begin{equation}
    \label{eq:ELO_1}
    \int_M k_A(p,g \cdot q) \,  f (g^{-1} \cdot p) \, \d p
    =
    \int_M k_A(p,q) \, f(p) \,\d p
\end{equation}
for all $q \in N$.

Fix $q \in N$ and let $F(p):=k_A(g \cdot p, g \cdot q) f(p)$ then observe that
\begin{equation*}
    (g \cdot F)(p)
    =
    k_A(g \cdot g^{-1} \cdot p, g \cdot q) f(g^{-1} \cdot p)
    =
    k_A(p,g \cdot q) \,  f (g^{-1} \cdot p)
    ,
\end{equation*}
which is the left integrand from \eqref{eq:ELO_1}.
Since we have assumed covariant integration we use Definition~\ref{def:covariant_integral} and have
\begin{equation*}
    \int_M \left( g \cdot F \right) (p) \, \d p
    =
    \chi_M(g) \, \int_M F(p) \,  \d p
    .
\end{equation*}
Applying this to \eqref{eq:ELO_1} we find
\begin{equation}
    \label{eq:ELO_2}
    \chi_M(g)
    \int_M k_A(g \cdot p,g \cdot q) \,  f (p) \, \d p
    =
    \int_M k_A(p,q) \, f(p) \,\d p
    .
\end{equation}
Since $f$ was arbitrary and $p \mapsto k_A(p,q)$ continuous it follows that
\begin{equation*}
     \chi_M(g) \, k_A(g \cdot p,g \cdot q)
     =
     k_A(p,q)
\end{equation*}
for all $p \in M$.

``$\Leftarrow$''

Assuming $\chi_M(g) \, k_A(g \cdot p,g \cdot q)=k_A(p,q)$ for all $g \in G$, $p \in M$ and $q \in N$ then \eqref{eq:ELO_2} follows for any choice of $f \in \iident{C}(M) \cap \iident{B}(M)$, $g \in G$ and $q \in N$.
Substituting the covariant integral the other way yields 
\eqref{eq:ELO_1}, which implies \eqref{eq:equivariant_A_2} since $q \in N$ is arbitrary.
The function $f$ and group element $g$ were also chosen arbitrarily so \eqref{eq:equivariant_A_2} follows for all $f \in \iident{C}(M) \cap \iident{B}(M)$ and $g \in G$.

Boundedness of $A$ follows from
\begin{align*}
    \left\Vert A f \right\Vert_{\infty}
    &=
    \sup_{q \in N} 
    \left| 
        \int_M k_A(p,q) \, f(p) \d p
    \right|
    \\
    &\leq
    \sup_{q \in N}
    \int_M
    | k_A(p,q) | \, |f(p)| \d p
    \\
    &\leq
    \left\Vert f \right\Vert_\infty
    \cdot\ 
    \sup_{q \in N}
    \int_M | k_A(p,q) | \d p
    \\
    &
    \overset{\eqref{eq:kernel_boundedness_requirement}}{<} \infty
    .
\end{align*}

\end{proof}

The condition on the kernel \eqref{eq:kernel_boundedness_requirement} is partially redundant with the symmetry requirement as the following lemma shows.

\begin{lemma}
    \label{lem:kernel_L1}
    In the same setting as Lemma~\ref{lem:equivariant_integral_operator}. If the kernel $k_A \in \iident{C}(M \times N)$ satisfies the symmetry \eqref{eq:equivariant_kernel_symmetry} and condition \eqref{eq:kernel_boundedness_requirement} then
    \begin{equation*}
        \left\Vert k_A(\ \cdot\ ,q_1) \right\Vert_{L^1(M)}
        =
        \left\Vert k_A(\ \cdot\ ,q_2) \right\Vert_{L^1(M)}
    \end{equation*}
    for all $q_1,q_2 \in N$.
\end{lemma}

\begin{proof}

Since $N$ is a homogeneous space then for all $q_1,q_2 \in N$ there exists a $g \in G$ so that $q_1 = g \cdot q_2$, then
\begin{align*}
    \int_M \left| k_A(p,q_1) \right| \d p
    &=
    \int_M \left| k_A(p,g \cdot q_2) \right| \d p
    \\
    &=
    \int_M \left| k_A(g \cdot g^{-1} \cdot p, g \cdot q_2) \right| \d p
    \\
    {\footnotesize \eqref{eq:equivariant_kernel_symmetry}}
    &=
    \frac{1}{\chi_M(g)}
    \int_M \left| k_A(g^{-1} \cdot p, q_2) \right| \d p
    \\
    \text{\footnotesize (Def.~\ref{def:covariant_integral})}
    &=
    \frac{\chi_M(g)}{\chi_M(g)}
    \int_M \left| k_A(p, q_2) \right| \d p
    \\
    &=
    \int_M \left| k_A(p,q_2) \right| \d p
    .
\end{align*}

\end{proof}

The condition on the kernel from Lemma~\ref{lem:equivariant_integral_operator} can be exploited to express it as a function on $M$ instead of $M \times N$.
If we fix a $q_0 \in N$ and for all $q \in N$ we choose a $g_q \in G_{q_0,q}$ (i.e. so that $g_q \cdot q_0=q$) then by \eqref{eq:equivariant_kernel_symmetry} we have
\begin{equation*}
    k_A(p,q) 
    = 
    \chi_M(g_q^{-1}) \ k_A(g_q^{-1} \cdot p, g_q^{-1} \cdot q)
    =
    \chi_M(g_q^{-1}) \ k_A(g_q^{-1} \cdot p, q_0)
    ,
\end{equation*}
which fixes the second input of $k_A$.
Consequently we could contain all the information of our kernel in a function that exists only on $M$ as $\kappa_A(p) := k_A(p,q_0)$.
This reduced kernel $\kappa_A$ still has some restrictions placed on it for the resulting operator to be equivariant, as the following theorem makes precise.

\begin{theorem}[Equivariant linear operators]
    \label{thm:equivariant_linear_operators}
    Let $M$ and $N$ be homogeneous spaces of a Lie group $G$ so that $M$ admits a covariant integral with respect to a character $\chi_M$ of $G$.
    Fix a $q_0 \in N$ and let $\kappa_A \in \iident{C}(M) \cap \iident{L}^1(M)$ be \emph{compatible}, i.e. have the property that
    \begin{equation}
        \label{eq:kernel_compatibility}
        \forall h \in G_{q_0}
        :
        h \cdot \kappa_A = \chi_M (h) \, \kappa_A.
    \end{equation}
    
    Then the operator $A$ defined by
    \begin{equation*}
        (Af)(q) := \frac{1}{\chi_M(g_q)} \int_M (g_q \cdot \kappa_A) (p) \, f(p) \, \d p
    \end{equation*}
    where for all $q \in N$ we can choose any $g_q$ so that $g_q \cdot q_0 = q$, is a \emph{well defined bounded linear operator} from $\iident{C}(M) \cap \iident{B}(M)$ to $\iident{C}(N) \cap \iident{B}(N)$ that is \emph{equivariant} with respect to $G$.
    
    Conversely every equivariant integral operator with a kernel $k_A \in \iident{C}(M\times N)$ and with $k_A(\,\cdot\,,q) \in \iident{L}^1(M)$ for some $q \in N$ is of this form.
\end{theorem}

\begin{proof}
\ \\
``$\Rightarrow$''

Assuming we have a $\kappa_A \in \iident{C}(M) \cap \iident{L}^1(M)$ that satisfies \eqref{eq:kernel_compatibility}.
Define $k_A \in C(M \times N)$ by
\begin{equation*}
    k_A(p,q) := \frac{1}{\chi_M(g_q)} (g_q \cdot \kappa_A)(p).
\end{equation*}
Then $k_A$ is well defined since it does not depend on the choice of $g_q$ for a given $q \in N$.
If $g_q'$ is another group element with $g_q \cdot q_0 = q$ then there exists a $h \in G_{q_0}$ so that $g_q' = g_q h$, we can check $k_A$ is invariant under choice of $h \in G_{q_0}$:
\begin{align*}
    \frac{1}{\chi_M(g_q h)} (g_q \cdot h \cdot \kappa_A)(p)
    =
    \frac{\chi_M(h)}{\chi_M(g_q) \chi_M(h)} (g_q \cdot \kappa_A)(p)
    =
    \frac{1}{\chi_M(g_q)} (g_q \cdot \kappa_A)(p)
    .
\end{align*}
The kernel $k_A$ also satisfies the symmetry requirement \eqref{eq:equivariant_kernel_symmetry} from Lemma~\ref{lem:equivariant_integral_operator}:
\begin{align*}
    \chi_M(g) \, k_A(g \cdot p,g \cdot q)
    &=
    \chi_M(g) \, \frac{1}{\chi_M(g_{(g\cdot q)})} (g_{(g\cdot q)} \cdot \kappa_A)(g \cdot p)
    \\ 
    &=
    \chi_M(g) \, \frac{1}{\chi_M(g g_{q})} (g \cdot g_{q} \cdot \kappa_A)(g \cdot p)
    \\
    &=
    \frac{\chi_M(g)}{\chi_M(g)\chi_M(g_{q})} (g_{q} \cdot \kappa_A)(g^{-1}g \cdot p)
    \\
    &=
    \frac{1}{\chi_M(g_q)} (g_q \cdot \kappa_A)(p)
    \\
    &=
    k_A(p,q)
    .
\end{align*}
By Lemma~\ref{lem:kernel_L1} we have
\begin{equation*}
    \sup_{q \in N} \int_{M} |k_A(p,q)| \d p 
    = 
    \left\Vert k_A(\,\cdot\,,q_0) \right\Vert_{L^1(M)}
    =
    \left\Vert \kappa_A \right\Vert_{L^1(M)}
    < \infty
    .
\end{equation*}
Consequently, $A$ also satisfies \eqref{eq:kernel_boundedness_requirement} and is a bounded equivariant linear operator per Lemma~\ref{lem:equivariant_integral_operator}.

``$\Leftarrow$''
~\ \\~
Assuming we have an equivariant linear operator $A$ with kernel $k_A \in \iident{C}(M \times N)$ then we pick a fixed $q_0 \in N$ and define $\kappa_A \in \iident{C}(M)$
\begin{equation*}
    \kappa_A(p) := k_A(p,q_0)
    .
\end{equation*}
This reduced kernel $\kappa_A$ satisfies the compatibility condition \eqref{eq:kernel_compatibility} since if $h \in G_{q_0}$ then
\begin{align*}
    (h \cdot \kappa_A)(p)
    &=
    k_A(h^{-1} \cdot p, q_0)
    \\
    &=
    k_A(h^{-1} \cdot p, h^{-1} \cdot q_0)
    \\
    &=
    \chi_M(h) \, k_A(p, q_0)
    \\
    &=
    \chi_M(h) \, \kappa_A(p)
    .
\end{align*}
Since we required $k_A(\,\cdot\,,q) \in \iident{L}^1(M)$ for some $q \in N$, we apply Lemma~\ref{lem:kernel_L1} to find
\begin{equation*}
    \left\Vert \kappa_A \right\Vert_{L^1(M)}
    =
    \left\Vert k_A(\ \cdot\ ,q_0) \right\Vert_{L^1(M)}
    =
    \left\Vert k_A(\ \cdot\ ,q) \right\Vert_{L^1(M)}
    <
    \infty.
\end{equation*}

\end{proof}

Theorem~\ref{thm:equivariant_linear_operators} is the at the core of group equivariant CNNs since it allows us to generalize the familiar convolution operation present in CNNs to general linear operators that are equivariant with respect to a group of choice.

\begin{example}[Group convolution]
    \label{example:group_convolution}
    Let $G=M=N$ be some Lie group.
    A Lie group always admits a Haar integral, so we have a trivial character $\chi=1$.
    As reference element we obviously choose the unit element $e$, though any group element would do.
    Then $G_g = \{ e \}$ and $G_{e,g}=\{ g \}$ are both trivial.
    Hence we have no symmetry condition on the kernel.
    Any $\kappa_A \in C(G) \cap L^1(G)$ defines a linear operator $A: C(G) \cap B(G) \to C(G) \cap B(G)$ by
    \begin{equation*}
        (Af)(h)
        =
        \int_G (h \cdot \kappa_A)(g) \, f(g) \, \d g
        =
        \int_G \kappa_A (h^{-1} g) \, f(g) \, \d g
    \end{equation*}
    We also call this operation \emph{group cross-correlation} and denote it as
    \begin{equation*}
        (\kappa \star_G f)(h)
        :=
        \int_G (h \cdot \kappa)(g) \, f(g) \, \d g
        .
    \end{equation*}
    As in the familiar $\mathbb{R}^n$ setting, group cross-correlation is closely related to \emph{group convolution}, which is defined as
    \begin{equation*}
        (\check{\kappa} *_G f)(h)
        :=
        \int_G \check{\kappa} (g^{-1} h) \, f(g) \, \d g
        .
    \end{equation*}
    We leave relating the two kernels $\kappa$ and $\check{\kappa}$ as an exercise: when is $\kappa \star_G f = \check{\kappa} *_G f$?

    As in the $\mathbb{R}^n$ case, when we talk about group convolution we mean both group cross-correlation and group convolution since they are interchangeable.
\end{example}

\begin{example}[Rotation-translation equivariance in $\mathbb{R}^2$]
    \label{example:rototranslation_equivariance_r2}
    
    Let $G=\iident{SE}(2)= \mathbb{R}^2 \rtimes \iident{SO}(2)$ and $M=N=\mathbb{R}^2$.
    The Lebesgue measure on $\mathbb{R}^2$ is rotation-translation invariant so we have a G-invariant integral on $\mathbb{R}^2$.
    Choose $\bm{y}_0=\bm{0}$ as the reference element then $G_{\bm{y}_0}=\left\{ (\bm{0},\, R(\theta)) \in G \ \middle\vert\  \theta \in [0,2\pi) \right\}$ is the stabilizer of $\bm{y}_0$.
    A kernel $\kappa_A$ on $\mathbb{R}^2$ is then compatible if
    \begin{equation*}
        (\bm{0},\, R(\theta)) \cdot \kappa_A = \kappa_A
        \qquad
        \forall \theta \in [0,2\pi)
        ,
    \end{equation*}
    i.e. $\kappa_A$ needs to be radially symmetric.
    Now, we could have figured that out without building up the whole equivariance framework. 
    But the next section will show how we can use the equivariance framework to step over the severe restriction that is imposed on the allowable kernels here.
\end{example}

\section{Building a Rotation-translation Equivariant CNN}

How do we now use this framework to construct a rotation-translation invariant network for images?
From Example~\ref{example:rototranslation_equivariance_r2} we know that we do not have a lot of freedom in training a rotation-translation equivariant operator from $\mathbb{R}^2$ to $\mathbb{R}^2$. 
We can buy ourselves a lot more freedom by making the first linear operation in our network one that maps functions on $\mathbb{R}^2$ to functions on $\iident{SE}(2)$.
In the context of image analysis the process of transforming an image to a higher dimensional representation is called \emph{lifting}. 
Therefore we will call the first layer in our network the \emph{lifting layer}.
Once we are operating on the group we have much more freedom since group convolution is equivariant. 
Just like in a conventional CNN we can have a series of \emph{group convolution layers} that make up the bulk of our network.
Of course we might not want our final product to live on the group, we might want to go back to $\mathbb{R}^2$, but we also have a recipe for that. Since going from the 3 dimensional space $\iident{SE}(2)$ to the 2 dimensional space $\mathbb{R}^2$ is akin to projection we call a layer that does this a \emph{projection layer}.

\begin{figure}[ht!]
    \centering
    \includegraphics[width=1.0\textwidth]{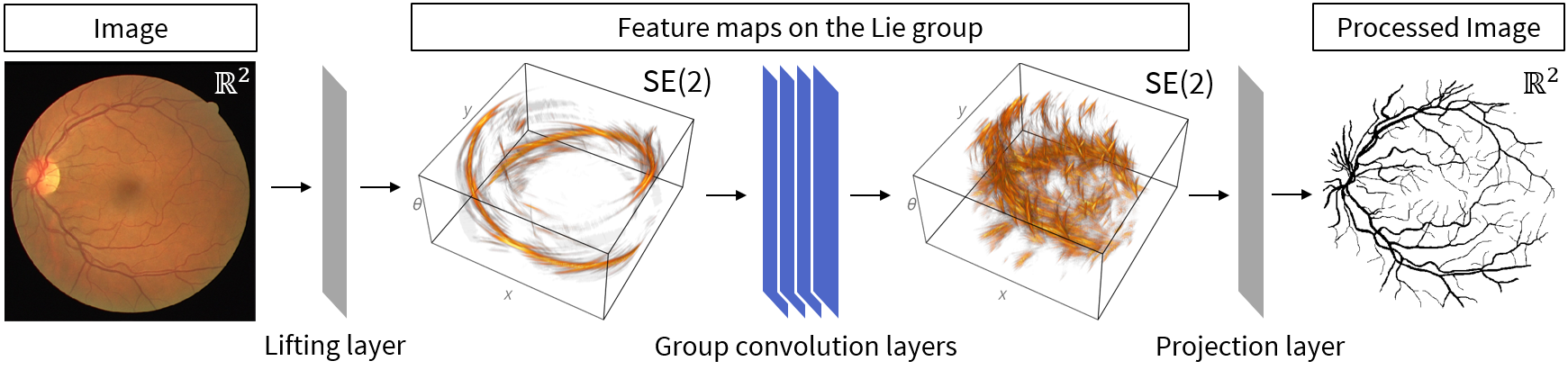}
    \caption{A G-CNN for retinal vessel segmentation that is rotation-translation equivariant. Lifting data to a higher dimensional space affords us more freedom in the kernels we train while maintaining equivariance.}
    \label{fig:drive_cnn}
\end{figure}

This three stage design is illustrated in Figure~\ref{fig:drive_cnn} for the retinal vessel segmentation application. For some more examples of the use of this type of network in medical imaging applications see \textcite{bekkers2018roto}.

\subsection{Lifting Layer}

Let $G=\iident{SE}(2) \equiv \mathbb{R}^2 \rtimes [0, 2\pi)$ using the parameterization from Example~\ref{example:special_euclidean_group}. Let $M=\mathbb{R}^2$ and $N=G$.
Choose $e=(\bm{0},0) \in N$ as the reference element then the stabilizer $G_e$ is trivially $\{e\}$, so any kernel on $M=\mathbb{R}^2$ is compatible.
The Lebesgue measure is rotation-translation equivariant so we have an invariant integral.

Let $n_0 \in  \{ 1,3 \}$ be the number of input channels and denote the input functions as $f^{(0)}_{j}:\mathbb{R}^2 \to \mathbb{R}$ for $j$ from $1$ to $n_0$. Let us denote the number of desired feature maps in the first layers as $n_1$.
Recall that there are two conventions for convolution layers: single channel and multi channel. 

In the multi channel setup we associate with each output channel a number of kernels equal to the amount of input channels. 
Our parameters would be a set $\{ \kappa^{(1)}_{ij}\}_{ij} \subset C(\mathbb{R}^2) \cap L^1(\mathbb{R}^2)$ of kernels and a set $\{ b^{(1)}_i \}_{i} \subset \mathbb{R}$ of biases for $i$ from $1$ to $n_1$ and $j$ from $1$ to $n_0$. 
The calculation for output channel $i$ is then given by
\begin{equation*}
    f^{(1)}_{i}(\bm{x},\theta)
    =
    \sigma \left(
    \sum_{j=1}^{n_0}
    \int_{\mathbb{R}^2} \left( (\bm{x},\theta) \cdot \kappa^{(1)}_{ij}\right) (\bm{y}) \, f^{(0)}_{j}(\bm{y}) \, \d \bm{y} + b^{(1)}_{i}
    \right)
    ,
\end{equation*}
for some choice of activation function $\sigma$.

In the single channel setup we associate a kernel with each input channel and then make linear combinations of the convolved input channels to generate output channels.
Our parameters would then consist of a set of kernels $\{ \kappa^{(1)}_{j} \}_{j} \subset C(\mathbb{R}^2) \cap L^1(\mathbb{R}^2)$ and a set of weights $\{ a^{(1)}_{ij} \}_{ij} \subset \mathbb{R}$ and biases $\{ b^{(1)}_i \}_{i} \subset \mathbb{R}$ for $i$ from $1$ to $n_1$ and $j$ from $1$ to $n_0$.
The calculation for output channel $i$ is then given by
\begin{equation*}
    f^{(1)}_{i}(\bm{x},\theta)
    =
    \sigma \left(
    \sum_{j=1}^{n_0}
    a^{(1)}_{ij}
    \int_{\mathbb{R}^2}
    \left( (\bm{x},\theta) \cdot \kappa^{(1)}_{j} \right)(\bm{y})
    \, f^{(0)}_{j}(\bm{y}) \, \d\bm{y}
    + b^{(1)}_{i}
    \right)
    .
\end{equation*}
In either case the actual lifting happens by translating and rotating the kernel over the image, a particular translation and rotation gives us a particular scalar value at the corresponding location in $\iident{SE}(2)$.

\begin{remark}[Orientation score transform]
    If you followed the course \textit{Differential Geometry for Image Processing (2MMA70)} this will seem very familiar to you. 
    Indeed this an \emph{orientation score transform} except we do not design the wavelet filter (the kernel) ourselves but leave it up to the network to train.
\end{remark}

\subsection{Group Convolution Layer}

We already saw in Example~\ref{example:group_convolution} how to do convolution on the group itself.
On the Lie group we always have an invariant integral (the left Haar integral) and the symmetry requirement on the kernel is trivial so we have no restrictions to take into account for training the kernel (unlike for the case $\mathbb{R}^2 \to \mathbb{R}^2$).
In layer $\ell \in \mathbb{N}$ with $n_{\ell-1}$ input channels and $n_{\ell}$ output channels the calculation for output channel $i$ is given (for the single channel setup) by 
\begin{equation*}
    f^{(\ell)}_{i} 
    =
    \sigma \left(
    \sum_{j=1}^{n_{\ell-1}}
    a^{(\ell)}_{ij}
    \left(
        \kappa^{(\ell)}_{j} 
        \star_G
        f^{(\ell-1)}_{j}
    \right)
    +
    b^{(\ell)}_{i}
    \right)
\end{equation*}
for all $i \in \{1, \ldots, n_{\ell} \}$ or
\begin{equation*}
    f^{(\ell)}_{i} (\bm{x},\theta)
    =
    \sigma \left(
    \sum_{j=1}^{n_{\ell}}
    a^{(\ell)}_{ij}
    \int_{\mathbb{R}^2}
    \int_{0}^{2\pi}
    \left( (\bm{x},\theta) \cdot \kappa^{(\ell)}_{j} \right)
    (\bm{y},\alpha)
    \ 
    f^{(\ell-1)}_{j}(\bm{y},\alpha)
    \ 
    \d\alpha \d\bm{y}
    +
    b^{(\ell)}_{i}
    \right)
    ,
\end{equation*}
for some choice of activation function $\sigma$. Here the kernels $\kappa^{(\ell)}_{j} \in C(G)\times L^1(G)$, the weights $a^{(\ell)}_{i j} \in \mathbb{R}$ and the biases $b^{(\ell)}_i \in \mathbb{R}$ are the trainable parameters. 
Deducing the formula for the multi channel setup we leave as an exercise.

Group convolution layers can be stacked sequentially just like normal convolution layers in a CNN to make up the heart of a G-CNN, see Figure~\ref{fig:drive_cnn}.

\subsection{Projection}

The desired output of our network is likely not a feature map on the group or some other higher dimensional homogeneous space. 
So at some point we have to transition away from them.

In traditional CNNs used for classification we saw that at some point we flattened our multi-dimensional array by `forgetting' the spatial dimensions. 
Once we have discretized, flattening is of course also a viable approach for a G-CNN when the goal is classification. However we might not want to throw away our spatial structure, if the goal of the network is to transform its input in some way then we want to go back to our original input space (like in the example in Figure~\ref{fig:drive_cnn}).

Applying our equivariance framework again to the case $G=M=\iident{SE}(2)$ and $N=\mathbb{R}^2$. 
Choose $\bm{0} \in N$ as the reference element then its stabilizer is the subgroup of just rotations. 
So to construct an equivariant linear operator from $\iident{SE}(2)$ to $\mathbb{R}^2$ requires a kernel $\kappa$ on $\iident{SE}(2)$ that satisfies
\begin{equation*}
    (\bm{0},\, \beta) \cdot \kappa = \kappa
    \quad\Leftrightarrow\quad
    \kappa(\bm{x},\,\theta)
    =
    \kappa\left( R(-\beta) \bm{x},\, \theta - \beta \right)
    \qquad
    \forall \beta,\theta \in [0,2\pi),\, \bm{x} \in \mathbb{R}^2
    ,
\end{equation*}
where $R(-\theta)$ is the rotation matrix over $-\theta$. 
Consequently we can reduce the trainable (unrestricted) part of the kernel $\kappa$ to a 2 dimensional slice:
\begin{equation*}
    \kappa(\bm{x},\theta)
    =
    \kappa\left( R(-\theta)\bm{x},\, 0 \right)
    .
\end{equation*}
A kernel like this gives us the desired equivariant linear operator and with a set of them we can construct a layer in the same fashion as the lifting and group convolution layer.

In practice this type of operator with trainable kernel is not what is used for projection from $\iident{SE}(2)$ to $\mathbb{R}^2$. 
Instead the much simpler (and non-trainable) integration over the $\theta$ axis is used, let $f \in \iident{B}(\iident{SE}(2))$ then the operator $P:C(\iident{SE}(2)) \cap B(\iident{SE}(2)) \to C(\mathbb{R}^2) \cap B(\mathbb{R}^2)$ given by
\begin{equation}
    \label{eq:integration_over_theta}
    (Pf)(\bm{x})
    :=
    \int_{0}^{2\pi} f(\bm{x},\theta) \, \d\theta,
\end{equation}
is a bounded linear operator that is rotation-translation equivariant.

\begin{remark}
    Note that the projection operator \eqref{eq:integration_over_theta} is one of our equivariant linear operators if we take the kernel to be $\kappa_P(\bm{x},\theta)=\delta(\bm{x})$ where $\delta$ is the Dirac delta on $\mathbb{R}^2$. The Dirac delta is not a function in $C(G) \cap L^1(G)$ but we can take a sequence in $C(G) \cap L^1(G)$ that has $\kappa_P$ as the limit in the weak sense, such as a sequence of narrowing Gaussians.
\end{remark}

A common alternative to integrating over the orientation axis is taking the maximum over that axis:
\begin{equation}
    \label{eq:max_projection_layer}
    \left( P_{\max} f \right)(\bm{x})
    :=
    \max_{\theta \in [0,2\pi)}
    \,
    f(\bm{x},\theta).
\end{equation}
This is not a linear operator but it is rotation-translation equivariant, we will revisit this projection operator later.

After we have once again obtained feature maps on $\mathbb{R}^2$ we can proceed to our desired output format in the same way as we would do with a classic CNN. Either we forget the spatial dimensions and transition to a fully connected network for classification applications or we take a linear combination of the obtained 2D feature maps to generate an output image such as in Figure~\ref{fig:drive_cnn} and \textcite{bekkers2018roto}.

\subsection{Discretization}

\begin{figure}[ht]
    \centering
    \begin{subfigure}[t]{0.48\textwidth}
        \vskip 0pt
        \centering
        \includegraphics[width=\textwidth]{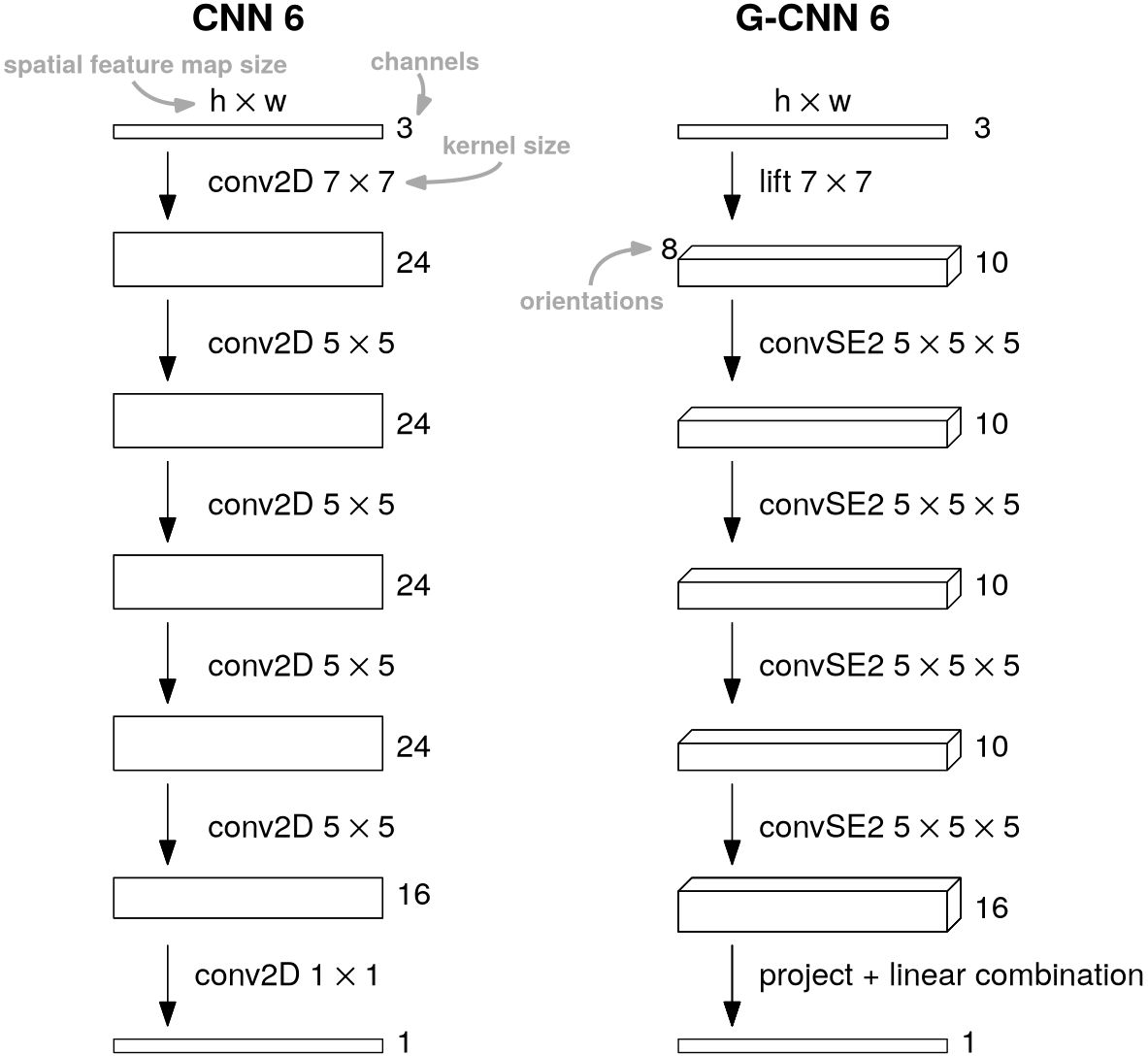}
        \vspace{2pt}
        \caption{A traditional CNN (left) versus a rotation-translation equivariant G-CNN (right) for segmenting a $h \times w$ color image. Figure~\ref{fig:drive_cnn} shows an application of this type.
        }
        \label{fig:drive_models}
    \end{subfigure}
    \hfill
    \begin{subfigure}[t]{0.48\textwidth}
        \vskip 0pt
        \centering
        \includegraphics[width=\textwidth]{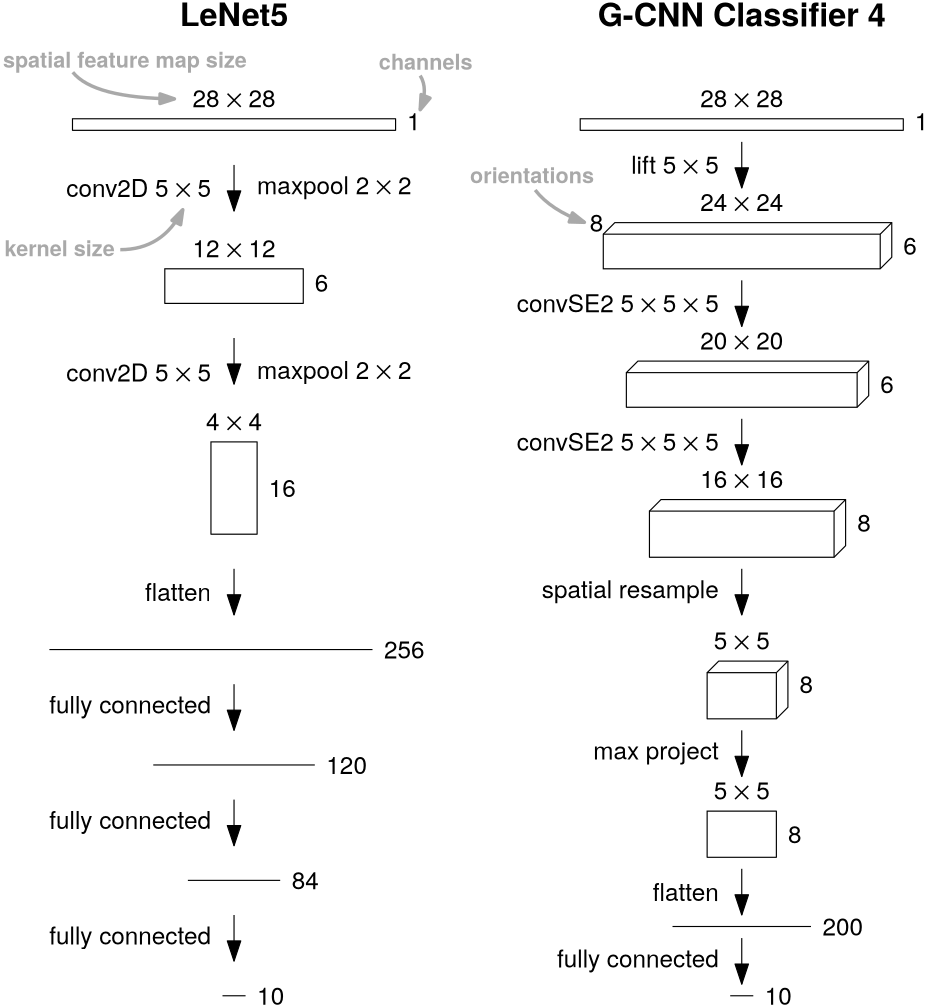}
        \vspace{2pt}
        \caption{A traditional CNN (left) versus a rotation-translation equivariant G-CNN (right) for classifying $28 \times 28$ grayscale images into $10$ classes. Digit classification falls into this category.}
        \label{fig:mnist_models}
    \end{subfigure}  
    \caption{Example $\iident{SE}(2)$ G-CNNs architectures versus similar traditional CNN architectures. 
    The shapes show the size of the data tensors at each stage in the network.
    }
    \label{fig:gcnn_models}
\end{figure}

To implement our developed G-CNN in practice we will need to switch to a discretized setting.
For our specific case of an $\iident{SE}(2)$ G-CNN the lifting layer typically uses kernels of size $5 \times 5$ to $7 \times 7$. 
We usually choose the number of discrete orientations to be $8$, so an input of $\mathbb{R}^{H \times W}$ would be lifted to $\mathbb{R}^{8\times H \times W}$, this may seem low but empirically this is around the sweet-spot between performance and memory usage/computation time.
The group convolution layers usually employ $5 \times 5 \times 5$ kernels.
In both cases we need to sample the kernel off-grid to be able to rotate them, for that we almost always use linear interpolation.
Example G-CNN implementations are illustrated in Figure~\ref{fig:gcnn_models} for both segmentation and classification, examples for various medical applications can be found in \textcite{bekkers2018roto}.
% See \textcite[Table 1]{bekkers2018roto} or \textcite[Figure 15 \& 17]{smets2021pdebased} for example architectures.

As a general rule of thumb in deep learning we discretize as coarsely as we can get away with. 
Increasing the size of the kernels or the number of orientations does increase performance but nowhere near proportional to the increase in memory and computation time this causes.
Keeping coarse kernels and increasing the depth of the network is a better way of spending a given memory/time budget.

\begin{remark}[Linear interpolation alternatives]
Higher order polynomial interpolation methods are highly undesirable on such coarse grids, as the oscillations can make the network behave erratically.
More advanced interpolation techniques have been proposed, see for example \textcite{bekkers2021bspline}, but the added computationally complexity can be a drawback.
Just as with discretization, the rule of thumb for interpolation is: as coarsely as you can get away with.
\end{remark}

%%%%%%%%%%%%%%%%%%%%%%%%%%%%%%%%%%%%%%%%%%%%%%%%%%%%%%%%%%%%%%%%%%%%%%%%%%%%%%%%%
\section{Tropical Operators}

We previously generalized the idea of a convolution to that of equivariant linear operators in a Lie group setting, essentially generalizing the domain of our functions. 
But we can go further by looking at the codomain and generalizing what we mean by `linear'.

Linear and non-linear are generally thought of as an absolute dichotomy. 
Another way of thinking about a linear function or operator is as a map that preserves the algebraic properties of the field of real numbers.
But could we not look at maps that preserve the structure of other algebraic structures rather than the field of real numbers?

We could in principle develop this idea using any algebraic structure, but to be useful in the context of neural networks we want the underlying set to be numeric and the operations to be able to be performed by a computer.

Additionally, restricting ourselves to fields would be very limiting, a more congenial algebraic structure is the semiring.
Semirings turn up in many places, indeed the first mathematical structure we learn about, the set of natural numbers $\mathbb{N}$, is a semiring.
Semirings arise in many areas of mathematics such as functional analysis, topology, graphs, combinatorics, optimization, etc. 
See \textcite{golan1999semirings} for an extensive survey of the applications of semirings.

\subsection{Semirings}

A semiring is an algebraic structure in which we can add and multiply elements, but in which neither subtraction nor
division are necessarily possible. 

\begin{definition}[Semiring]\label{def:semiring}
    A \emph{semiring} is a set $R$ equipped with two binary operations $\oplus$ and $\odot$, called addition and multiplication, such that
    \begin{enumerate}[label=(\roman*)]
        \item addition and multiplication are associative,
        \item addition is commutative,
        \item addition has an identity element $\mathbb{0}$,
        \item multiplication has an identity element $\mathbb{1}$,
        \item multiplication distributes over addition:
            \begin{equation*}
                \begin{split}
                    a \odot (b \oplus c) = a \odot b \oplus a \odot c,
                    \\
                    (a \oplus b) \odot c = a \odot c \oplus b \odot c,
                \end{split}
            \end{equation*}
        \item multiplication by $\mathbb{0}$ annihilates:
            \begin{equation*}
                \mathbb{0} \odot a = a \odot \mathbb{0} = \mathbb{0}.
            \end{equation*}
    \end{enumerate}
Additionally if $a \oplus a = a$ we say the semiring is \emph{idempotent} and if $a \odot b = b \odot a$ we say the semiring is \emph{commutative} or \emph{abelian}.
\end{definition}

Just like with standard multiplication it is conventional to abbreviate $a b \equiv a \odot b$ if there can be no confusion. We also let $\odot$ take precedence over $\oplus$, i.e. $a \odot b \oplus c = (a \odot b) \oplus c$.

\begin{remark}[Rig]
    Some works refer to semirings as \emph{rigs}, from rings `without negatives', hence the missing `n'.
\end{remark}

\begin{example}[Real linear semiring]
The real numbers form a commutative semiring $(\mathbb{R},+, \,\cdot\,)$ under standard addition and multiplication. 
Naturally all fields and rings are also semirings.
\end{example}

\begin{example}[Viterbi semiring]
    \label{example:viterbi_semiring}
    The Viterbi semiring is given by the unit interval $[0,1]$ and the operations
    \begin{equation*}
        \begin{split}
        a \oplus b &:= \max\{a,b\},
        \\
        a \odot b &:= a b
        .
        \end{split}
    \end{equation*}
    The additive unit is $0$ and multiplicative unit is $1$. This semiring is both idempotent and commutative.
    Note that there exists no element $(-a)$ so that $a \oplus (-a) = \mathbb{0}=0$ and no element $a^{-1}$ so that $a \odot a^{-1} = \mathbb{1}=1$. So neither subtraction nor division is possible in this semiring.
\end{example}

\begin{example}[Log semiring]
    The log semiring is given by the set $\mathbb{R} \cup \{ -\infty,\,+\infty\}$ and the operations
        \begin{equation*}
        \begin{split}
        a \oplus b &:= \log \left( e^a + e^b \right),
        \\
        a \odot b &:= a + b.
        \end{split}
    \end{equation*}
    The additive unit is $-\infty$ and the multiplicative unit is $0$.
\end{example}

The central object of study in linear algebra is the vector space over a field (usually $\mathbb{R}$ or $\mathbb{C}$).
The analogue to the vector space in our generalized setting is the semimodule.

\begin{definition}[Semimodule]
    Let $(R,\oplus,\odot)$ be a semiring.
    A left $R$-semimodule or \emph{semimodule} over $R$ is a commutative monoid $(M,+)$ with additive identity $0_M$ and a map $R \times M \to M$ denoted by $(a,m) \mapsto a m$, called \emph{scalar multiplication}, so that for all $a,b \in R$ and $m,m' \in M$ the following hold:
    \begin{enumerate}[label=(\roman*)]
        \item $(a \odot b) m = a (b m)$,
        
        \item $a(m+m')=a m + a m'$,
        
        \item $(a \oplus b)m=a m + b m$,
        
        \item $\mathbb{1} m = m$,
        
        \item $a 0_M = 0_M = \mathbb{0} m$.
    \end{enumerate}
\end{definition}
Naturally every semiring is a semimodule over itself just like every field is a vector space over itself.

\begin{example}
Let $R$ be a semiring and $S$ a non-empty set, then $R^S$ (the set of all functions $S \to R$) is a semimodule with addition and scalar multiplication defined element-wise:
let $f,f' \in R^S$ and $r \in R$ then
\begin{equation*}
    (f \oplus f')(s)
    :=
    f(s) \oplus f'(s)
    \quad
    \text{and}
    \quad
    (r \odot f)(s) 
    := 
    r \odot f(s)
\end{equation*}
for all $s \in S$. Here we used $\oplus$ and $\odot$ for the operations in the semimodule as well since they correspond with the $\oplus$ and $\odot$ operations in the semiring.
The additive identity of the semimodule is the constant function $s \mapsto \mathbb{0}$.
This generalizes the prototypical vector space $\mathbb{R}^n$ to the semimodule $R^n$ using the notational convention $n\equiv\{1,\ldots,n\}$.
For non-finite $S$ we get the generalization of function vector spaces to function semimodules.
\end{example}

\begin{example}
    Let $R=([0,1],\max,\,\cdot\,)$ be the Viterbi semiring from Example~\ref{example:viterbi_semiring} and let $S$ be a manifold.
    Let $f,f': S \to [0,1]$ and define addition and scalar multiplication as
    \begin{equation*}
        (f + f')(s)
        :=
        \max \left\{ f(s) ,\, f'(s)\right\}
        \quad
        \text{and}
        \quad
        (r f)(s)
        :=
        r f(s),
    \end{equation*}
    then the functions from $S$ to $[0,1]$ form a semimodule.
\end{example}

Now we can formulate a generalization of linear maps in the form of semiring homomorphisms.

\begin{definition}[Semimodule homomorphism]
    Let $R$ be a semiring and let $X$ and $Y$ be semimodules over $R$.
    Then a map $A:X \to Y$ is an \emph{$R$-homomorphic map} or an \emph{$R$-homomorphism} if for all $a,b \in R$ and $f,f' \in X$:
    \begin{equation*}
        A(a f + b f')
        =
        a (A f) + b (A f')
        ,
    \end{equation*}
    where on the left the addition and scalar multiplication happen in $X$ and on the right the addition and scalar multiplication happen in $Y$.
\end{definition}
Just like with the definition of linearity the single condition above is equivalent to the following two conditions:
\begin{equation*}
    A(a f) = a (A f')
    \quad
    \text{and}
    \quad
    A(f + f')
    =
    A f + A f'
    \qquad
    \forall a \in R ,\, f,f' \in X.
\end{equation*}

Under this definition we can understand linear as meaning homomorphic with respect to the real linear semigroup $(\mathbb{R},+,\,\cdot\,)$.
But now, instead of just considering linear maps we can pick another semigroup and consider homomorphisms with respect to this semigroup.
This allows us to construct equivariant semimodule homomorphic operators in the same fashion as we did with equivariant linear operators in section~\ref{subsection:equivariant_linear_operators}. 
We will develop such a class of equivariant operators for a particular choice of semiring.

\subsection{Tropical Semiring}

\begin{definition}[Tropical semiring]\label{def:tropical_semiring}
    The max tropical semiring or max-plus algebra or simply \emph{tropical semiring} is the semiring $\mathbb{R}_{\max}:=\left(\mathbb{R}
    \cup \{-\infty\} ,\oplus,\odot \right)$, with the operations
    \begin{equation*}
        \begin{split}
        a \oplus b &:= \max\{a, b\}
        ,
        \\
        a \odot b &:= a + b.
        \end{split}
    \end{equation*}
    Where the additive identity is $\mathbb{0}:=-\infty$ and the multiplicative identity is $\mathbb{1}:=0$.
\end{definition}

Since $a \oplus a = \max \{ a, a \} = a$ and $a \odot b = a+b=b+a = b \odot a$ the tropical semiring is both idempotent and commutative.
By definition we set $a + (-\infty) := -\infty$ so that we satisfy the annihilation property $a \odot \mathbb{0} = \mathbb{0} \odot a = \mathbb{0}$.

\begin{remark}
    The tropical semiring can alternatively be defined as $(\mathbb{R} \cup \{ +\infty \}, \min, +)$, which is then also called the min tropical semiring or min-plus algebra.
    But we observe that one is isomorphic to the other via negation so we make a style choice and go with the max version.
\end{remark}

\begin{example}[ReLU]
    Recall that the rectified linear unit is defined as $x \mapsto \{ x,\, 0\}$ for $x \in \mathbb{R}$. 
    In the tropical setting we can write this as $x \mapsto x \oplus 0$ for $x \in \mathbb{R}_{\max}$, hence the ReLU may not be a linear or affine function but it is \emph{tropically affine}.
    So we can think about a typical ReLU neural network as really alternating operations from two distinct semirings.
\end{example}

We based our construction of equivariant linear operators on integration. 
When you consider an integral as a limit of a sum of products we can see how we could define a type of integration with respect to another semigroup.

Recall that Riemannian integration is defined in terms of Darboux sums over ever smaller partitions of the underlying space.
Let $M$ be a manifold with some Radon measure $\mu$ and $(R,\oplus,\odot)$ a semiring. 
Let $P$ be a partition of $M$ and define $\mu(P) := \sup_{S \in P} \mu(S)$. Then we can generalize a Darboux sum

\begin{minipage}{\textwidth}
\begin{equation*}
    \everymath{\displaystyle}
    \tikzstyle{every node}=[anchor=base,rounded corners=1mm]
    \lim_{\mu(P) \to 0}\ 
    \tikz[baseline]{
        \node[fill=RoyalBlue!20] (1)
        {$\sum_{S \in P}$};
    }
        \,
        f(p_S)
        \,
        \tikz[baseline]{
        \node[fill=Plum!20] (2)
        {$\cdot\vphantom{\vert}$};
        }
        \,
        \mu(S)
    ,
\end{equation*}
as:
\begin{equation}
    \label{eq:generalized_darboux_sum}
    \everymath{\displaystyle}
    \tikzstyle{every node}=[anchor=base,rounded corners=1mm]
    \lim_{\mu(P) \to 0}\ 
    \tikz[baseline]{
        \node[fill=RoyalBlue!20] (3)
        {$\bigoplus_{S \in P}$};
    }
        \,
        f(p_S)
        \,
        \tikz[baseline]{
        \node[fill=Plum!20] (4)
        {$\odot\vphantom{\vert}$};
        }
        \,
        \mu(S)
    ,
\begin{tikzpicture}[overlay]
    \path[-latex, very thick, color=RoyalBlue!50] (1) edge [bend left=15] (3);
    \path[-latex, very thick, color=Plum!50] (2) edge [bend right=20] (4);
\end{tikzpicture}
\end{equation}
\end{minipage}

where $p_S \in M$ is any arbitrary point in the partition element $S$ and assuming that for all $\varepsilon>0$ we can find a partition $P$ so that $\mu(P)<\varepsilon$.
Filling in the tropical semiring operations we obtain:
\begin{align}
    \nonumber
    & 
    \lim_{\mu(P) \to 0}\ 
    \max_{S \in P}
    \left(
       f(p_S)
        +
        \mu(S)
    \right)
    \intertext{($\mu(S) \to 0$ since $\mu(P) \to 0$)}
    &=
    \lim_{\mu(P) \to 0}\ 
    \max_{S \in P}
       f(p_S)
      \label{eq:lim_max_f}
    \intertext{(Assuming $f$ is such that the limit exists and is unique)}
    \nonumber
    &=
    \sup_{p \in M} f(p)
    .
\end{align}
Lebesgue integration can be generalized similarly. 
This would yield the same $\sup_{p \in M} f(p)$ formula but for all measurable functions that are bounded from above rather then all the functions for which the Darboux sum has a unique limit.
We will not detail the Lebesgue construction (see \textcite{kolokoltsov1997idempotent} for that) but proceed with the $\sup_{p \in M} f(p)$ formula as our definition of tropical integral.

\begin{remark}
Classic example of a function that is Lebesgue integrable but not Riemann integrable is the indicator function of the rational numbers $\mathbb{1}_{\mathbb{Q}}$. 
The limit \eqref{eq:lim_max_f} does not exist depending on our choice of points $p_S$ but in the Lebesgue sense we simply get $\sup_{x \in \mathbb{R}} \mathbb{1}_{\mathbb{Q}}(x)=1$ as expected.
\end{remark}

\begin{definition}
    Let $M$ be a manifold, then we define the set of measurable $\mathbb{R}_{\max}$-valued functions that are bounded from above as
    \begin{equation*}
        \iident{BA}(M)
        :=
        \iident{BA}(M,\mathbb{R}_{\max})
        :=
        \left\{\,
            f \in \mathbb{R}_{\max}^M
            \ \middle\vert \  
            \textstyle\sup_{p \in M} f(p) < \infty
            \ \text{and $f$ is measurable}
            \,
        \right\}
        .
    \end{equation*}
    Additionally if $f$ is not identical to $-\infty$ everywhere we say $f$ is \emph{proper}.
\end{definition}

Clearly BA($M$) is a tropical semimodule (i.e. a semimodule with respect to the tropical semiring) under pointwise addition and multiplication:
\begin{equation*}
    (f \oplus f')(p) := \max\{f(p),f'(p)\}
    \quad
    \text{and}
    \quad
    (a \odot f)(p) := a + f(p)
    ,
\end{equation*}
for all $a \in \mathbb{R}_{\max}$, $f,f' \in \iident{BA}(M)$ and all $p \in M$.

The functions in BA($M$) are exactly those for which the tropical integral exists.
\begin{definition}[Tropical integral]
    Let $M$ be a manifold then we call the mapping $\iident{BA}(M) \to \mathbb{R}_{\max}$ defined by
    \begin{equation*}
        f \mapsto \sup_{p \in M} f(p)
    \end{equation*}
    for $f \in \iident{BA}(M)$ the \emph{tropical integral} over $M$.
\end{definition}

We can easily check that the tropical integral is a \emph{tropical map} (i.e. a semimodule homomorphism with respect to the tropical semiring) from $\iident{BA}(M)$ to $\mathbb{R}_{\max}$ in the same way that the (linear) integral is a linear map from the integrable functions to $\mathbb{R}$. For all $a,b \in \mathbb{R}_{\max}$ and $f,f' \in \iident{BA}(M)$ we have:
\begin{equation*}
    \textstyle
    \sup_{p \in M}
    (a \odot f \oplus b \odot f')(p)
    =
    a \odot \left( \sup_{p \in M} f(p) \right)
    \oplus
    b \odot \left( \sup_{p \in M} f'(p) \right)
    .
\end{equation*}
        
Let $M$ be a homogeneous space of the Lie group $G$, let `$\cdot$' denote both the action of $G$ on $M$ and the corresponding representation on functions on $M$.
Since this representation does not affect the codomain of the functions it does not change the supremum, hence the tropical integral is always invariant:
\begin{equation*}
    \sup_{p \in M}\, (g \cdot f)(p)
    =
    \sup_{p \in M} f(p).
\end{equation*}

So now we have developed an alternative notion of linearity and an alternative to the (linear) invariant integral. 
We will use these elements to construct a new type of equivariant operator in the same manner as we did before with linear operators.

\subsection{Equivariant Tropical Operators}

The starting point for our equivariant linear operators was the integral operator from \eqref{eq:integral_operator}. We obtain the tropical analogue simply by replacing the linear semiring operations:

\begin{minipage}{\textwidth}
\begin{align}
    \nonumber
    (A f)(q) &:= 
    \tikzstyle{every node}=[anchor=base,rounded corners=1mm]
    \everymath{\displaystyle}
    \tikz[baseline]{
        \node[fill=RoyalBlue!20] (integral)
        {$\int_M$};
    }
    \ k_A (p,q)
    \,
    \tikz[baseline]{
        \node[fill=Plum!20] (multiplication)
        {$\cdot\vphantom{\vert}$};
    }
    \,
    f(p) \,\d p,
    \intertext{with the tropical semiring operations:}
    \label{eq:tropical_integral_operator}
    (T f)(q) &:=
    \tikzstyle{every node}=[anchor=base,rounded corners=1mm]
    \everymath{\displaystyle}
    \tikz[baseline]{
        \node[fill=RoyalBlue!20] (max)
        {$\sup_{p \in M}\vphantom{\int}$};
    }
    \ k_T (p,q)
    \,
    \tikz[baseline]{
        \node[fill=Plum!20] (addition)
        {$+\vphantom{\vert}$};
    }
    \,
    f(p).
\end{align}
{\begin{tikzpicture}[overlay]
        \path[-latex, very thick, color=RoyalBlue!50] (integral) edge [bend left] (max);
        \path[-latex, very thick, color=Plum!50] (multiplication) edge [bend right] (addition);
\end{tikzpicture}}
\end{minipage}

We can see that if $k_T \in \iident{BA}(M \times N)$ and $f \in \iident{BA}(M)$ then $Tf \in \iident{BA}(N)$.
It is straightforward to verify that $T$ is a tropical operator from  $\iident{BA}(N)$ to $\iident{BA}(N)$.

Now we can proceed in the exact same way as in section~\ref{subsection:equivariant_linear_operators} to find out when $T$ is equivariant.

\begin{lemma}[Equivariant tropical integral operators]
    \label{lem:equivariant_tropical_integral_operator}
    Let $M$ and $N$ be homogeneous spaces of a Lie group $G$.
    Let $T$ be a tropical integral operator \eqref{eq:tropical_integral_operator} from $\iident{BA}(M)$ to $\iident{BA}(N)$ with a kernel $k_T \in \iident{BA}(M \times N)$.
    Then 
    \begin{equation*}
    T(g\cdot f) = g \cdot (T f)
    \end{equation*}
    for all $g \in G$ and $f \in \iident{BA}(M)$ if and only if
    \begin{equation}
    \label{eq:tropical_equivariant_kernel_symmetry}
        k_T(g \cdot p,g \cdot q)
        =
        k_T(p,q)
    \end{equation}
    for all $g \in G$, $p \in M$ and $q \in N$.
\end{lemma}

\begin{proof}
    First we show that $T f \in \iident{BA}(N)$:
    \begin{align*}
        \sup_{q \in N} \ (T f)(q) 
        &= \sup_{q \in N} \ \left( \sup_{p \in M}\ k_T(p,q) + f(p) \right)
        \\
        &\leq 
        \left(\sup_{(p,q) \in M \times N}\ k_T(p,q)\right) + \left(\sup_{p \in M} f(p)\right)
        \\
        &
        < \infty
        ,
    \end{align*}
    since both $k_T$ and $f$ are bounded from above.
    
    ``$\Rightarrow$''

    Assuming $T$ to be equivariant, take an arbitrary $g \in G$ and $f \in \iident{BA}(M)$ and substitute the definition of the group representation and $T$ in
    \begin{equation*}
        g^{-1} \cdot T(g\cdot f)
        =
        T f
    \end{equation*}
    to find
    \begin{equation}
        \label{eq:TLO_1}
        \sup_{p \in M} k_T(p,g \cdot q) +  f (g^{-1} \cdot p)
        =
        \sup_{p \in M} k_T(p,q) + f(p)
    \end{equation}
    for all $q \in N$.
    Since the tropical integral is invariant under domain transformation we can substitute $p$ with $g \cdot p$ on the left without changing the result:
    \begin{equation*}
        \sup_{p \in M} k_T(g \cdot p,g \cdot q) +  f (p)
        =
        \sup_{p \in M} k_T(p,q) + f(p)
    \end{equation*}
    for all $q \in N$.
    The equality only holds for all $f$ if
    \begin{equation}
        \label{eq:TLO_2}
         k_T(g \cdot p,g \cdot q)
         =
         k_T(p,q)
    \end{equation}
    for all $p \in M$, $q \in N$ and $g \in G$.
    
    ``$\Leftarrow$''
    
    Assuming $k_T(g \cdot p,g \cdot q)=k_T(p,q)$ for all $g \in G$, $p \in M$ and $q \in N$ then \eqref{eq:TLO_2} follows for any choice of $f \in \iident{BA}(M)$, $g \in G$ and $q \in N$.
    Substituting the invariant tropical integral the other way yields 
    \eqref{eq:TLO_1}, which implies the equivariance 
    \begin{equation*}
        g^{-1} \cdot T(g\cdot f)
        =
        T f
    \end{equation*}
    since $q \in N$ is arbitrary.
    The function $f$ and group element $g$ were also chosen arbitrarily so equivariance follows for all $f \in \iident{BA}(M)$ and $g \in G$.
\end{proof}

Now we can use the same strategy as we did in the linear case and use the symmetry \eqref{eq:tropical_equivariant_kernel_symmetry} to fix the second argument of the kernel $k_T$.
Pick a $q_0 \in N$ then for all $q \in N$ there exists at least one $g_q \in G_{q_0}$, consequently
\begin{equation*}
    k_T(p,q) 
    = 
    k_T(p, g_q \cdot q_0)
    =
    k_T(g_q \cdot g_q^{-1} \cdot p, g_q \cdot q_0)
    =
    k_T(g_q^{-1} \cdot p, q_0)
\end{equation*}
for all $p \in M$ and $q \in N$.
From which we can define a reduced kernel $\kappa_T \in \iident{BA}(M)$ as
\begin{equation*}
    \kappa_T(p) := k_A (p, q_0)
\end{equation*}
that still allows us to recover the full kernel by
\begin{equation*}
    k_T(p,q)
    =
    k_T(g_q^{-1} \cdot p, q_0)
    =
    \kappa_T(g_q^{-1} \cdot p)
    .
\end{equation*}
This whole construction in the end gives us the same type of operator as in Theorem~\ref{thm:equivariant_linear_operators} but with the linear operations switched for tropical operations:
\begin{equation*}
    \everymath{\displaystyle}
    \tikzstyle{every node}=[anchor=base,rounded corners=1mm]
    \begin{split}
    (Af)(q) &:= 
    \tikz[baseline]{
        \node[fill=RoyalBlue!20] (integral)
        {$\int_M$};
    }
    \ (g_q \cdot \kappa_A)(p) 
    \,
    \tikz[baseline]{
        \node[fill=Plum!20] (multiplication)
        {$\cdot\vphantom{\vert}$};
    }
    \,
    f(p) \,\d p,
    \\[20pt]
    (T f)(q) &:=
    \tikz[baseline]{
        \node[fill=RoyalBlue!20] (max)
        {$\sup_{p \in M}\vphantom{\int}$};
    }
    \ (g_q \cdot \kappa_T)(p) 
    \,
    \tikz[baseline]{
        \node[fill=Plum!20] (addition)
        {$+\vphantom{\vert}$};
    }
    \,
    f(p).
    \end{split}
\end{equation*}
    \begin{tikzpicture}[overlay]
        \path[-latex, very thick, color=RoyalBlue!50] (integral) edge [bend left] (max);
        \path[-latex, very thick, color=Plum!50] (multiplication) edge [bend right] (addition);
    \end{tikzpicture}

We summarize this result in the following theorem.
\begin{theorem}[Equivariant tropical operators]
    \label{thm:equivariant_tropical_operators}
    Let $M$ and $N$ be homogeneous spaces of a Lie group $G$.
    Fix a $q_0 \in N$ and let $\kappa_T \in \iident{BA}(M)$ be \emph{compatible}, i.e
    \begin{equation*}
        \forall h \in G_{q_0}
        :
        h \cdot \kappa_A =  \kappa_A.
    \end{equation*}
    
    Then the operator $T$ defined by
    \begin{equation*}
        (T f)(q) :=  \sup_{p \in M}\ (g_q \cdot \kappa_T) (p) + f(p)
    \end{equation*}
    where for all $q \in N$ we choose any $g_q$ so that $g_q \cdot q_0 = q$, is a \emph{well defined tropical operator} from $\iident{BA}(M)$ to $\iident{BA}(N)$ that is \emph{equivariant} with respect to $G$.
    
    Conversely every tropical integral operator \eqref{eq:tropical_integral_operator} with a kernel in $\iident{BA}(M \times N)$ that is equivariant is of this form.
\end{theorem}

\begin{proof}(sketch)
   \ \\
    
    ``$\Rightarrow$''
    
    Assume we have a $\kappa_T \in \iident{BA}(M)$ that is compatible. Define
    \begin{equation*}
        k_T(p,q) := (g_q \cdot \kappa_T)(p)
    \end{equation*}
    we can check that $k_T$ is well defined and satisfies the requirements of Lemma~\ref{lem:equivariant_tropical_integral_operator} because of the compatibility condition on $\kappa_T$.
    
    ``$\Leftarrow$''
    
    Assume we have a tropical integral operator $T$, then by Lemma~\ref{lem:equivariant_tropical_integral_operator} we have a kernel $k_T \in \iident{BA}(M \times N)$ that satisfies \eqref{eq:tropical_equivariant_kernel_symmetry}.
    Fix a $q_0 \in N$ and define
    \begin{equation*}
        \kappa_T(p) := k_T(p,q_0)
        ,
    \end{equation*}
    then we can check that $\kappa_T$ satisfies the compatibility condition.
\end{proof}

\begin{example}[Max pooling]
Let $G=(\mathbb{R}^n,+)$ be the translation group and $M=N=\mathbb{R}^n$.
Pick a subset $S \subset \mathbb{R}^n$ and define
\begin{equation*}
    \kappa_T(p)
    :=
    \begin{cases}
        \ 0 & \text{if } p \in S,
        \\
        \ -\infty & \text{elsewhere}.
    \end{cases}
\end{equation*}
Then the corresponding operator $T$ equals
\begin{align*}
    (T f)(\bm{y})
    &=
    \sup_{\bm{x} \in \mathbb{R}^n}
    (\bm{y} \cdot \kappa_T)(\bm{x})
    +
    f(\bm{x})
    \\
    &=
    \sup_{\bm{x} \in \mathbb{R}^n}
    \kappa_T(\bm{x}-\bm{y})
    +
    f(\bm{x})
    \\
    &=
    \sup_{\bm{x} \in \bm{y}+S} f(\bm{x})
    ,
\end{align*}
so at each point $\bm{y}$ the output equals the supremum of the input in the subset $\bm{y}+S$.
Think of this a continous version the shift-invariant max pooling operation usually seen in CNNs.
In this light we can see that max pooling is a tropical operator.
\end{example}

\begin{example}[Tropical convolution]
Let $G=M=N$ be some Lie group, then we call the equivariant tropical operation \emph{tropical convolution} or \emph{morphological convolution}, which we denote as
\begin{equation*}
    (\kappa \square_G f)(h)
    :=
    \sup_{g \in G}\ (h \cdot \kappa)(g) + f(g)
    .
\end{equation*}
The name morphological convolution comes from the field of grayscale morphology where these types of operators have previously been used for image processing applications, see \textcite{wikipedia2021morphology}.
An example use of these types of operations in neural networks can be found in \textcite{smets2021pdebased}.
\end{example}

\begin{example}[Pointwise ReLU]
    Let $G=M=N$ be some Lie group and let $f \in \iident{BA}(M)$.
    Define a kernel
    \begin{equation*}
        \kappa_T(g) := 
        \begin{cases}
            \ 0 \qquad & \text{if } g=e,
            \\
            \ -\sup_{h \in G} f(h) & \text{else.}
        \end{cases}
    \end{equation*}
    Then applying the corresponding operator to $f$ gives
    \begin{align*}
        (Tf)(h)
        &=
        \sup_{g \in G} 
        (h \cdot \kappa_T)(g)
        +
        f(g)
        \\
        &=
        \max \left\{
            f(h)
            ,\,
            \sup_{g \in G}
            \left( -\sup_{z \in G} f(z) \right)
            +f(g)
        \right\}
        \\
        &=
        \max \left\{
            f(h)
            ,\,
            0
        \right\}
        \\
        &=
        \relu(f(h))
        .
    \end{align*}
\end{example}

We have not said anything about the boundedness of $T$ in Theorem~\ref{thm:equivariant_tropical_operators} since it is not clear what metric or norm we should consider on the function space BA.
In the following lemma we detail some reasonable conditions under which $T$ will be bounded in the supremum norm.

\begin{lemma}
    In the setting of Theorem~\ref{thm:equivariant_tropical_operators}: if $f \in \iident{B}(M)$ and $\sup_{p \in M} \kappa_T(p)=0$ then $Tf \in \iident{B}(N)$ and $T$ is a bounded operator from $\iident{B}(M)$ to $\iident{B}(N)$ in the supremum norm.
\end{lemma}

\begin{proof}
\allowdisplaybreaks
\setlength{\jot}{2.0ex}
    \begin{align*}
        \left\| T f \right\|_\infty
        &=
        \sup_{q \in N}
        \left|\ 
            \sup_{p \in M}\ 
            (g_q \cdot \kappa_T)(p)
            +
            f(p)
        \ \right|
        \\
        &\leq
        \sup_{q \in N}
        \left|\ 
            \sup_{p \in M}
            (g_q \cdot \kappa_T)(p)
            +
            \sup_{p \in M}
            f(p)
        \ \right|
        \\
        &=
        \sup_{q \in N}
        \left|\ 
            \sup_{p \in M}
            \kappa_T(p)
            +
            \sup_{p \in M}
            f(p)
        \ \right|
        \\
        &=
        \left|\ 
        \sup_{p \in M}
            \kappa_T(p)
            +
            \sup_{p \in M}
            f(p)
        \ \right|
        \\
        &=
        \left|\ 
            \sup_{p \in M}
            f(p)
        \ \right|
        \\
        &\leq
        \sup_{p \in M} \left| f(p) \right|
        \\
        &= 
        \left\|\, f \,\right\|_\infty
        .
    \end{align*}
\end{proof}

We have seen how we can use semirings, and the tropical semiring in particular, to develop classes of equivariant operators other than linear.
From the examples we saw we can conclude that many operators currently in use in neural networks (particularly the ReLU and max pooling activation functions) are special cases of tropical or tropically affine operators and fit well inside the equivariant semimodule homomorphism framework.

\setlength{\bibitemsep}{\baselineskip}
\renewcommand{\mkbibnamefamily}[1]{\textsc{#1}}
\printbibliography

\end{document}